\documentclass{article} 
\usepackage[preprint]{configs/colm2026_conference}

\usepackage{configs/commands}

\usepackage{microtype}
\usepackage{hyperref}
\usepackage{url}
\usepackage{booktabs}
\usepackage{multirow}
\usepackage{wrapfig}

\usepackage{lineno}

\usepackage[utf8]{inputenc}
\usepackage[T1]{fontenc}
\usepackage{caption} 
\usepackage{colortbl}
\usepackage{xcolor}
\usepackage{graphicx}
\usepackage{pgfplots}
\pgfplotsset{compat=1.18}
\usetikzlibrary{patterns,positioning,decorations.text}
\usepgfplotslibrary{groupplots}
\usepackage{configs/tkz-kiviat}
\usepackage[table]{xcolor}
\usepackage{pgf}
\usepackage[export]{adjustbox}

\usepackage{longtable}
\usepackage{appendix}
\usepackage{setspace}
\usepackage{titletoc}
\usepackage{makecell}
\usepackage{mdframed}
\usepackage{soul}
\usepackage{float}
\usepackage[misc]{ifsym}
\usepackage{wasysym}

\definecolor{darkblue}{rgb}{0, 0, 0.5}
\definecolor{deepgreen}{RGB}{0,100,0}

\definecolor{heat00}{HTML}{D18683}
\definecolor{heat08}{HTML}{D99693}
\definecolor{heat17}{HTML}{E8BBB6}
\definecolor{heat33}{HTML}{F3DAD6}
\definecolor{heat50}{HTML}{FBF4EC}
\definecolor{heat67}{HTML}{D9E8DC}
\definecolor{heat80}{HTML}{9FC4AE}

\definecolor{g1col}{RGB}{30,64,175}    
\definecolor{g1row}{RGB}{239,246,255}  
\definecolor{g2col}{RGB}{154,52,18}    
\definecolor{g2row}{RGB}{255,247,237}  
\definecolor{g3col}{RGB}{20,83,45}     
\definecolor{g3row}{RGB}{240,253,244}  
\definecolor{g4col}{RGB}{76,29,149}    
\definecolor{g4row}{RGB}{245,243,255}  
\definecolor{g5col}{RGB}{159,29,29}    
\definecolor{g5row}{RGB}{255,241,242}  
\newcommand{\rgb}[1]{%
\pgfmathsetmacro{\clamped}{min(max(#1,0),80)}%
\pgfmathtruncatemacro{\leqA}{\clamped <= 8.33}%
\pgfmathtruncatemacro{\leqB}{\clamped <= 16.67}%
\pgfmathtruncatemacro{\leqC}{\clamped <= 33.33}%
\pgfmathtruncatemacro{\leqD}{\clamped <= 50}%
\pgfmathtruncatemacro{\leqE}{\clamped <= 66.67}%
\ifnum\leqA=1
\pgfmathsetmacro{\percent}{\clamped / 8.33 * 100}%
\edef\temp{\noexpand\cellcolor{heat08!\percent!heat00}}\temp%
\else\ifnum\leqB=1
\pgfmathsetmacro{\percent}{(\clamped - 8.33) / (16.67 - 8.33) * 100}%
\edef\temp{\noexpand\cellcolor{heat17!\percent!heat08}}\temp%
\else\ifnum\leqC=1
\pgfmathsetmacro{\percent}{(\clamped - 16.67) / (33.33 - 16.67) * 100}%
\edef\temp{\noexpand\cellcolor{heat33!\percent!heat17}}\temp%
\else\ifnum\leqD=1
\pgfmathsetmacro{\percent}{(\clamped - 33.33) / (50 - 33.33) * 100}%
\edef\temp{\noexpand\cellcolor{heat50!\percent!heat33}}\temp%
\else\ifnum\leqE=1
\pgfmathsetmacro{\percent}{(\clamped - 50) / (66.67 - 50) * 100}%
\edef\temp{\noexpand\cellcolor{heat67!\percent!heat50}}\temp%
\else
\pgfmathsetmacro{\percent}{(\clamped - 66.67) / (80 - 66.67) * 100}%
\edef\temp{\noexpand\cellcolor{heat80!\percent!heat67}}\temp%
\fi\fi\fi\fi\fi%
}

\newcommand{\swatch}[1]{%
\raisebox{0.2ex}{\fcolorbox{gray!50}{#1}{\rule{0pt}{0.9ex}\rule{0.9ex}{0pt}}}%
}
\hypersetup{colorlinks=true, citecolor=darkblue, linkcolor=darkblue, urlcolor=darkblue}

\title{WildTableBench: Benchmarking Multimodal Foundation Models on Table Understanding In the Wild}

\author{
Junzhe Huang$^{1}$,
Xiaoxiao Sun$^{2}$,
Yan Yang$^{3}$,
Yuxuan Hou$^{4}$,
Ruotian Zhang$^{4}$,
Sirui Li$^{5}$,\\[2pt]
~\textbf{Hehe Fan}$^{4}$,
\textbf{Serena Yeung-Levy}$^{2}$,
\textbf{Xin Yu}$^{6}$\,$^{\textrm{\Letter}}$
\\[6pt]
$^{1}$The University of Queensland,
$^{2}$Stanford University,
$^{3}$The Australian National University
\\[2pt]
$^{4}$Zhejiang University,
$^{5}$Murdoch University,
$^{6}$The University of Adelaide \\[2pt]
\textbf{Dataset:} \raisebox{-0.2em}{\includegraphics[height=1.1em]{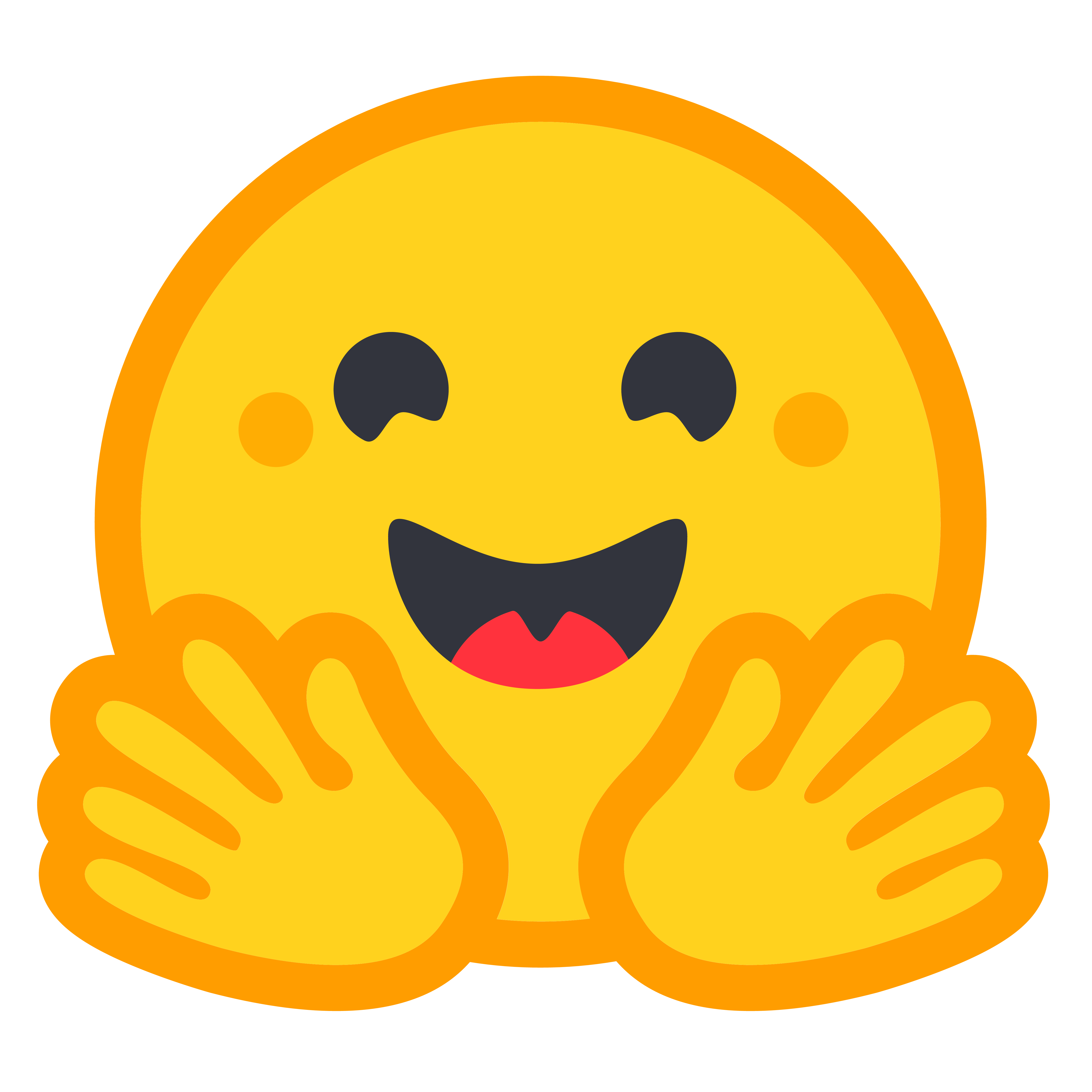}}
\ \href{https://huggingface.co/datasets/jzhuang/WildTableBench}{\texttt{https://huggingface.co/datasets/jzhuang/WildTableBench}}\\
\textbf{Code:} \raisebox{-0.2em}{\includegraphics[height=1.1em]{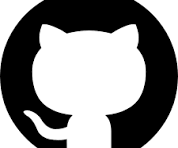}} \                                                           
  \href{https://github.com/hjzhe/WildTableBench}{\texttt{https://github.com/hjzhe/WildTableBench}}\\
\textbf{Leaderboard:} \raisebox{-0.2em}{\includegraphics[height=1.1em]{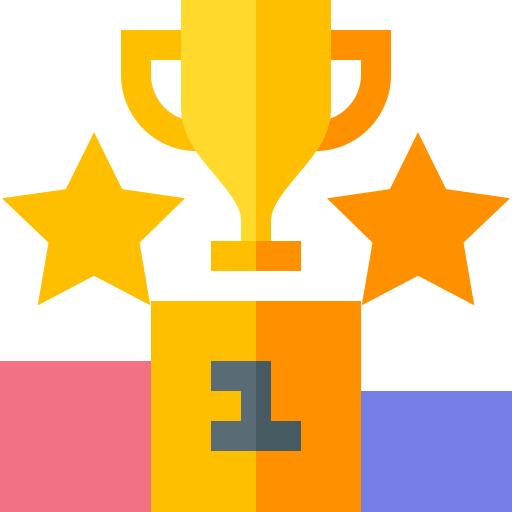}}
\ \href{https://hjzhe.github.io/WildTableBench}{\texttt{https://hjzhe.github.io/WildTableBench}}
}

\newcommand{\eg}{\textit{e.g.}}

\begin{document}

\newcommand{\tablerowpad}{\raisebox{0pt}[1.05em][1.05em]{}}

\ifcolmsubmission
\linenumbers
\fi

\maketitle

\definecolor{gbar}{RGB}{52,168,83}
\definecolor{obar}{RGB}{90,90,90}
\definecolor{kbar}{RGB}{66,133,244}
\definecolor{bbar}{RGB}{245,140,40}
\definecolor{cbar}{RGB}{196,113,67}
\definecolor{qbar}{RGB}{115,87,217}
\definecolor{zbar}{RGB}{210,60,60}
\colorlet{bc0}{gbar!90!black}  \colorlet{bc2}{gbar!50}
\colorlet{bc1}{kbar}
\colorlet{bc3}{obar}           \colorlet{bc9}{obar!65}
\colorlet{bc12}{obar!50}       \colorlet{bc13}{obar!35}
\colorlet{bc4}{bbar}
\colorlet{bc5}{cbar}           \colorlet{bc6}{cbar!55}
\colorlet{bc7}{qbar}           \colorlet{bc8}{qbar!65}
\colorlet{bc10}{qbar!40}
\colorlet{bc11}{zbar}

\newcommand{\cattag}[2]{%
  \tikz[baseline=(t.base)]\node(t)[rounded corners=3pt,fill=#1,%
    inner xsep=3pt,inner ysep=1.2pt]{\fontsize{5.5}{6}\selectfont\textbf{#2}};%
}
\newcommand{\metatag}[2]{%
  \tikz[baseline=(t.base)]\node(t)[rounded corners=2.5pt,fill=#1,%
    inner xsep=4pt,inner ysep=1.8pt]{\fontsize{7}{8.5}\selectfont\textbf{#2}};%
}

\begin{figure*}[h]
\vspace{-5pt}
\centering

\newsavebox{\imgbox}%
\newcommand{\wrongbox}[1]{%
  \begin{tikzpicture}[baseline=(num.base)]
    \node[inner sep=2.8pt, minimum width=2.4em, minimum height=1.5em,
          fill=black!2, draw=black!22, line width=0.3pt,
          font=\small\bfseries, text=black!85]
      (num) {#1};
    \draw[red!45, line width=0.55pt, line cap=round]
      ([shift={(2pt,2pt)}]num.south west) -- ([shift={(-2pt,-2pt)}]num.north east)
      ([shift={(-2pt,2pt)}]num.south east) -- ([shift={(2pt,-2pt)}]num.north west);
  \end{tikzpicture}%
}
\noindent
\begin{minipage}[t]{0.475\textwidth}\vspace{0pt}%
  \setlength{\fboxsep}{0pt}\setlength{\fboxrule}{0.25pt}%
  \fbox{\includegraphics[width=\dimexpr\linewidth-0.5pt\relax]{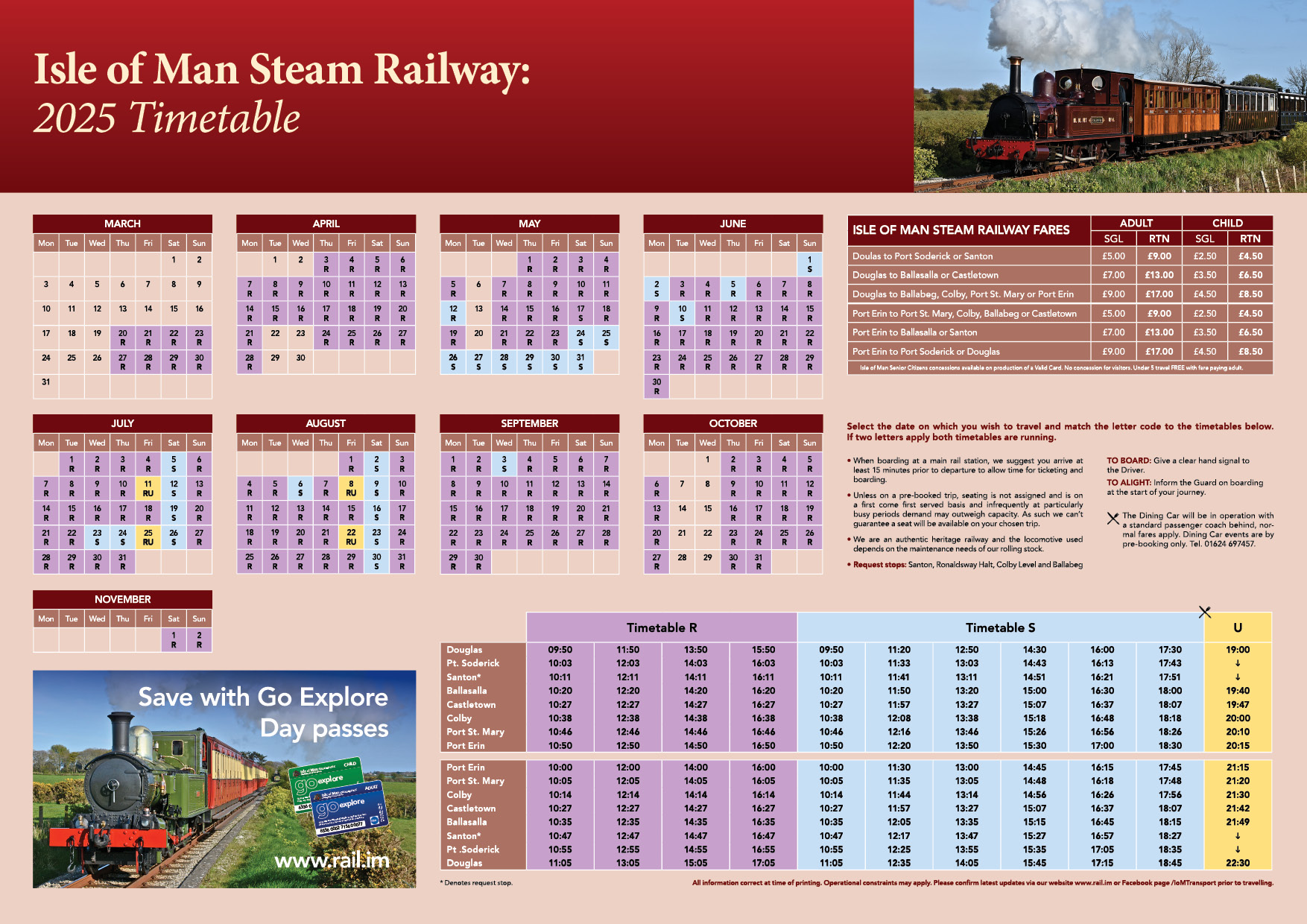}}%
\end{minipage}\hfill
\begin{minipage}[t]{0.50\textwidth}\vspace{0pt}%
  \small
  \hspace*{0.01\linewidth}%
  \begin{minipage}[t]{0.95\linewidth}\vspace{0pt}%
    \raggedright
    \textbf{Question:} \textit{A person plans to experience the scenic train ride after 5:00~PM but before 7:00~PM. How many days fit this schedule in 2025?}%
    \par\vspace{2pt}
    \textbf{Answer:}\;\,\colorbox{green!50}{\;\textbf{26}\;}%
    \par\vspace{2pt}
    {\small\begin{tabular*}{\linewidth}{@{}l@{\extracolsep{\fill}}c@{}}
      \raisebox{-2.3pt}{\includegraphics[height=11pt]{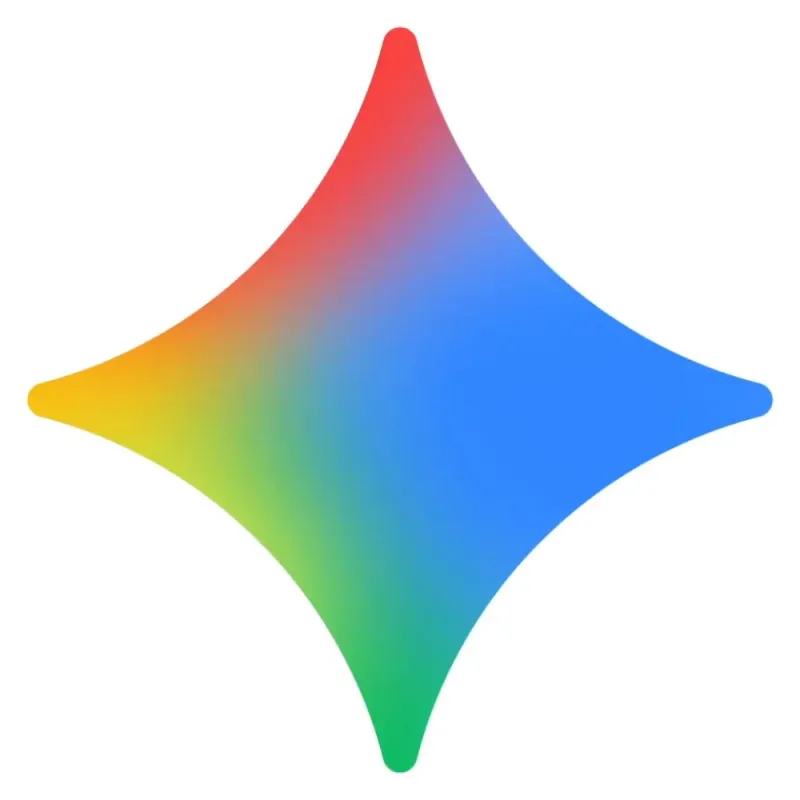}}\hspace{4pt}Gemini-3-Pro:
        & \wrongbox{18} \\[5pt]
      \raisebox{-2.3pt}{\includegraphics[height=11pt]{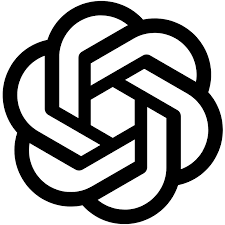}}\hspace{4pt}GPT-5.2-Thinking:
        & \wrongbox{2} \\[5pt]
      \raisebox{-2.3pt}{\includegraphics[height=11pt]{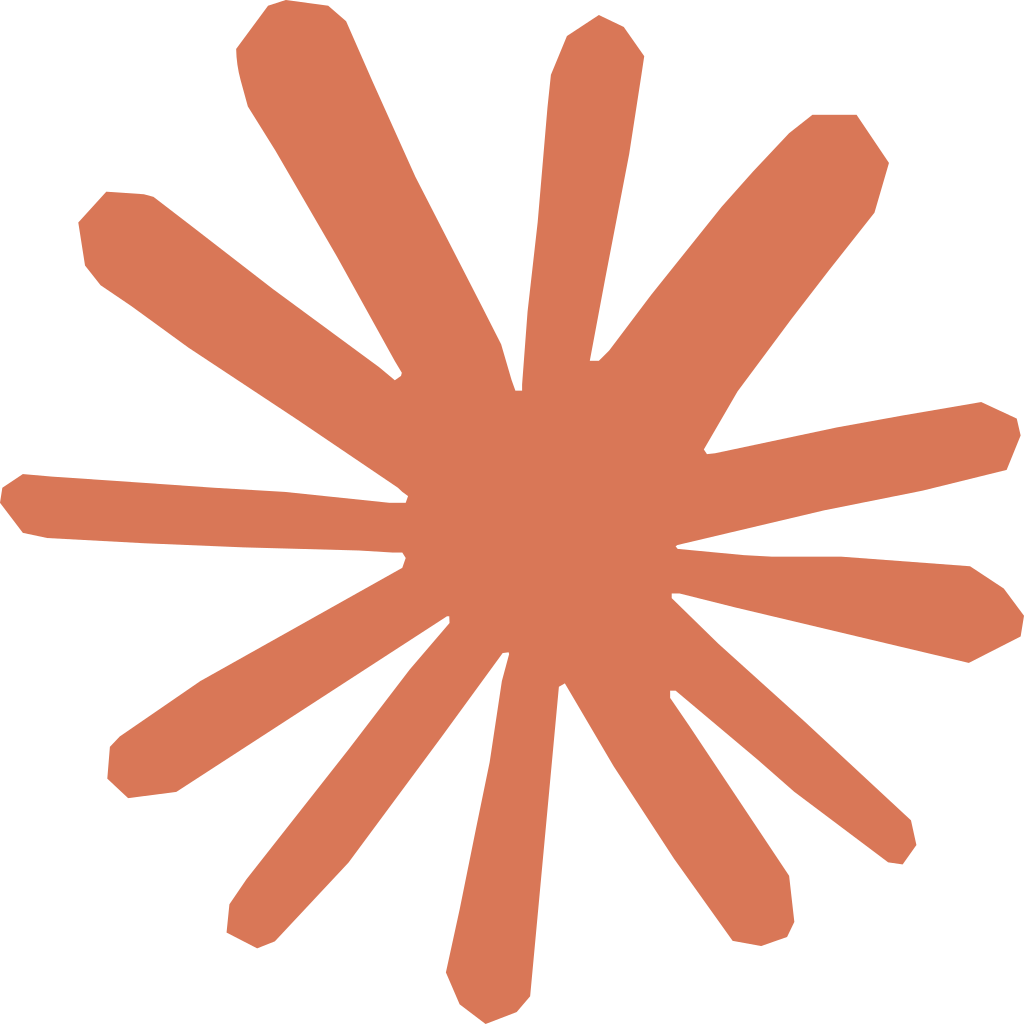}}\hspace{4pt}Claude-Opus-4.6:
        & \wrongbox{4} \\
    \end{tabular*}}%
    \par\vspace{2pt}
    {\color{black!50}\hrule height 0.3pt}\par\vspace{5pt}
    \fontsize{7}{9}\selectfont
    \textcolor{black!55}{{Image Domain:}}\;\metatag{blue!8}{Transportation}\quad\\
    \textcolor{black!55}{{Question Category:}}\;\metatag{teal!10}{Color-based Counting}\\
    \textcolor{black!55}{{Required Skills:}}\;\,\metatag{orange!10}{Multi-hop Reasoning}\;\metatag{red!7}{Visual Parsing}\;%
  \end{minipage}%
\end{minipage}%
\par\vspace{2pt}{\centering\small\textbf{(a)} A benchmark example\par}
\vspace{-5pt}
\noindent
\begin{minipage}[t]{0.65\textwidth}
\null\vspace{-105pt}
\begin{tikzpicture}[yscale=0.035]
  \def\bw{0.38}
  \def\sp{0.69}
  \pgfmathsetmacro{\xmax}{13*\sp+\bw/2+0.1}

  \foreach \y in {10,20,...,70}{
    \draw[black!8] ({-\bw/2-0.08},\y) -- (\xmax,\y);
    \node[left,font=\fontsize{5}{5.5}\selectfont,text=black!40]
      at ({-\bw/2-0.1},\y) {\y};
  }
  \draw[black!15] ({-\bw/2-0.08},0) -- (\xmax,0);
  \node[left,font=\fontsize{5}{5.5}\selectfont,text=black!40]
    at ({-\bw/2-0.1},0) {0};
  \node[rotate=90,font=\fontsize{6}{7}\selectfont,anchor=south]
    at ({-\bw/2-0.42},35) {Accuracy (\%)};

  \foreach \i/\val in {0/67.9, 1/49.9, 2/48.2, 3/46.6,
                        4/46.2, 5/45.7, 6/35.0, 7/34.7,
                        8/28.2, 9/24.8, 10/24.5, 11/24.4,
                        12/14.8, 13/5.7}{
    \pgfmathsetmacro{\xc}{\i*\sp}
    \fill[bc\i, rounded corners=1pt, draw=bc\i!55!black, line width=0.15pt]
      ({\xc-\bw/2},0) rectangle ({\xc+\bw/2},\val);
    \node[above, font=\fontsize{4.5}{5}\selectfont\bfseries,
          inner sep=0.3pt] at (\xc,\val) {\val};
  }

  \pgfmathsetmacro{\xm}{0*\sp}
  \node[above=4pt, inner sep=0pt] at (\xm,67.9)
    {\includegraphics[height=5.5pt]{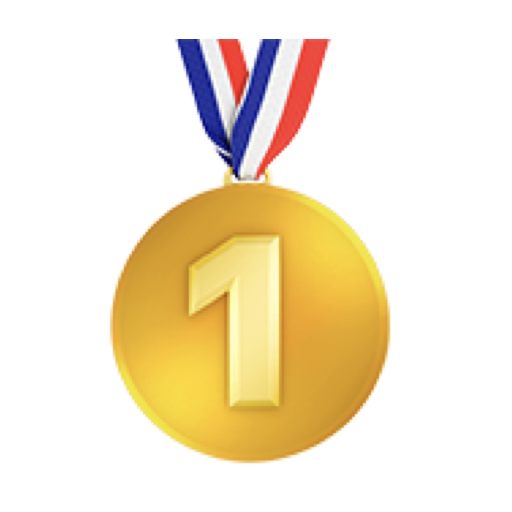}};
  \pgfmathsetmacro{\xm}{1*\sp}
  \node[above=4pt, inner sep=0pt] at (\xm,49.9)
    {\includegraphics[height=5.5pt]{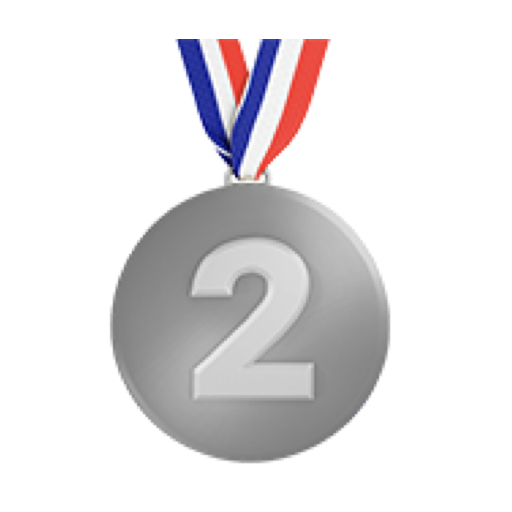}};
  \pgfmathsetmacro{\xm}{2*\sp}
  \node[above=4pt, inner sep=0pt] at (\xm,48.2)
    {\includegraphics[height=5.5pt]{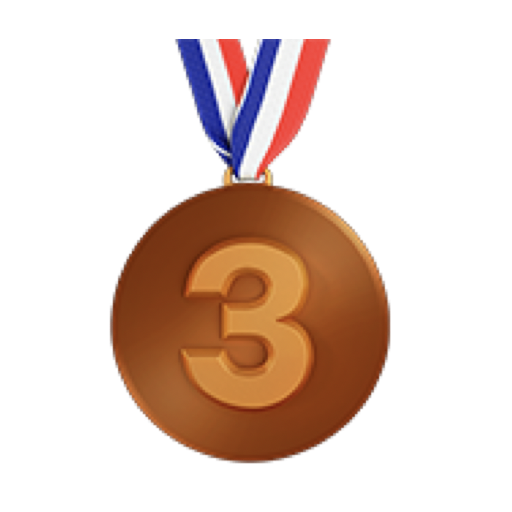}};

  \foreach \i/\icon in {0/gemini, 1/kimi, 2/gemini, 3/openai,
                         4/bytedance, 5/claude, 6/claude, 7/qwen,
                         8/qwen, 9/openai, 10/qwen, 11/zai,
                         12/openai, 13/openai}{
    \pgfmathsetmacro{\xc}{\i*\sp}
    \node[below=2pt, inner sep=0] at (\xc,0)
      {\includegraphics[height=6.5pt]{icon/\icon.png}};
  }

  \foreach \i/\lbl in {%
    0/{Gemini-3-Pro}, 1/{Kimi-K2.5}, 2/{Gemini-3-Flash},
    3/{GPT-5.2}, 4/{Seed-2.0-Pro}, 5/{Claude-Opus-4.6},
    6/{Claude-Sonnet-4.6}, 7/{Qwen3-VL-235B-Thinking},
    8/{Qwen3-VL-32B-Thinking}, 9/{GPT-5-mini},
    10/{Qwen3-VL-32B-Instruct}, 11/{GLM-4.6V},
    12/{o3}, 13/{GPT-4o}}{
    \pgfmathsetmacro{\xc}{\i*\sp}
    \node[below=11pt, rotate=20, anchor=north east,
          font=\fontsize{4.5}{5}\selectfont, inner sep=0] at (\xc,0) {\lbl};
  }
\end{tikzpicture}
\par\vspace{.2em}{\centering\small\textbf{(b)} Overall model accuracy\par}
\end{minipage}\hfill
\begin{minipage}[t]{0.3\textwidth}
\raggedleft
\begin{tikzpicture}[xshift=0.8cm]
  \path[use as bounding box] (-1.8,-2) rectangle (1.7,1.85);
  \tkzKiviatDiagram[
    lattice      = 4,
    gap          = 0.32,
    space        = 0.14,
    label space  = 0.14,
    step         = 1,
    radial style/.style  = {thin, black!18},
    lattice style/.style = {thin, black!12},
    label style/.style   = {inner sep=0pt, outer sep=0pt}
  ]{{},{},{},{},{}}

  \def\arcr{1.48}
  \foreach \sep in {36, 108, 180, 252, 324} {
    \fill[black!15] (\sep:\arcr) circle (0.5pt);
  }
  \draw[black!40, line width=0.6pt, line cap=round]
    ({-27}:\arcr) arc ({-27}:{27}:\arcr);
  \path[decorate, decoration={text along path,
    text={|\fontsize{5}{6}\selectfont\bfseries\color{black!40}|Color},
    text align=center, raise=4pt, reverse path}]
    ({-14}:\arcr) arc ({-14}:{14}:\arcr);
  \draw[black!40, line width=0.6pt, line cap=round]
    ({45}:\arcr) arc ({45}:{99}:\arcr);
  \path[decorate, decoration={text along path,
    text={|\fontsize{5}{6}\selectfont\bfseries\color{black!40}|Cell-Level},
    text align=center, raise=4pt, reverse path}]
    ({50}:\arcr) arc ({50}:{94}:\arcr);
  \draw[black!40, line width=0.6pt, line cap=round]
    ({117}:\arcr) arc ({117}:{171}:\arcr);
  \path[decorate, decoration={text along path,
    text={|\fontsize{5}{6}\selectfont\bfseries\color{black!40}|Numerical},
    text align=center, raise=4pt, reverse path}]
    ({122}:\arcr) arc ({122}:{166}:\arcr);
  \draw[black!40, line width=0.6pt, line cap=round]
    ({189}:\arcr) arc ({189}:{243}:\arcr);
  \path[decorate, decoration={text along path,
    text={|\fontsize{5}{6}\selectfont\bfseries\color{black!40}|Verification},
    text align=center, raise=-7pt}]
    ({194}:\arcr) arc ({194}:{238}:\arcr);
  \draw[black!40, line width=0.6pt, line cap=round]
    ({261}:\arcr) arc ({261}:{315}:\arcr);
  \path[decorate, decoration={text along path,
    text={|\fontsize{5}{6}\selectfont\bfseries\color{black!40}|Hypothetical},
    text align=center, raise=-7pt}]
    ({266}:\arcr) arc ({266}:{310}:\arcr);


  \def\minA{20.0}  \def\maxA{56.5}   
  \def\minB{47.0}  \def\maxB{62.3}   
  \def\minC{45.9}  \def\maxC{71.8}   
  \def\minD{56.7}  \def\maxD{71.6}   
  \def\minE{53.9}  \def\maxE{75.3}   

  \newcommand{\norm}[3]{\pgfmathsetmacro{\normed}{1.0+(#1-#2)/(#3-#2)*3.3}}

  \norm{56.5}{\minA}{\maxA}\let\gA\normed
  \norm{62.3}{\minB}{\maxB}\let\gB\normed
  \norm{71.8}{\minC}{\maxC}\let\gC\normed
  \norm{71.6}{\minD}{\maxD}\let\gD\normed
  \norm{75.3}{\minE}{\maxE}\let\gE\normed
  \tkzKiviatLine[semithick, gbar!85!black,
    fill=gbar, opacity=0.12,
    mark=*, mark size=0.8pt](\gA, \gB, \gC, \gD, \gE)

  \norm{28.7}{\minA}{\maxA}\let\gA\normed
  \norm{52.3}{\minB}{\maxB}\let\gB\normed
  \norm{45.9}{\minC}{\maxC}\let\gC\normed
  \norm{56.7}{\minD}{\maxD}\let\gD\normed
  \norm{55.1}{\minE}{\maxE}\let\gE\normed
  \tkzKiviatLine[semithick, obar,
    fill=obar, opacity=0.08,
    mark=*, mark size=0.8pt](\gA, \gB, \gC, \gD, \gE)

  \norm{38.3}{\minA}{\maxA}\let\gA\normed
  \norm{47.0}{\minB}{\maxB}\let\gB\normed
  \norm{51.1}{\minC}{\maxC}\let\gC\normed
  \norm{67.2}{\minD}{\maxD}\let\gD\normed
  \norm{55.1}{\minE}{\maxE}\let\gE\normed
  \tkzKiviatLine[semithick, kbar,
    fill=kbar, opacity=0.08,
    mark=*, mark size=0.8pt](\gA, \gB, \gC, \gD, \gE)

  \norm{20.0}{\minA}{\maxA}\let\gA\normed
  \norm{51.7}{\minB}{\maxB}\let\gB\normed
  \norm{46.6}{\minC}{\maxC}\let\gC\normed
  \norm{61.2}{\minD}{\maxD}\let\gD\normed
  \norm{53.9}{\minE}{\maxE}\let\gE\normed
  \tkzKiviatLine[semithick, cbar,
    fill=cbar, opacity=0.08,
    mark=*, mark size=0.8pt](\gA, \gB, \gC, \gD, \gE)

  \newcommand{\ann}[5]{%
    \node[font=\fontsize{3.3}{4}\selectfont\bfseries, text=black, fill=white, fill opacity=0.8, text opacity=1, rounded corners=0.5pt, anchor=#4, inner sep=0.5pt] at (#1,#2) {#3};}

  \ann{1.00}{0.0}{56.5}{west}{gbar!80!black}
  \ann{0.56}{1.10}{62.3}{south west}{gbar!80!black}
  \ann{-0.72}{0.72}{71.8}{south east}{gbar!80!black}
  \ann{-0.80}{-0.90}{71.6}{north east}{gbar!80!black}
  \ann{0.15}{-1.26}{75.3}{north west}{gbar!80!black}

  \ann{0.52}{0.08}{28.7}{south west}{obar}
  \ann{0.30}{0.60}{52.3}{south west}{obar}
  \ann{0.10}{0.12}{45.9}{south east}{obar}
  \ann{0.10}{-0.10}{56.7}{north east}{obar}
  \ann{0.15}{-0.42}{55.1}{north west}{obar}

  \ann{0.70}{-0.08}{38.3}{north west}{kbar}
  \ann{0.00}{0.35}{47.0}{south west}{kbar}
  \ann{-0.50}{0.20}{51.1}{south east}{kbar}
  \ann{-0.90}{-0.69}{67.2}{north west}{kbar}

  \ann{0.15}{0.05}{20.0}{north}{cbar}
  \ann{-0.74}{-0.40}{61.2}{north west}{cbar}

  \node[rounded corners=3pt, draw=black!20, line width=0.3pt,
        fill=white, inner sep=2pt, anchor=south,
        font=\fontsize{4}{5.5}\selectfont]
    at (0,-2.4) {%
      \begin{tabular}{@{}l@{\enspace}l@{}}
        \tikz[baseline=-0.5ex]\fill[gbar!90!black] (0,0) circle (1.5pt);~Gemini-3-Pro &
        \tikz[baseline=-0.5ex]\fill[obar] (0,0) circle (1.5pt);~GPT-5.2 \\
        \tikz[baseline=-0.5ex]\fill[kbar] (0,0) circle (1.5pt);~Kimi-K2.5 &
        \tikz[baseline=-0.5ex]\fill[cbar] (0,0) circle (1.5pt);~Claude-Opus-4.6
      \end{tabular}};
\end{tikzpicture}
\\\vspace{1em}{\centering\small\textbf{(c)} Domain accuracy}\hspace{1em}
\end{minipage}

\vspace{-4pt}
\caption{
WildTableBench overview.
\textbf{(a)} A benchmark example requiring multi-hop reasoning over a real-world train schedule. All three frontier models answer incorrectly.
\textbf{(b)} Overall accuracy of 14 representative models; most fall below 50\%.
\textbf{(c)} Category-level comparison of the top-4 models from different providers. Performance varies markedly across question types, with Color-related questions proving most challenging.
}
\label{fig:teaser}
\end{figure*}

\begin{abstract}


Using multimodal foundation models to analyze table images is a high-value yet challenging application in consumer and enterprise scenarios. Despite its importance, current evaluations rely largely on structured-text tables or clean rendered images, leaving the visual complexity of \emph{in-the-wild} table images underexplored. Such images feature varied layouts and diverse domains that demand sophisticated structural perception and numerical reasoning. To bridge this gap, we introduce WildTableBench, the first question-answering benchmark for naturally occurring table images from real-world settings. WildTableBench comprises 402 high-information-density table images collected from online forums and websites across diverse domains, together with 928 manually annotated and verified questions spanning 17 subtypes across five categories. We evaluate 21 frontier proprietary and open-source multimodal foundation models on this benchmark. Only one model exceeds 50\% accuracy, while all remaining models range from 4.1\% to 49.9\%. We further conduct diagnostic analyses to characterize model failures and reveal persistent weaknesses in structural perception and reasoning. These results and analyses provide useful insights into current model capabilities and establish WildTableBench as a valuable diagnostic benchmark for table image understanding.

\end{abstract}

\section{Introduction}


Table image understanding has emerged as an important and increasingly studied capability for multimodal foundation models~\citep{kim2404tablevqa,wu2025tablebench}, particularly in practical settings where tables are often encountered as screenshots, scans, or photographs. Unlike clean digital tables, these real-world table images are often visually dense, structurally irregular, and difficult to parse. However, existing benchmarks~\citep{zhu2025tableeval,wu2025realhitbench} are largely built on structured-text tables or rendered images, creating a gap between current evaluation settings and the challenges of practical table understanding.


To address this gap, we introduce WildTableBench, a benchmark for evaluating multimodal foundation models on naturally occurring table images from the real-world setting. WildTableBench contains 402 table images collected from publicly available web sources, covering diverse visual styles, structural patterns, and thematic domains, such as \textit{worksheets}, \textit{transportation}, \textit{sports}, \textit{education}, \textit{finance}, \textit{science}, \textit{engineering}, and \textit{healthcare}. These images include \textit{screenshots}, \textit{scans}, and \textit{photographs} of tables shared in realistic contexts.
As shown in Figure~\ref{fig:teaser}, they exhibit visual complexity and structural irregularities that are common in real-world table images, such as irregular cell spans, dense content, decorative formatting, and skewed orientations. These characteristics are largely underrepresented in prior evaluations. By incorporating these characteristics, WildTableBench significantly narrows the gap between existing table evaluations and practical applications.



Beyond image collection efforts, we also design and annotate high-quality evaluation questions tailored to real-world table understanding. We first examine practical use cases from public forums, websites, and user-shared table images, and organize them into a taxonomy of 17 question subtypes in five categories. This taxonomy supports a comprehensive evaluation of model capabilities, including cell-level understanding, numerical reasoning, fact verification, hypothetical reasoning, and color-based reasoning. For each subtype, we use LLMs to draft seed questions, which are then refined and extended by professional annotators to ensure that they are clear, unambiguous, and verifiable from the image alone. Rather than emphasizing simple surface-level lookup, we prioritize questions that require non-trivial visual grounding and reasoning over table structure and values. The final benchmark contains 928 high-quality questions, each paired with a manually verified reference answer and cross-checked by annotators.

We systematically evaluate 21 frontier proprietary (\eg, Gemini, GPT, Claude, Seed-series) and open-source (\eg, Kimi, GLM, Qwen-series) multimodal foundation models, including their instruction-tuned and reasoning variants. The best model, Gemini-3-Pro, answers only 67.9\% of questions correctly. All other models score below 50.0\%, in some cases less than 5\%, with substantial variations across model families and parameter sizes. 
These results suggest that the diversity of real-world table images and the breadth of our question set pose significant challenges for current models, particularly in accurately perceiving table structure and reasoning over cell values, indicating that robust table image understanding remains far from solved.

Our contributions are threefold:
\begin{enumerate}
    \item[1)] We introduce WildTableBench, the first table image understanding benchmark using table images from real-world settings. The benchmark comprises 402 images sourced from diverse scenarios, together with 928 human-crafted and verified questions spanning 17 subtypes across five categories, substantially bridging the gap between existing table evaluation settings and practical applications.
    \item[2)] We conduct a comprehensive evaluation of both frontier proprietary and open-source models on WildTableBench.
    Our study provides a systematic picture of current model capabilities on naturally occurring table images and reveals substantial performance gaps that remain in this setting.
    \item[3)] To better guide future research and development, we design diagnostic experiments that quantify the impact of reasoning difficulty and reasoning budget on model performance. 
    We also identify consistent performance declines on cells located in deeper rows or columns, revealing common failure patterns that call for improved visual grounding and reasoning.
\end{enumerate}

\section{Related Work}

\noindent{\textbf{Text-based table benchmarks.}}
The majority of existing table understanding benchmarks represent
tables as structured text. TabFact~\citep{chentabfact} scrapes Wikipedia
tables into CSV format for fact verification. TableBench~\citep{wu2025tablebench}
consolidates tables from WTQ~\citep{wtq}, SQA~\citep{sqa},
FinQA~\citep{finqa}, and related datasets to evaluate LLMs across
18 task subtypes. NeedleInATable~\citep{wang2025needleinatable} converts
tables from WTQ, TabFact, HiTab~\citep{cheng2022hitab}, and TABMWP~\citep{tabmwp}
into HTML and Markdown text formats to study long-context retrieval in LLMs.
TableEval~\citep{zhu2025tableeval} collects tables from real-world PDF
documents,
converting them into Markdown format for multilingual question answering.
While these benchmarks target sophisticated reasoning,
they provide models with explicit structural representations,
circumventing visual perception
and enabling models to exploit
dataset-level correlations without genuinely parsing table layout.

\noindent{\textbf{Rendered table image benchmarks.}}
To evaluate visual table understanding, several benchmarks programmatically convert structured table data into images.
TableVQA-Bench~\citep{kim2404tablevqa} produces images in two ways:
tables from WTQ and TabFact are rendered by attaching Wikipedia's
CSS stylesheet to their HTML source and capturing screenshots via
Puppeteer, while tables from FinTabNet~\citep{fintabnet} are cropped
from Fortune~500 annual report PDFs.
M\textsuperscript{2}-TabFact~\citep{zhou2025m2} extends TabFact
with a four-step pipeline that automatically renders Wikipedia
tables into visual form.
RealHiTBench~\citep{wu2025realhitbench} collects hierarchically structured tables from 13 open data platforms (\eg, Kaggle, OECD, government statistics portals)
in Excel format and programmatically converts them across three modalities: LaTeX, HTML, and PNG. 
Although these benchmarks extend evaluation beyond purely text-based table understanding,
they remain limited in two key respects. First, 
most visual inputs for these benchmarks are generated through controlled rendering or conversion pipelines, rather than being sourced from naturally occurring table images. Second, these pipelines reduce the variation in visual presentation, producing relatively clean and homogeneous inputs in typography, color, spacing, and formatting. As a result, they do not fully capture the visual complexity of real-world images, where screenshots, scans, and photos often contain compression artifacts, blur, perspective distortion, partial occlusion, background clutter, and source-specific styling variations.


\noindent{\textbf{Summary.}}
In summary, existing work either
operates on structured text without visual perception or uses
programmatically rendered images that remain clean and layout-regular. Since no existing benchmark directly collects table images from real-world visual sources, this paper builds WildTableBench to address this gap by sourcing images from web search engines, capturing the full spectrum of domain diversity, layout irregularity, visual noise, and structural complexity present in naturally occurring table images.

\begin{figure*}[t]
  \centering
    \begin{minipage}[t]{0.47\textwidth}
    \centering
    \adjustbox{width=\linewidth}{\includegraphics{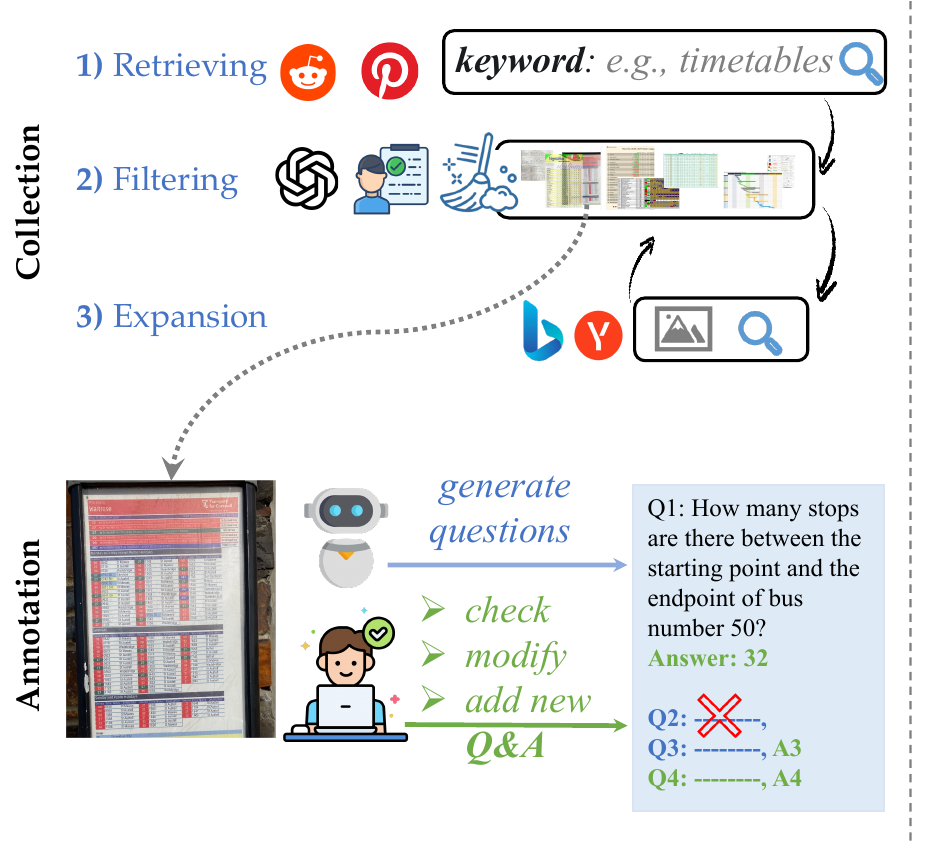}}%
    \par\vspace{2pt}
    {\small\textbf{(a)} Collection and annotation}
  \end{minipage}%
  \begin{minipage}[t]{0.50\textwidth}
    \centering
    \adjustbox{width=\linewidth}{\includegraphics{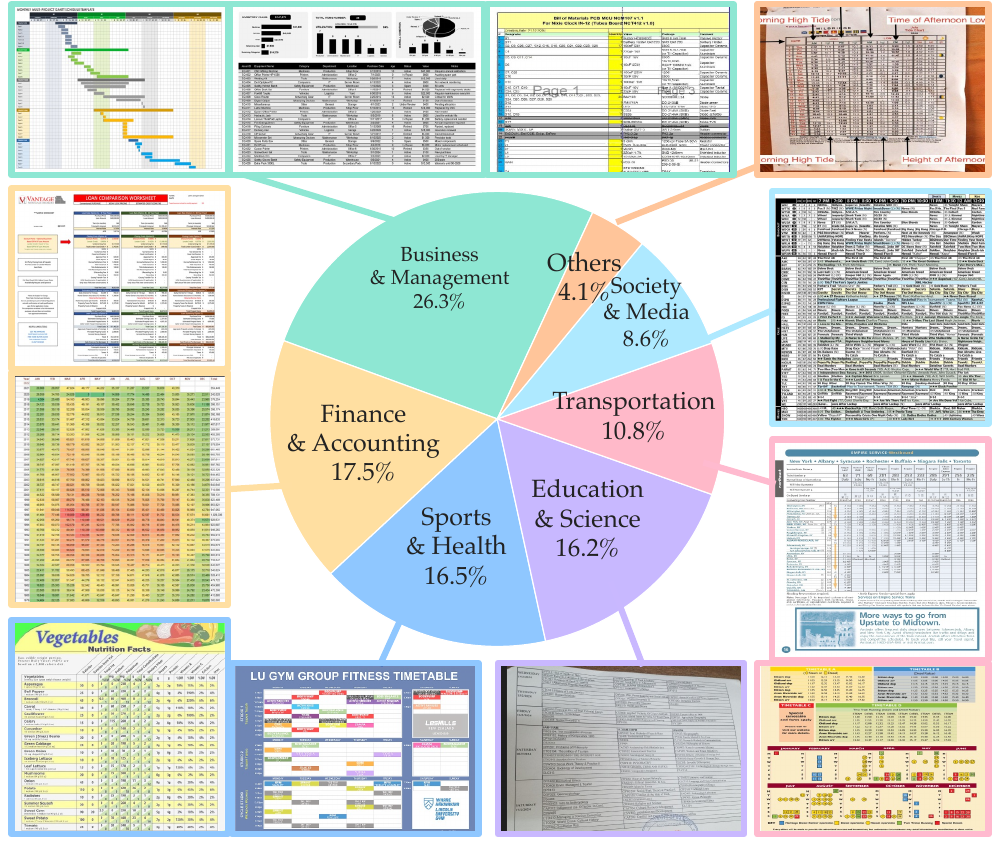}}%
    \par\vspace{2pt}
    {\small\textbf{(b)} Image domain and examples}
  \end{minipage}
  \vspace{-.3em}
  \caption{Overview of WildTableBench. \textbf{(a)} Data construction pipeline.
  \textbf{(b)} Domain distribution: WildTableBench covers diverse real-world scenarios, incorporating both high-fidelity digital screenshots and natural photographs across various professional and daily domains.}
  \label{fig:distribution}
\end{figure*}

\section{Benchmark Construction}
\label{sec:construction}
\subsection{Image Collection}
To ensure that WildTableBench reflects practical table-understanding scenarios, we collect real-world table images from publicly available web sources, where users share screenshots, scans, and photographs of tables. Rather than using artificially rendered tables, we design a collection process to ensure visual diversity and structural complexity in the collected images. Figure~\ref{fig:distribution}~(a) shows our pipeline. It has three stages: 

\textbf{Stage 1: Searching images by keywords.}
Candidate images are collected from platforms where users actively share table content, such as forums (e.g., Reddit) and visual discovery sites (e.g., Pinterest). We use a diverse set of table-related keywords, such as \textit{train timetable}, \textit{match schedule}, and \textit{nutrition label} (\autoref{app:keyword}), covering multiple domains.

\textbf{Stage 2: Quality filtering.}
Collected images are first automatically filtered using a GPT-4o-based~\citep{gpt4o2024} VLM method to remove non-tabular content or tables with insufficient structural complexity. Remaining images are manually reviewed to remove low-quality samples, and any personal information (e.g., names or addresses) is masked.

\textbf{Stage 3: Candidate expansion.}
The filtered images are used as queries on Bing and Yandex to expand the dataset with additional table images. Retrieved images are manually reviewed to ensure structural complexity, visual quality, diversity, and de-identification.


The final dataset comprises \textbf{402 table images} collected from publicly available web sources, covering a wide range of topics and scenarios across seven domain categories (Figure~\ref{fig:distribution}~(b); see Appendix~\ref{app:categories} for a full breakdown with fine-grained category labels).


\subsection{Task Taxonomy and Question Annotation}
\label{sec:taxonomy}

\begin{table*}[t]
\centering
\setlength{\extrarowheight}{1.2pt}
\small
\setlength{\tabcolsep}{5pt}
\caption{Taxonomy of 17 question subtypes in \textsc{WildTableBench}, organized into five categories.
\textbf{Reference} describes the table element the question targets;
\textbf{Output} describes the required answer type.}
\label{tab:taxonomy}
\resizebox{\textwidth}{!}{%
\begin{tabular}{>{\raggedright\arraybackslash}m{2.3cm} >{\raggedright\arraybackslash}m{2.8cm} >{\centering\arraybackslash}m{1.15cm} >{\raggedright\arraybackslash}m{3.4cm} >{\raggedright\arraybackslash}m{3.0cm} >{\raggedright\arraybackslash}m{5.2cm}}
\hline
\noalign{\vskip 3pt}
\textbf{Category} & \textbf{Subtype} & \textbf{ID} & \textbf{Reference} & \textbf{Output} & \textbf{Example Question} \\
\noalign{\vskip 3pt}
\hline
\rowcolor{g1row}& Transcription              & C1-T  & A column / row range                          & Ordered list of cell values                         & List column headers from B5 to H5 in left-to-right order. \\
\arrayrulecolor{gray!50}\cline{2-6}\arrayrulecolor{black}
\rowcolor{g1row}\tablerowpad& Cell Locating              & C1-L  & A cell address (e.g., E5)                     & Value at the given address                          & What is the value in cell G11? \\
\arrayrulecolor{gray!50}\cline{2-6}\arrayrulecolor{black}
\rowcolor{g1row}\tablerowpad& Semantic Lookup \& Struct. & C1-S  & A row label + column label / structural query & Intersection cell value or structural count         & What is the total sales for the ``Electronics'' row in the ``Q3'' column? \\
\arrayrulecolor{gray!50}\cline{2-6}\arrayrulecolor{black}
\rowcolor{g1row}\multirow{-7}{*}{\makecell[l]{\textcolor{g1col}{\small\textbf{C1}}\\[-0.1em]\textcolor{g1col}{\small\textbf{Cell-Level}}}}
& Excel Formula              & C1-F  & A cell range + target operation               & Excel formula string (e.g., \texttt{=SUM(U49:U52)}) & If ``2023'' is in cell F15, what formula calculates the total for that year? \\
\hline
\rowcolor{g2row}& Basic Numerical            & C2-B  & A column / row                                & Single operation (sum, difference, average, median) & What is the sum of ``Total Production'' for the first five countries listed? \\
\arrayrulecolor{gray!50}\cline{2-6}\arrayrulecolor{black}
\rowcolor{g2row}\tablerowpad& Ranking                   & C2-R  & A column + rank criterion                     & Max, min, or value/entity at the specified rank     & Which product has the 3rd highest revenue in the table? \\
\arrayrulecolor{gray!50}\cline{2-6}\arrayrulecolor{black}
\rowcolor{g2row}\tablerowpad& Conditional Numerical      & C2-C  & A filter condition + target column            & Aggregate over filtered rows                        & What percentage of Views came from External traffic on 4/10/2020? \\
\arrayrulecolor{gray!50}\cline{2-6}\arrayrulecolor{black}
\rowcolor{g2row}\multirow{-7}{*}{\makecell[l]{\textcolor{g2col}{\small\textbf{C2}}\\[-0.1em]\textcolor{g2col}{\small\textbf{Numerical}}}}
& Multi-step Conditional     & C2-M  & Multiple nested conditions                    & Multi-step numeric result                           & For tasks with ``Percent Complete'' $>75\%$, what is the total variance between Estimate and Actual? \\
\hline
\rowcolor{g3row}& Value Verification         & C3-V  & A factual claim about specific cell values    & True / False                                        & True or False: Total Revenue for 2026 is \$5,443,860 and the growth rate shows 43\% \\
\arrayrulecolor{gray!50}\cline{2-6}\arrayrulecolor{black}
\rowcolor{g3row}\tablerowpad& Aggregate Verification     & C3-A  & A numeric aggregate claim (sum, avg, count)   & True / False                                        & Are there exactly 10 occurrences of ``Tacna'' in the ``Ciudad Nac.'' range? \\
\arrayrulecolor{gray!50}\cline{2-6}\arrayrulecolor{black}
\rowcolor{g3row}\multirow{-5}{*}{\makecell[l]{\textcolor{g3col}{\small\textbf{C3}}\\[-0.1em]\textcolor{g3col}{\small\textbf{Verification}}}}
& Conditional Verification   & C3-C  & A conditional / filtered factual claim        & True / False                                        & True or False: The count of ``Argentina'' rows is less than the count of ``Australia'' rows \\
\hline
\rowcolor{g4row}& Row Operation              & C4-R  & A row addition / removal condition            & Recalculated result after row edit                  & If rows where ACTUAL exceeds BUDGET by $\geq$\$50 are removed, what is the new total? \\
\arrayrulecolor{gray!50}\cline{2-6}\arrayrulecolor{black}
\rowcolor{g4row}\tablerowpad& Value Modification         & C4-M  & A cell-value change specification             & Updated aggregate after the edit                    & If ``Expired'' status rows are changed to ``Used'', what is the new Used Count total? \\
\arrayrulecolor{gray!50}\cline{2-6}\arrayrulecolor{black}
\rowcolor{g4row}\multirow{-6}{*}{\makecell[l]{\textcolor{g4col}{\small\textbf{C4}}\\[-0.1em]\textcolor{g4col}{\small\textbf{Hypothetical}}}}
& Hypothetical Condition     & C4-H  & A hypothetical what-if scenario               & Derived numeric or Boolean result                   & If rows with MARK=`X' receive a 10\% pay increase, what is the new total pay? \\
\hline
\rowcolor{g5row}& Color Identification          & C5-I  & A cell, region, legend entry, or color-coded category & Color name or the category/status the color encodes & What is the background color of the ``CanFan 10'' row? \\
\arrayrulecolor{gray!50}\cline{2-6}\arrayrulecolor{black}
\rowcolor{g5row}\tablerowpad& Color-based Counting       & C5-C  & A target color or color-encoded category      & Count of matching cells or records                  & How many cells in the table contain white-colored text? \\
\arrayrulecolor{gray!50}\cline{2-6}\arrayrulecolor{black}
\rowcolor{g5row}\multirow{-5}{*}{\makecell[l]{\textcolor{g5col}{\small\textbf{C5}}\\[-0.1em]\textcolor{g5col}{\small\textbf{Color}}}}
& Color-based Reasoning      & C5-R  & A color filter plus an additional condition or operation & Conditional count or arithmetic result over color-selected subset & Sum all cells filled with a yellow background. \\
\hline
\end{tabular}}
\end{table*}

After collecting and filtering candidate images, we define a question taxonomy to guide annotation and evaluation.
WildTableBench organizes questions into a two-level taxonomy with five categories and 17 subtypes.
Table~\ref{tab:taxonomy} summarizes each subtype, including its identifier (e.g., C1-T for Cell-Level Transcription), the targeted table reference, the expected output type, and an example question. 

\noindent\textbf{Question type taxonomy.} The taxonomy spans a broad range of capabilities, from local cell perception and retrieval to multi-step reasoning. \textcolor{g1col}{C1: Cell-Level Understanding} covers direct transcription, cell localization, and label-based lookup of table content. \textcolor{g2col}{C2: Numerical Reasoning} requires arithmetic, ranking, and conditional aggregation over extracted values. \textcolor{g3col}{C3: Fact Verification} asks models to judge the truth of claims about cell values or aggregates, optionally under filtering conditions. \textcolor{g4col}{C4: Hypothetical Reasoning} evaluates counterfactual recomputation under explicit row- or value-level edits described in the question. \textcolor{g5col}{C5: Color-Based Reasoning} targets color-encoded cues, including color identification and computation over color-selected subsets.

\noindent\textbf{Annotation strategy.} All questions were annotated in English by five university students using a dedicated platform. For each candidate image, GPT-5.2~\citep{gpt522026} first generates 8 draft questions with prompts designed to cover taxonomy subtypes relevant to that image. These drafts serve only as annotation aids rather than final benchmark items. Following the annotation pipeline shown in Figure~\ref{fig:distribution}~(a), the annotators then review, revise, discard, and supplement the drafts to ensure that every retained Q\&A pair is unambiguous, image-grounded, and verifiable.

The final benchmark contains \textbf{928 questions} across 402 images, with 2.3 questions per image on average. More than 95\% of the questions were newly written or substantially revised by humans, and all reference answers were verified by humans to ensure quality and reliability.

\section{Experiments}
\label{sec:experiments}

\subsection{Experimental Setup}

\textbf{Models.} Using WildTableBench, we evaluate the table understanding capabilities of \textbf{21 models}, including both proprietary API-based and open-source models.


\textit{1) Proprietary models (9).}
These nine models are accessed through their official APIs and cover a range of model families and capability--cost trade-offs. They include \includegraphics[width=0.3cm,valign=c]{icon/openai.png} OpenAI: GPT-o3~\citep{o32024}, GPT-4o~\citep{gpt4o2024}, GPT-5-mini~\citep{gpt5mini2026}, and GPT-5.2~\citep{gpt522026}
; \includegraphics[width=0.3cm,valign=c]{icon/gemini.png} Google: Gemini~3~Pro~\citep{gemini3pro2026} and
Gemini~3~Flash~\citep{gemini3flash2026}; \includegraphics[width=0.3cm,valign=c]{icon/claude.png} Anthropic: Claude~Sonnet~4.6~\citep{claudesonnet462026} and Claude~Opus~4.6~\citep{claudeopus462026}; and \includegraphics[width=0.3cm,valign=c]{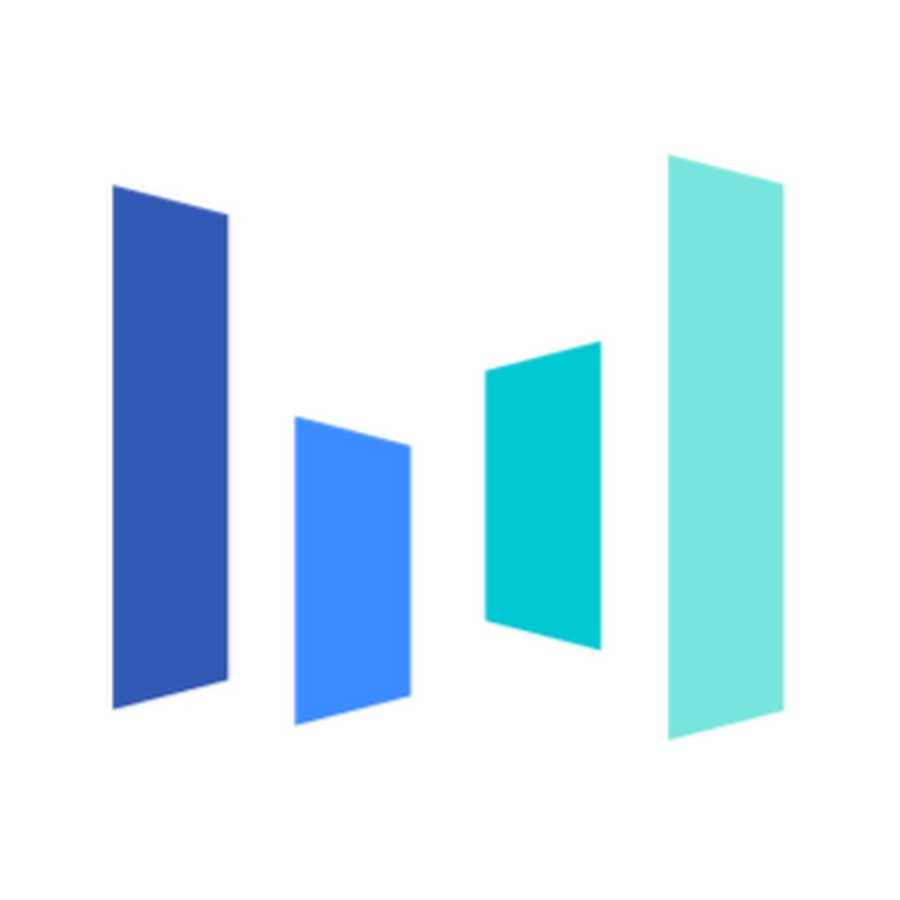} Bytedance: Seed-2.0-Pro~\citep{seed2026}. For models that support configurable reasoning, we use high reasoning effort.

\textit{2) Open-source models (12).} These models come from three families. \includegraphics[width=0.3cm,valign=c]{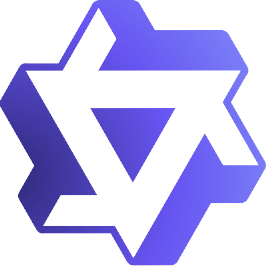} Qwen3-VL~\citep{qwen3vl2025} is a series of vision-language models available
in five parameter scales: 2B, 4B, 8B, 32B, and 235B.
For each scale, we evaluate both the standard instruction-tuned variant
(\textit{Instruct}) and the chain-of-thought reasoning variant
(\textit{Thinking}), yielding ten models in total.
\includegraphics[width=0.3cm,valign=c]{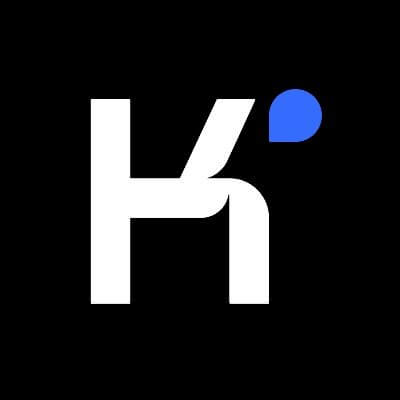} Kimi~K2.5~\citep{team2026kimi} is a 1-trillion parameter Mixture-of-Experts multimodal model with open weights released under a
Modified MIT License. \includegraphics[width=0.3cm,valign=c]{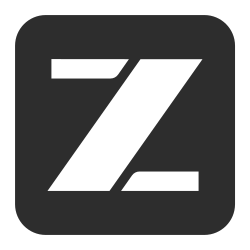} GLM-4.6V~\citep{glm46v2025} is a 106B-parameter MoE vision-language model with open weights supporting 128K context. 
Full implementation details are provided in Appendix~\ref{app:impl}.


\noindent{\textbf{Task format.}} All questions in WildTableBench are answered in a free-form format. We avoid multiple-choice options to prevent answer-selection shortcuts and to better reflect practical table understanding scenarios. Each model is prompted to analyze the table image step by step and end
its response with ``Final answer: [answer]''. We extract the prediction from the last occurrence of this pattern. The full inference prompt is provided in Appendix~\ref{app:response_prompt}.


\noindent{\textbf{Evaluation protocol: LLM-as-judge.}}
Following prior work~\citep{zheng2023judging,chen2024mllm}, we use GPT-5.2 as an LLM judge. It receives the question, reference answer, final answer, and full model response, applying format-insensitive matching rules (Appendix~\ref{app:general_prompt} --~\ref{app:transcription_prompt}) for numerical, text, Boolean, and list answers. We validate reliability through 4 random audits of 50 responses each, finding 96\%${\pm2\%}$ agreement with human judgments.



\subsection{Main Results}  

Table~\ref{tab:main_results} reports accuracy for all evaluated models across subtypes and categories. Several findings stand out:

\begin{table*}[t]
  \centering
  \tiny
  \caption{Accuracy (\%) on \textsc{WildTableBench} by subtype and by category (see Table~\ref{tab:taxonomy}). Avg columns average subtypes within each category. \textbf{Bold} = column best.}
  \label{tab:main_results}
  \label{tab:subcategory_results}
  \setlength{\tabcolsep}{2.0pt}{\begin{tabular}{l|cccc|c|cccc|c|ccc|c|ccc|c|ccc|c|c}
  \hline
  \noalign{\vskip 3pt}
  \multirow{2}{*}{\textbf{Model}} &
  \multicolumn{5}{c|}{\textbf{C1: Cell-Level}} &
  \multicolumn{5}{c|}{\textbf{C2: Numerical}} &
  \multicolumn{4}{c|}{\textbf{C3: Verification}} &
  \multicolumn{4}{c|}{\textbf{C4: Hypo.}} &
  \multicolumn{4}{c|}{\textbf{C5: Color}} &
  \textbf{Ovr.} \\
  & \rotatebox{90}{C1-T} & \rotatebox{90}{C1-L} & \rotatebox{90}{C1-S} & \rotatebox{90}{C1-F} & \rotatebox{90}{\textbf{Avg}} &
  \rotatebox{90}{C2-B} & \rotatebox{90}{C2-R} & \rotatebox{90}{C2-C} & \rotatebox{90}{C2-M} & \rotatebox{90}{\textbf{Avg}} &
  \rotatebox{90}{C3-V} & \rotatebox{90}{C3-A} & \rotatebox{90}{C3-C} & \rotatebox{90}{\textbf{Avg}} &
  \rotatebox{90}{C4-R} & \rotatebox{90}{C4-M} & \rotatebox{90}{C4-H} & \rotatebox{90}{\textbf{Avg}} &
  \rotatebox{90}{C5-I} & \rotatebox{90}{C5-C} & \rotatebox{90}{C5-R} & \rotatebox{90}{\textbf{Avg}} &
  \rotatebox{90}{\textbf{Overall}} \\
  \noalign{\vskip 3pt}
  \hline
  \noalign{\vskip 1pt}
  \multicolumn{24}{l}{\textbf{Proprietary LMMs}} \\
  \noalign{\vskip 1pt}
  \hline
  Gemini-3-Pro & \rgb{66.7}\textbf{66.7} & \rgb{71.4}\textbf{71.4} & \rgb{46.7}46.7 & \rgb{61.1}\textbf{61.1} & \textbf{62.8} & \rgb{77.1}\textbf{77.1} & \rgb{80.0}\textbf{80.0} & \rgb{74.1}\textbf{74.1} & \rgb{66.4}\textbf{66.4} & \textbf{71.9} & \rgb{73.9}\textbf{73.9} & \rgb{54.5}54.5 & \rgb{75.8}\textbf{75.8} & \textbf{71.6} & \rgb{77.5}\textbf{77.5} & \rgb{61.9}61.9 & \rgb{82.1}\textbf{82.1} & \textbf{75.3} & \rgb{55.6}55.6 & \rgb{62.5}\textbf{62.5} & \rgb{51.6}\textbf{51.6} & \textbf{55.8} & \rgb{67.9}\textbf{67.9} \\
  Gemini-3-Flash & \rgb{43.3}43.3 & \rgb{61.2}61.2 & \rgb{50.0}50.0 & \rgb{44.4}44.4 & 51.0 & \rgb{37.1}37.1 & \rgb{42.5}42.5 & \rgb{54.3}54.3 & \rgb{51.0}51.0 & 49.7 & \rgb{43.5}43.5 & \rgb{54.5}54.5 & \rgb{69.7}69.7 & 58.2 & \rgb{57.5}57.5 & \rgb{57.1}57.1 & \rgb{50.0}50.0 & 55.1 & \rgb{50.0}50.0 & \rgb{37.5}37.5 & \rgb{33.9}33.9 & 37.5 & \rgb{49.4}49.4 \\
  Seed-2.0-Pro & \rgb{50.0}50.0 & \rgb{67.3}67.3 & \rgb{56.7}\textbf{56.7} & \rgb{58.3}58.3 & 59.3 & \rgb{40.0}40.0 & \rgb{37.5}37.5 & \rgb{52.6}52.6 & \rgb{42.0}42.0 & 44.9 & \rgb{73.9}\textbf{73.9} & \rgb{63.6}63.6 & \rgb{66.7}66.7 & 68.7 & \rgb{47.5}47.5 & \rgb{42.9}42.9 & \rgb{42.9}42.9 & 44.9 & \rgb{55.6}55.6 & \rgb{37.5}37.5 & \rgb{22.6}22.6 & 32.5 & \rgb{47.8}47.8 \\
  GPT-5.2 & \rgb{43.3}43.3 & \rgb{65.3}65.3 & \rgb{46.7}46.7 & \rgb{50.0}50.0 & 53.1 & \rgb{45.7}45.7 & \rgb{45.0}45.0 & \rgb{42.2}42.2 & \rgb{49.0}49.0 & 45.8 & \rgb{52.2}52.2 & \rgb{45.5}45.5 & \rgb{63.6}63.6 & 56.7 & \rgb{45.0}45.0 & \rgb{66.7}\textbf{66.7} & \rgb{60.7}60.7 & 55.1 & \rgb{50.0}50.0 & \rgb{30.0}30.0 & \rgb{22.6}22.6 & 29.2 & \rgb{46.6}46.6 \\
  Claude-Opus-4.6 & \rgb{46.7}46.7 & \rgb{65.3}65.3 & \rgb{33.3}33.3 & \rgb{52.8}52.8 & 51.7 & \rgb{54.3}54.3 & \rgb{42.5}42.5 & \rgb{47.4}47.4 & \rgb{45.5}45.5 & 46.7 & \rgb{73.9}\textbf{73.9} & \rgb{45.5}45.5 & \rgb{57.6}57.6 & 61.2 & \rgb{67.5}67.5 & \rgb{33.3}33.3 & \rgb{50.0}50.0 & 53.9 & \rgb{44.4}44.4 & \rgb{25.0}25.0 & \rgb{11.3}11.3 & 20.8 & \rgb{45.7}45.7 \\
  Claude-Sonnet-4.6 & \rgb{20.0}20.0 & \rgb{40.8}40.8 & \rgb{46.7}46.7 & \rgb{33.3}33.3 & 35.9 & \rgb{34.3}34.3 & \rgb{25.0}25.0 & \rgb{37.1}37.1 & \rgb{34.3}34.3 & 34.1 & \rgb{52.2}52.2 & \rgb{36.4}36.4 & \rgb{51.5}51.5 & 49.3 & \rgb{50.0}50.0 & \rgb{42.9}42.9 & \rgb{53.6}53.6 & 49.4 & \rgb{33.3}33.3 & \rgb{17.5}17.5 & \rgb{12.9}12.9 & 17.5 & \rgb{35.0}35.0 \\
  GPT-5-mini & \rgb{23.3}23.3 & \rgb{24.5}24.5 & \rgb{20.0}20.0 & \rgb{13.9}13.9 & 20.7 & \rgb{28.6}28.6 & \rgb{20.0}20.0 & \rgb{27.6}27.6 & \rgb{21.7}21.7 & 24.3 & \rgb{52.2}52.2 & \rgb{27.3}27.3 & \rgb{39.4}39.4 & 41.8 & \rgb{37.5}37.5 & \rgb{28.6}28.6 & \rgb{32.1}32.1 & 33.7 & \rgb{44.4}44.4 & \rgb{15.0}15.0 & \rgb{12.9}12.9 & 18.3 & \rgb{25.3}25.3 \\
  GPT-o3 & \rgb{10.0}10.0 & \rgb{16.3}16.3 & \rgb{20.0}20.0 & \rgb{22.2}22.2 & 17.2 & \rgb{14.3}14.3 & \rgb{5.0}5.0 & \rgb{9.5}9.5 & \rgb{14.0}14.0 & 11.4 & \rgb{47.8}47.8 & \rgb{18.2}18.2 & \rgb{27.3}27.3 & 32.8 & \rgb{10.0}10.0 & \rgb{9.5}9.5 & \rgb{10.7}10.7 & 10.1 & \rgb{27.8}27.8 & \rgb{10.0}10.0 & \rgb{14.5}14.5 & 15.0 & \rgb{14.8}14.8 \\
  GPT-4o & \rgb{6.7}6.7 & \rgb{4.1}4.1 & \rgb{0.0}0.0 & \rgb{5.6}5.6 & 4.1 & \rgb{2.9}2.9 & \rgb{0.0}0.0 & \rgb{3.4}3.4 & \rgb{5.6}5.6 & 3.9 & \rgb{17.4}17.4 & \rgb{9.1}9.1 & \rgb{33.3}33.3 & 23.9 & \rgb{0.0}0.0 & \rgb{0.0}0.0 & \rgb{0.0}0.0 & 0.0 & \rgb{16.7}16.7 & \rgb{5.0}5.0 & \rgb{4.8}4.8 & 6.7 & \rgb{5.7}5.7 \\
  \hline
  \noalign{\vskip 1pt}
  \multicolumn{24}{l}{\textbf{Open-source LMMs: Thinking}} \\
  \noalign{\vskip 1pt}
  \hline
  Kimi-K2.5 & \rgb{46.7}46.7 & \rgb{51.0}51.0 & \rgb{56.7}\textbf{56.7} & \rgb{36.1}36.1 & 47.6 & \rgb{51.4}51.4 & \rgb{62.5}62.5 & \rgb{52.6}52.6 & \rgb{45.5}45.5 & 50.6 & \rgb{56.5}56.5 & \rgb{81.8}\textbf{81.8} & \rgb{69.7}69.7 & 67.2 & \rgb{57.5}57.5 & \rgb{52.4}52.4 & \rgb{53.6}53.6 & 55.1 & \rgb{66.7}\textbf{66.7} & \rgb{27.5}27.5 & \rgb{35.5}35.5 & 37.5 & \rgb{49.9}49.9 \\
  Qwen3-VL-235B-T & \rgb{26.7}26.7 & \rgb{46.9}46.9 & \rgb{33.3}33.3 & \rgb{38.9}38.9 & 37.9 & \rgb{37.1}37.1 & \rgb{22.5}22.5 & \rgb{37.9}37.9 & \rgb{32.2}32.2 & 33.5 & \rgb{26.1}26.1 & \rgb{36.4}36.4 & \rgb{54.5}54.5 & 41.8 & \rgb{30.0}30.0 & \rgb{42.9}42.9 & \rgb{50.0}50.0 & 39.3 & \rgb{55.6}55.6 & \rgb{22.5}22.5 & \rgb{21.0}21.0 & 26.7 & \rgb{34.7}34.7 \\
  Qwen3-VL-32B-T & \rgb{13.3}13.3 & \rgb{28.6}28.6 & \rgb{46.7}46.7 & \rgb{25.0}25.0 & 28.3 & \rgb{37.1}37.1 & \rgb{25.0}25.0 & \rgb{31.0}31.0 & \rgb{23.8}23.8 & 27.8 & \rgb{39.1}39.1 & \rgb{18.2}18.2 & \rgb{48.5}48.5 & 40.3 & \rgb{35.0}35.0 & \rgb{4.8}4.8 & \rgb{10.7}10.7 & 20.2 & \rgb{55.6}55.6 & \rgb{25.0}25.0 & \rgb{22.6}22.6 & 28.3 & \rgb{28.2}28.2 \\
  GLM-4.6V & \rgb{26.7}26.7 & \rgb{20.4}20.4 & \rgb{10.0}10.0 & \rgb{13.9}13.9 & 17.9 & \rgb{25.7}25.7 & \rgb{10.0}10.0 & \rgb{28.4}28.4 & \rgb{23.8}23.8 & 24.0 & \rgb{17.4}17.4 & \rgb{27.3}27.3 & \rgb{48.5}48.5 & 34.3 & \rgb{35.0}35.0 & \rgb{23.8}23.8 & \rgb{17.9}17.9 & 27.0 & \rgb{66.7}\textbf{66.7} & \rgb{17.5}17.5 & \rgb{19.4}19.4 & 25.8 & \rgb{24.4}24.4 \\
  Qwen3-VL-8B-T & \rgb{10.0}10.0 & \rgb{14.3}14.3 & \rgb{3.3}3.3 & \rgb{5.6}5.6 & 9.0 & \rgb{28.6}28.6 & \rgb{10.0}10.0 & \rgb{20.7}20.7 & \rgb{8.4}8.4 & 15.0 & \rgb{34.8}34.8 & \rgb{18.2}18.2 & \rgb{27.3}27.3 & 28.4 & \rgb{20.0}20.0 & \rgb{0.0}0.0 & \rgb{25.0}25.0 & 16.9 & \rgb{33.3}33.3 & \rgb{10.0}10.0 & \rgb{6.5}6.5 & 11.7 & \rgb{14.7}14.7 \\
  Qwen3-VL-4B-T & \rgb{6.7}6.7 & \rgb{8.2}8.2 & \rgb{0.0}0.0 & \rgb{2.8}2.8 & 4.8 & \rgb{5.7}5.7 & \rgb{2.5}2.5 & \rgb{8.6}8.6 & \rgb{3.5}3.5 & 5.4 & \rgb{26.1}26.1 & \rgb{9.1}9.1 & \rgb{30.3}30.3 & 25.4 & \rgb{10.0}10.0 & \rgb{0.0}0.0 & \rgb{3.6}3.6 & 5.6 & \rgb{16.7}16.7 & \rgb{12.5}12.5 & \rgb{4.8}4.8 & 9.2 & \rgb{7.7}7.7 \\
  Qwen3-VL-2B-T & \rgb{6.7}6.7 & \rgb{6.1}6.1 & \rgb{0.0}0.0 & \rgb{2.8}2.8 & 4.1 & \rgb{0.0}0.0 & \rgb{0.0}0.0 & \rgb{4.3}4.3 & \rgb{2.1}2.1 & 2.4 & \rgb{13.0}13.0 & \rgb{18.2}18.2 & \rgb{15.2}15.2 & 14.9 & \rgb{2.5}2.5 & \rgb{0.0}0.0 & \rgb{0.0}0.0 & 1.1 & \rgb{22.2}22.2 & \rgb{2.5}2.5 & \rgb{1.6}1.6 & 5.0 & \rgb{4.1}4.1 \\
  \hline
  \noalign{\vskip 1pt}
  \multicolumn{24}{l}{\textbf{Open-source LMMs: Instruct}} \\
  \noalign{\vskip 1pt}
  \hline
  Qwen3-VL-235B-I & \rgb{23.3}23.3 & \rgb{28.6}28.6 & \rgb{23.3}23.3 & \rgb{13.9}13.9 & 22.8 & \rgb{28.6}28.6 & \rgb{20.0}20.0 & \rgb{30.2}30.2 & \rgb{25.2}25.2 & 26.6 & \rgb{30.4}30.4 & \rgb{36.4}36.4 & \rgb{36.4}36.4 & 34.3 & \rgb{25.0}25.0 & \rgb{23.8}23.8 & \rgb{14.3}14.3 & 21.3 & \rgb{66.7}\textbf{66.7} & \rgb{15.0}15.0 & \rgb{12.9}12.9 & 21.7 & \rgb{25.2}25.2 \\
  Qwen3-VL-32B-I & \rgb{16.7}16.7 & \rgb{30.6}30.6 & \rgb{23.3}23.3 & \rgb{25.0}25.0 & 24.8 & \rgb{17.1}17.1 & \rgb{20.0}20.0 & \rgb{23.3}23.3 & \rgb{23.8}23.8 & 22.5 & \rgb{43.5}43.5 & \rgb{36.4}36.4 & \rgb{42.4}42.4 & 41.8 & \rgb{27.5}27.5 & \rgb{33.3}33.3 & \rgb{21.4}21.4 & 27.0 & \rgb{44.4}44.4 & \rgb{20.0}20.0 & \rgb{9.7}9.7 & 18.3 & \rgb{24.5}24.5 \\
  Qwen3-VL-8B-I & \rgb{10.0}10.0 & \rgb{16.3}16.3 & \rgb{3.3}3.3 & \rgb{8.3}8.3 & 10.3 & \rgb{8.6}8.6 & \rgb{10.0}10.0 & \rgb{11.2}11.2 & \rgb{6.3}6.3 & 8.7 & \rgb{17.4}17.4 & \rgb{18.2}18.2 & \rgb{18.2}18.2 & 17.9 & \rgb{15.0}15.0 & \rgb{0.0}0.0 & \rgb{7.1}7.1 & 9.0 & \rgb{44.4}44.4 & \rgb{5.0}5.0 & \rgb{1.6}1.6 & 9.2 & \rgb{9.9}9.9 \\
  Qwen3-VL-4B-I & \rgb{10.0}10.0 & \rgb{10.2}10.2 & \rgb{3.3}3.3 & \rgb{8.3}8.3 & 8.3 & \rgb{2.9}2.9 & \rgb{2.5}2.5 & \rgb{6.9}6.9 & \rgb{5.6}5.6 & 5.4 & \rgb{26.1}26.1 & \rgb{9.1}9.1 & \rgb{12.1}12.1 & 16.4 & \rgb{10.0}10.0 & \rgb{4.8}4.8 & \rgb{3.6}3.6 & 6.7 & \rgb{38.9}38.9 & \rgb{7.5}7.5 & \rgb{4.8}4.8 & 10.8 & \rgb{7.9}7.9 \\
  Qwen3-VL-2B-I & \rgb{3.3}3.3 & \rgb{6.1}6.1 & \rgb{6.7}6.7 & \rgb{0.0}0.0 & 4.1 & \rgb{2.9}2.9 & \rgb{0.0}0.0 & \rgb{3.4}3.4 & \rgb{4.2}4.2 & 3.3 & \rgb{21.7}21.7 & \rgb{18.2}18.2 & \rgb{24.2}24.2 & 22.4 & \rgb{2.5}2.5 & \rgb{0.0}0.0 & \rgb{3.6}3.6 & 2.2 & \rgb{38.9}38.9 & \rgb{2.5}2.5 & \rgb{3.2}3.2 & 8.3 & \rgb{5.8}5.8 \\
  \hline
  \end{tabular} }
  \end{table*}

\noindent{\textbf{WildTableBench is challenging even for frontier proprietary models.}} In terms of overall accuracy, Gemini-3-Pro achieves the best score at only 67.9\%, leaving nearly one-third of the questions incorrect. Within the proprietary tier, performance varies widely, and most evaluated models remain below 50\% overall: while Claude-Opus-4.6 (45.7\%), Gemini-3-Flash (49.4\%), and GPT-5.2 (46.6\%) approach the 50\% level, GPT-5-mini, GPT-o3, and GPT-4o fall much lower at 25.3\%, 14.8\%, and 5.7\%, respectively; GPT-4o also records 0.0\% on several subtypes. Overall, these results show that WildTableBench is difficult even for current frontier proprietary models, with substantial capability variation even within this tier.

\noindent{\textbf{Proprietary frontier models generally outperform open-source alternatives.}}
Among open-source models, Kimi-K2.5 achieves the highest overall accuracy at 49.9\%, followed by Qwen3-VL-235B-T at 34.7\%. Both remain well below Gemini-3-Pro, trailing it by 18.0\% and 33.2\%, respectively. The remaining open-source models perform substantially worse, consistent with the predominance of low-accuracy cells (\swatch{heat33}\,\swatch{heat17}\,\swatch{heat08}\,\swatch{heat00}, corresponding roughly to 30--0\%) in the lower portion of Table~\ref{tab:main_results}.
This gap highlights the difficulty of real-world table understanding for current open-source models.


\noindent{\textbf{Model performance varies substantially across question categories.}}
C3 (Verification question) is often among the highest-accuracy categories across models. Gemini-3-Pro, Seed-2.0-Pro, and Kimi-K2.5 reach 71.6\%, 68.7\%, and 67.2\% on C3, respectively, and even GPT-5-mini scores 41.8\%, higher than its C1, C2, and C5 results. In contrast, C5 (Color) is frequently among the weakest categories: GPT-5.2 scores 29.2\%, Claude-Opus-4.6 20.8\%, and Claude-Sonnet-4.6 17.5\%, while even Gemini-3-Pro reaches only 55.8\%. For C5, a closer look suggests that the challenge lies less in recognizing color itself than in using color for reasoning. Kimi-K2.5 drops from 66.7\% on C5-I to 27.5\% on C5-C and 35.5\% on C5-R, and GLM-4.6V drops from 66.7\% on C5-I to 17.5\% on C5-C and 19.4\% on C5-R. This suggests that the difficulty of WildTableBench lies not only in local perception, but also in compositional reasoning over grounded table evidence.

\noindent{\textbf{Larger reasoning models perform better.}}
Within the Qwen3-VL Thinking family, overall accuracy increases from 4.1\% (2B) to 7.7\% (4B), 14.7\% (8B), and 34.7\% (235B). A similar trend holds for the Instruct family, where performance rises from 5.8\% to 7.9\%, 9.9\%, and 25.2\% as model size increases. The advantage of Thinking over Instruct is small at 2B (4.1\% vs.\ 5.8\%), but becomes clear at 8B (14.7\% vs.\ 9.9\%) and 235B (34.7\% vs.\ 25.2\%), suggesting that explicit reasoning is more beneficial at \textit{larger scales}. Across question categories, this advantage is especially clear in structured reasoning tasks. For example, Qwen3-VL-235B-T outperforms Qwen3-VL-235B-I on C1 (37.9\% vs.\ 22.8\%), C2 (33.5\% vs.\ 26.6\%), and C4 (39.3\% vs.\ 21.3\%).

Overall, WildTableBench provides a fine-grained view of table understanding ability.
Beyond ranking models by overall accuracy, it reveals where they succeed and fail across different skills, including verification, numerical reasoning, hypothetical reasoning, and color-based reasoning. This makes the benchmark useful not only for comparing frontier and open-source models, but also for diagnosing their distinct strengths and weaknesses. More broadly, the results suggest that table understanding is not a single ability, but a combination of perception, grounding, and multi-step reasoning skills.

\subsection{Analysis and Discussion}

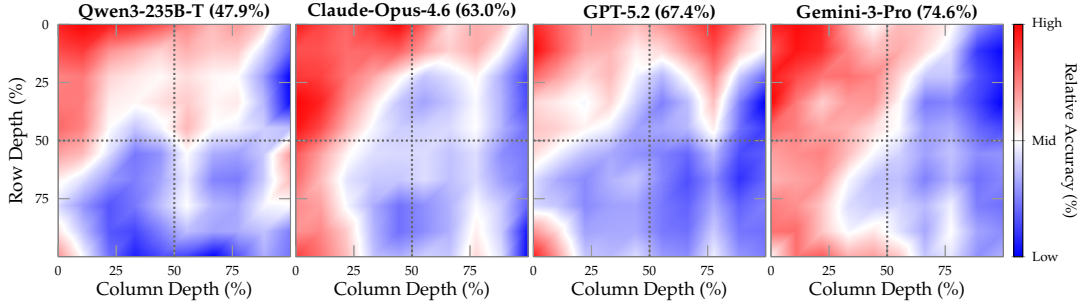
\begin{figure*}[t]
    \centering
  \pgfplotsset{
    colormap={RdBu}{
      rgb255(0)=(0,0,255)
      rgb255(250)=(128,128,255)
      rgb255(500)=(255,255,255)
      rgb255(750)=(255,128,128)
      rgb255(1000)=(255,0,0)
    },
    /pgfplots/heatmap base/.style={
      view={0}{90},
      width=0.22\textwidth,
      height=0.22\textwidth,
      colormap name=RdBu,
      point meta min=0, point meta max=1,
      y dir=reverse,
      xmin=0, xmax=9, ymin=0, ymax=9,
      enlargelimits=false,
      xtick={0,2.25,4.5,6.75},
      xticklabels={0,25,50,75},
      ytick={0,2.25,4.5,6.75},
      yticklabels={0,25,50,75},
      ztick=\empty, zlabel={},
      tick label style={font=\fontsize{5}{6}\selectfont},
      title style={font=\scriptsize\bfseries, at={(0.5,1.0)}, anchor=south, yshift=-8pt, align=center},
      xlabel style={font=\scriptsize, at={(0.5,0)}, anchor=north, yshift=-6pt},
      ylabel style={font=\scriptsize, at={(0,0.5)}, anchor=south, yshift=8pt,xshift=0pt},
      axis on top,
      axis line style={thin, black!50},
      tick style={thin, black!40},
      scale only axis,
      3d box=complete,
      after end axis/.code={
        \draw[densely dotted, black!60, line width=0.8pt]
          (rel axis cs:0.5,0) -- (rel axis cs:0.5,1);
        \draw[densely dotted, black!60, line width=0.8pt]
          (rel axis cs:0,0.5) -- (rel axis cs:1,0.5);
      },
    }
  }
  
  \begin{tikzpicture}

  \begin{axis}[
    heatmap base,
    name=plotA,
    at={(0,0)}, anchor=north west,
    title={Qwen3-235B-T (47.9\%)},
    xlabel={Column Depth (\%)}, ylabel={Row Depth (\%)},
  ]
  \addplot3[surf, shader=interp, mesh/cols=10] table[meta=C] {
  x y C
  0 0 0.941
  1 0 1.000
  2 0 0.945
  3 0 0.920
  4 0 0.843
  5 0 0.749
  6 0 0.640
  7 0 0.703
  8 0 0.585
  9 0 0.269
  0 1 0.861
  1 1 0.895
  2 1 0.724
  3 1 0.668
  4 1 0.594
  5 1 0.663
  6 1 0.606
  7 1 0.612
  8 1 0.424
  9 1 0.193
  0 2 0.748
  1 2 0.767
  2 2 0.590
  3 2 0.549
  4 2 0.515
  5 2 0.569
  6 2 0.519
  7 2 0.505
  8 2 0.306
  9 2 0.001
  0 3 0.766
  1 3 0.761
  2 3 0.544
  3 3 0.520
  4 3 0.568
  5 3 0.609
  6 3 0.524
  7 3 0.547
  8 3 0.362
  9 3 0.018
  0 4 0.802
  1 4 0.748
  2 4 0.506
  3 4 0.388
  4 4 0.482
  5 4 0.651
  6 4 0.499
  7 4 0.472
  8 4 0.409
  9 4 0.376
  0 5 0.691
  1 5 0.639
  2 5 0.438
  3 5 0.235
  4 5 0.290
  5 5 0.502
  6 5 0.340
  7 5 0.313
  8 5 0.401
  9 5 0.700
  0 6 0.535
  1 6 0.437
  2 6 0.299
  3 6 0.250
  4 6 0.323
  5 6 0.464
  6 6 0.260
  7 6 0.243
  8 6 0.401
  9 6 0.636
  0 7 0.434
  1 7 0.295
  2 7 0.166
  3 7 0.235
  4 7 0.376
  5 7 0.504
  6 7 0.358
  7 7 0.328
  8 7 0.482
  9 7 0.496
  0 8 0.561
  1 8 0.380
  2 8 0.172
  3 8 0.137
  4 8 0.255
  5 8 0.376
  6 8 0.323
  7 8 0.335
  8 8 0.384
  9 8 0.250
  0 9 0.728
  1 9 0.494
  2 9 0.219
  3 9 0.057
  4 9 0.148
  5 9 0.125
  6 9 0.000
  7 9 0.174
  8 9 0.344
  9 9 0.226
  };
  \end{axis}

  \begin{axis}[
    heatmap base,
    name=plotB,
    at={(plotA.north east)}, anchor=north west, xshift=2pt,
    title={Claude-Opus-4.6 (63.0\%)},
    xlabel={Column Depth (\%)}, yticklabels={},
  ]
  \addplot3[surf, shader=interp, mesh/cols=10] table[meta=C] {
  x y C
  0 0 0.962
  1 0 0.873
  2 0 0.806
  3 0 0.904
  4 0 0.989
  5 0 0.812
  6 0 0.567
  7 0 0.669
  8 0 0.623
  9 0 0.395
  0 1 0.850
  1 1 0.845
  2 1 0.794
  3 1 0.805
  4 1 0.824
  5 1 0.684
  6 1 0.594
  7 1 0.672
  8 1 0.502
  9 1 0.251
  0 2 0.894
  1 2 0.881
  2 2 0.777
  3 2 0.678
  4 2 0.532
  5 2 0.391
  6 2 0.443
  7 2 0.526
  8 2 0.338
  9 2 0.124
  0 3 1.000
  1 3 0.939
  2 3 0.731
  3 3 0.558
  4 3 0.391
  5 3 0.320
  6 3 0.391
  7 3 0.494
  8 3 0.346
  9 3 0.146
  0 4 0.976
  1 4 0.862
  2 4 0.649
  3 4 0.486
  4 4 0.406
  5 4 0.413
  6 4 0.431
  7 4 0.505
  8 4 0.350
  9 4 0.221
  0 5 0.842
  1 5 0.685
  2 5 0.521
  3 5 0.440
  4 5 0.428
  5 5 0.436
  6 5 0.383
  7 5 0.439
  8 5 0.279
  9 5 0.235
  0 6 0.805
  1 6 0.639
  2 6 0.450
  3 6 0.397
  4 6 0.394
  5 6 0.437
  6 6 0.378
  7 6 0.436
  8 6 0.310
  9 6 0.233
  0 7 0.736
  1 7 0.692
  2 7 0.489
  3 7 0.343
  4 7 0.223
  5 7 0.282
  6 7 0.351
  7 7 0.494
  8 7 0.396
  9 7 0.187
  0 8 0.719
  1 8 0.743
  2 8 0.601
  3 8 0.360
  4 8 0.220
  5 8 0.277
  6 8 0.348
  7 8 0.538
  8 8 0.398
  9 8 0.029
  0 9 0.880
  1 9 0.779
  2 9 0.634
  3 9 0.389
  4 9 0.343
  5 9 0.412
  6 9 0.391
  7 9 0.582
  8 9 0.438
  9 9 0.000
  };
  \end{axis}

  \begin{axis}[
    heatmap base,
    name=plotC,
    at={(plotB.north east)}, anchor=north west, xshift=2pt,
    title={GPT-5.2 (67.4\%)},
    xlabel={Column Depth (\%)}, yticklabels={},
  ]
  \addplot3[surf, shader=interp, mesh/cols=10] table[meta=C] {
  x y C
  0 0 0.938
  1 0 0.815
  2 0 0.695
  3 0 0.711
  4 0 0.630
  5 0 0.712
  6 0 0.820
  7 0 0.933
  8 0 0.753
  9 0 0.523
  0 1 1.000
  1 1 0.829
  2 1 0.675
  3 1 0.653
  4 1 0.532
  5 1 0.576
  6 1 0.712
  7 1 0.854
  8 1 0.619
  9 1 0.487
  0 2 0.811
  1 2 0.689
  2 2 0.575
  3 2 0.644
  4 2 0.478
  5 2 0.357
  6 2 0.522
  7 2 0.739
  8 2 0.405
  9 2 0.209
  0 3 0.587
  1 3 0.538
  2 3 0.489
  3 3 0.588
  4 3 0.408
  5 3 0.250
  6 3 0.372
  7 3 0.631
  8 3 0.256
  9 3 0.000
  0 4 0.624
  1 4 0.594
  2 4 0.526
  3 4 0.471
  4 4 0.323
  5 4 0.326
  6 4 0.361
  7 4 0.545
  8 4 0.226
  9 4 0.115
  0 5 0.576
  1 5 0.473
  2 5 0.377
  3 5 0.334
  4 5 0.288
  5 5 0.254
  6 5 0.200
  7 5 0.366
  8 5 0.150
  9 5 0.187
  0 6 0.441
  1 6 0.330
  2 6 0.271
  3 6 0.350
  4 6 0.393
  5 6 0.261
  6 6 0.156
  7 6 0.280
  8 6 0.097
  9 6 0.198
  0 7 0.430
  1 7 0.387
  2 7 0.269
  3 7 0.315
  4 7 0.322
  5 7 0.236
  6 7 0.194
  7 7 0.340
  8 7 0.193
  9 7 0.251
  0 8 0.721
  1 8 0.561
  2 8 0.319
  3 8 0.332
  4 8 0.335
  5 8 0.238
  6 8 0.220
  7 8 0.402
  8 8 0.231
  9 8 0.218
  0 9 0.953
  1 9 0.690
  2 9 0.390
  3 9 0.364
  4 9 0.438
  5 9 0.302
  6 9 0.276
  7 9 0.478
  8 9 0.294
  9 9 0.245
  };
  \end{axis}

  \begin{axis}[
    heatmap base,
    name=plotD,
    at={(plotC.north east)}, anchor=north west, xshift=2pt,
    title={Gemini-3-Pro (74.6\%)},
    xlabel={Column Depth (\%)}, yticklabels={},
    colorbar right,
    every colorbar/.append style={
      width=0.12cm,
      xshift=-5pt,
      ytick={0,0.5,1.0},
      yticklabels={Low,Mid,High},
      yticklabel style={font=\fontsize{5}{6}\selectfont},
      ylabel={},
      extra description/.code={\node[rotate=-90, font=\fontsize{6}{7}\selectfont, anchor=south]
        at (rel axis cs:4.20,0.5) {Relative Accuracy (\%)};},
    },  ]
  \addplot3[surf, shader=interp, mesh/cols=10] table[meta=C] {
  x y C
  0 0 0.937
  1 0 0.987
  2 0 0.938
  3 0 0.692
  4 0 0.509
  5 0 0.643
  6 0 0.609
  7 0 0.520
  8 0 0.293
  9 0 0.161
  0 1 0.844
  1 1 0.968
  2 1 0.919
  3 1 0.746
  4 1 0.631
  5 1 0.671
  6 1 0.575
  7 1 0.439
  8 1 0.112
  9 1 0.009
  0 2 0.952
  1 2 0.865
  2 2 0.778
  3 2 0.789
  4 2 0.747
  5 2 0.582
  6 2 0.395
  7 2 0.392
  8 2 0.165
  9 2 0.064
  0 3 1.000
  1 3 0.768
  2 3 0.602
  3 3 0.704
  4 3 0.709
  5 3 0.512
  6 3 0.240
  7 3 0.262
  8 3 0.134
  9 3 0.000
  0 4 0.807
  1 4 0.756
  2 4 0.681
  3 4 0.658
  4 4 0.572
  5 4 0.495
  6 4 0.313
  7 4 0.345
  8 4 0.209
  9 4 0.139
  0 5 0.693
  1 5 0.729
  2 5 0.754
  3 5 0.638
  4 5 0.488
  5 5 0.414
  6 5 0.315
  7 5 0.386
  8 5 0.280
  9 5 0.294
  0 6 0.654
  1 6 0.686
  2 6 0.716
  3 6 0.497
  4 6 0.379
  5 6 0.379
  6 6 0.253
  7 6 0.355
  8 6 0.221
  9 6 0.266
  0 7 0.759
  1 7 0.727
  2 7 0.615
  3 7 0.438
  4 7 0.401
  5 7 0.419
  6 7 0.362
  7 7 0.478
  8 7 0.248
  9 7 0.219
  0 8 0.786
  1 8 0.797
  2 8 0.618
  3 8 0.485
  4 8 0.550
  5 8 0.446
  6 8 0.327
  7 8 0.506
  8 8 0.323
  9 8 0.276
  0 9 0.515
  1 9 0.791
  2 9 0.830
  3 9 0.666
  4 9 0.702
  5 9 0.613
  6 9 0.389
  7 9 0.456
  8 9 0.283
  9 9 0.354
  };
  \end{axis}

  \end{tikzpicture}
    \vspace{-2em}
    \caption{Cell-retrieval accuracy across a 10$\times$10 (row $\times$ column) grid, based on 2{,}489 needles from 50 real-world spreadsheet images.
    Each subplot is normalised by its own min--max range to highlight \emph{within-model} positional sensitivity, and the overall accuracy of each model is shown in the subplot title.
    Results for eight additional models are provided in Appendix~\ref{app:needle_full}.}
    \label{fig:needle_heatmap}
  \end{figure*}

\textbf{Cell-position sensitivity.}
To examine whether retrieval accuracy depends on cell location, we construct 2{,}489 cell-retrieval needles from 50 spreadsheet images (Figure~\ref{fig:needle_heatmap}). Two patterns emerge consistently. First, retrieval is generally strongest near the top-left and tends to weaken as cells move toward deeper rows and farther columns, but the relative contribution of row depth and column depth varies across models. Second, several models show a mild recovery at the bottom boundary relative to the adjacent interior rows, suggesting a possible \emph{visual anchor} effect in which boundary rows are easier to localise than rows embedded in the middle of the table. Notably, this bottom-right disadvantage persists even in the strongest models, pointing to a systematic positional bias in visual grounding.


\noindent{\textbf{Reasoning budget scaling.}}
We study how accuracy scales with reasoning effort in three representative models. Figure~\ref{fig:reasoning-cost} plots accuracy against average reasoning tokens (top) and per-query cost (bottom) for Gemini-3-Pro, Gemini-3-Flash, and Kimi-K2.5. All three improve with more reasoning, but returns diminish: enabling reasoning gives Kimi-K2.5 a 15.4\% gain, whereas raising Gemini-3-Flash from minimal to high effort adds only 7.8\%. Gemini-3-Pro gets the best accuracy (67.9\%) but also the highest cost; Gemini-3-Flash at minimal reasoning attains 40.5\% for just \$0.0017 per query, making it the most cost-efficient setting tested. Overall, reasoning budget is an effective but model-dependent lever, and its cost-accuracy tradeoff varies substantially across families (see Appendix~\ref{app:reasoning_budget} for details).

\noindent{\textbf{Multi-hop reasoning.}} To examine how question difficulty scales with reasoning depth, we group questions by the number of table cells accessed to derive the answer: \textbf{L1} (1 cell, $n{=}43$), \textbf{L2} (2--10 cells, $n{=}150$), \textbf{L3} (11--20 cells, $n{=}150$), \textbf{L4} (21--30 cells, $n{=}150$), and \textbf{L5} ($>$30 cells, $n{=}150$). Results are reported on the 643 questions for which both evaluation outputs and hop-level annotations are available (Table~\ref{tab:hop_results}). Most models show a broadly decreasing trend as hop level increases. This decline is particularly pronounced for smaller open-source models: Qwen3-VL-2B-T drops from 11.6\% at L1 to 1.3\% at L5. By contrast, Gemini-3-Pro achieves the highest L1 accuracy (83.7\%) and remains relatively strong at L5 (57.3\%), whereas GPT-5.2 declines from 65.1\% to 41.3\%. These results suggest that robustness on high-complexity multi-hop questions varies substantially across model families.

\providecolor{rcObar}{RGB}{90,90,90}
\providecolor{rcGbarDk}{RGB}{44,143,71}
\providecolor{rcGbarLt}{RGB}{100,185,120}
\providecolor{rcCbar}{RGB}{66,133,244}
\providecolor{rcObarB}{RGB}{160,160,160}
\providecolor{rcGbarDkB}{RGB}{80,170,100}
\providecolor{rcGbarLtB}{RGB}{150,210,165}
\providecolor{rcCbarB}{RGB}{130,175,250}
\providecolor{rcObarF}{RGB}{210,210,210}
\providecolor{rcGbarDkF}{RGB}{180,225,190}
\providecolor{rcGbarLtF}{RGB}{200,235,210}
\providecolor{rcCbarF}{RGB}{200,220,250}

\newcommand{\lmark}[4]{%
  \node[inner sep=1pt,fill=white,draw=#4,line width=0.3pt,
        rounded corners=1pt] at (axis cs:#1,#2)
    {\includegraphics[height=4pt]{icon/#3.png}};%
}
\newcommand{\tlbl}[3]{%
  \node[font=\fontsize{4}{4.5}\selectfont\bfseries,text=black,
        inner sep=0.5pt,fill=white,fill opacity=0.75,text opacity=1,
        rounded corners=0.5pt] at (axis cs:#1,#2) {#3};%
}
\newcommand{\clbl}[4]{%
  \node[font=\fontsize{5}{6}\selectfont\bfseries,text=#4,
        inner sep=0.5pt,fill=white,fill opacity=0.8,text opacity=1,
        rounded corners=0.5pt] at (axis cs:#1,#2) {#3};%
}

\begin{figure*}[t]
\centering
\hspace{-15pt}
\begin{minipage}[t]{0.48\textwidth}
\vspace{-2pt}
\centering
\begin{tikzpicture}
\begin{groupplot}[
  group style={
    group size=1 by 2,
    vertical sep=30pt,
  },
  width=\linewidth, height=0.61\linewidth,
  enlarge x limits=0.05,
  grid=major,
  grid style={line width=0.15pt, black!8},
  ymin=32, ymax=72,
  ytick={35,40,45,50,55,60,65,70},
  yticklabel style={font=\scriptsize, text=black!40},
  tick label style={font=\scriptsize},
  label style={font=\scriptsize},
  clip=false,
  every axis plot/.append style={thick,forget plot},
]

\nextgroupplot[
  ylabel={\scriptsize Accuracy\,(\%)},
  xmin=-400, xmax=6800,
  xtick={0,2000,4000,6000},
  xticklabels={0,2K,4K,6K},
]
\addplot[rcGbarDkF,fill=rcGbarDkF,opacity=0.15,draw=none,forget plot]
  coordinates {(112,32)(112,61.30)(4944,67.90)(4944,32)} \closedcycle;
\addplot[rcGbarDk, thick, mark=none]
  coordinates {(112,61.30)(4944,67.90)};
\lmark{112}{61.30}{gemini}{rcGbarDkB}
\lmark{4944}{67.90}{gemini}{rcGbarDkB}
\clbl{100}{64.5}{low}{rcGbarDk}
\clbl{4510}{70.0}{high}{rcGbarDk}

\addplot[rcGbarLtF,fill=rcGbarLtF,opacity=0.15,draw=none,forget plot]
  coordinates {(0,32)(0,40.45)(1374,46.90)(2828,47.79)(3515,48.20)(3515,32)} \closedcycle;
\addplot[rcGbarLt, thick, densely dashed, mark=none]
  coordinates {(0,40.45)(1374,46.90)(2828,47.79)(3515,48.20)};
\lmark{0}{40.45}{gemini}{rcGbarLtB}
\lmark{1374}{46.90}{gemini}{rcGbarLtB}
\lmark{2828}{47.79}{gemini}{rcGbarLtB}
\lmark{3515}{48.20}{gemini}{rcGbarLtB}
\clbl{-150}{37.5}{minimal}{rcGbarLt!80!black}
\clbl{1700}{50.5}{low}{rcGbarLt!80!black}
\clbl{3000}{44.2}{medium}{rcGbarLt!80!black}
\clbl{3930}{49.2}{high}{rcGbarLt!80!black}

\addplot[rcCbarF,fill=rcCbarF,opacity=0.15,draw=none,forget plot]
  coordinates {(0,32)(0,34.53)(6431,49.90)(6431,32)} \closedcycle;
\addplot[rcCbar, thick, mark=none]
  coordinates {(0,34.53)(6431,49.90)};
\lmark{0}{34.53}{kimi}{rcCbarB}
\lmark{6431}{49.90}{kimi}{rcCbarB}
\clbl{870}{34.0}{disabled}{rcCbar}
\clbl{6100}{53.5}{enabled}{rcCbar}

\nextgroupplot[
  ylabel={\scriptsize Accuracy\,(\%)},
  x coord trafo/.code={%
    \pgfmathparse{#1<0.02 ? #1*2.0 : 0.04+(#1-0.02)*0.71429}%
  },
  x coord inv trafo/.code={%
    \pgfmathparse{#1<0.04 ? #1/2.0 : 0.02+(#1-0.04)/0.71429}%
  },
  xmin=-0.0005, xmax=0.090,
  xtick={0,0.005,0.01,0.015,0.02,0.04,0.06,0.08},
  xticklabels={0,.005,.010,.015,.020,.040,.060,.080},
  scaled x ticks=false,
]
\addplot[rcGbarDkF,fill=rcGbarDkF,opacity=0.15,draw=none,forget plot]
  coordinates {(0.0110,32)(0.0110,61.30)(0.0671,67.90)(0.0671,32)} \closedcycle;
\addplot[rcGbarDk, thick, mark=none]
  coordinates {(0.0110,61.30)(0.0671,67.90)};
\lmark{0.0110}{61.30}{gemini}{rcGbarDkB}
\lmark{0.0671}{67.90}{gemini}{rcGbarDkB}
\clbl{0.0120}{64.5}{low}{rcGbarDk}
\clbl{0.0550}{70.0}{high}{rcGbarDk}

\addplot[rcGbarLtF,fill=rcGbarLtF,opacity=0.15,draw=none,forget plot]
  coordinates {(0.0017,32)(0.0017,40.45)(0.0057,46.90)(0.0101,47.79)(0.0121,48.20)(0.0121,32)} \closedcycle;
\addplot[rcGbarLt, thick, densely dashed, mark=none]
  coordinates {(0.0017,40.45)(0.0057,46.90)(0.0101,47.79)(0.0121,48.20)};
\lmark{0.0017}{40.45}{gemini}{rcGbarLtB}
\lmark{0.0057}{46.90}{gemini}{rcGbarLtB}
\lmark{0.0101}{47.79}{gemini}{rcGbarLtB}
\lmark{0.0121}{48.20}{gemini}{rcGbarLtB}
\clbl{0.0010}{37.5}{minimal}{rcGbarLt!80!black}
\clbl{0.0030}{47.5}{low}{rcGbarLt!80!black}
\clbl{0.0100}{44.7}{medium}{rcGbarLt!80!black}
\clbl{0.0130}{51.5}{high}{rcGbarLt!80!black}

\addplot[rcCbarF,fill=rcCbarF,opacity=0.15,draw=none,forget plot]
  coordinates {(0.003,32)(0.003,34.53)(0.021,49.90)(0.021,32)} \closedcycle;
\addplot[rcCbar, thick, mark=none]
  coordinates {(0.003,34.53)(0.021,49.90)};
\lmark{0.003}{34.53}{kimi}{rcCbarB}
\lmark{0.021}{49.90}{kimi}{rcCbarB}
\clbl{0.0080}{34.5}{disabled}{rcCbar}
\clbl{0.025}{53.5}{enabled}{rcCbar}

\end{groupplot}

\node[draw=black!30, rounded corners=2pt, inner xsep=3pt, inner ysep=1.5pt,
      fill=white, line width=0.3pt, anchor=south east]
  at ($(group c1r2.south east)+(-3pt,3pt)$)
  {\fontsize{6}{7}\selectfont
    \begin{tabular}{@{}c@{\,}c@{\,}l@{}}
      \raisebox{-0.5pt}{\includegraphics[height=4pt]{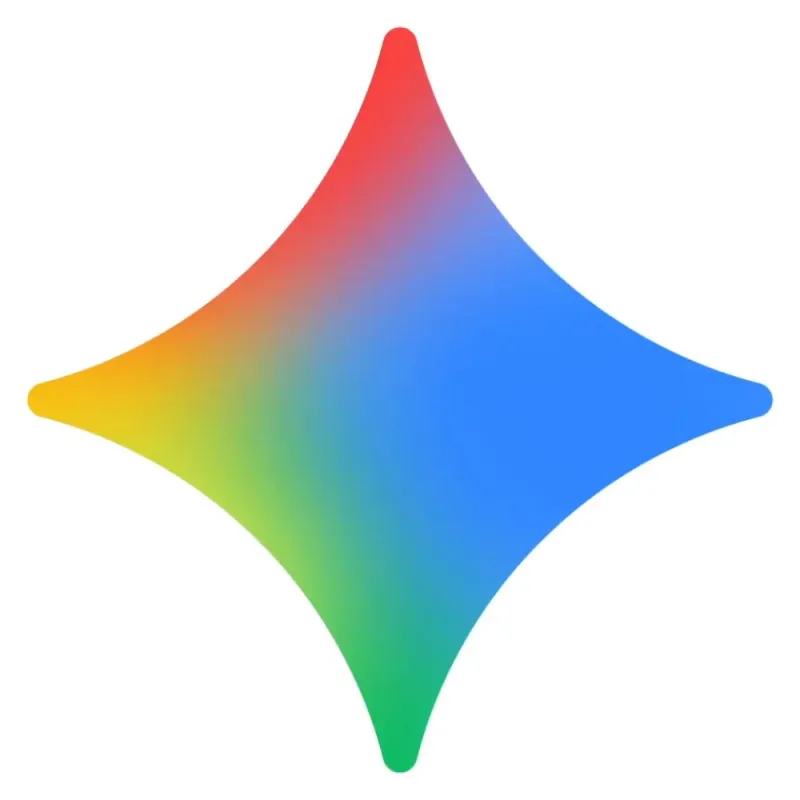}} &
      \makebox[8pt]{\color{rcGbarDk}\rule[2pt]{6pt}{0.8pt}} &
      Gemini-3-Pro\\[0.5pt]
      \raisebox{-0.5pt}{\includegraphics[height=4pt]{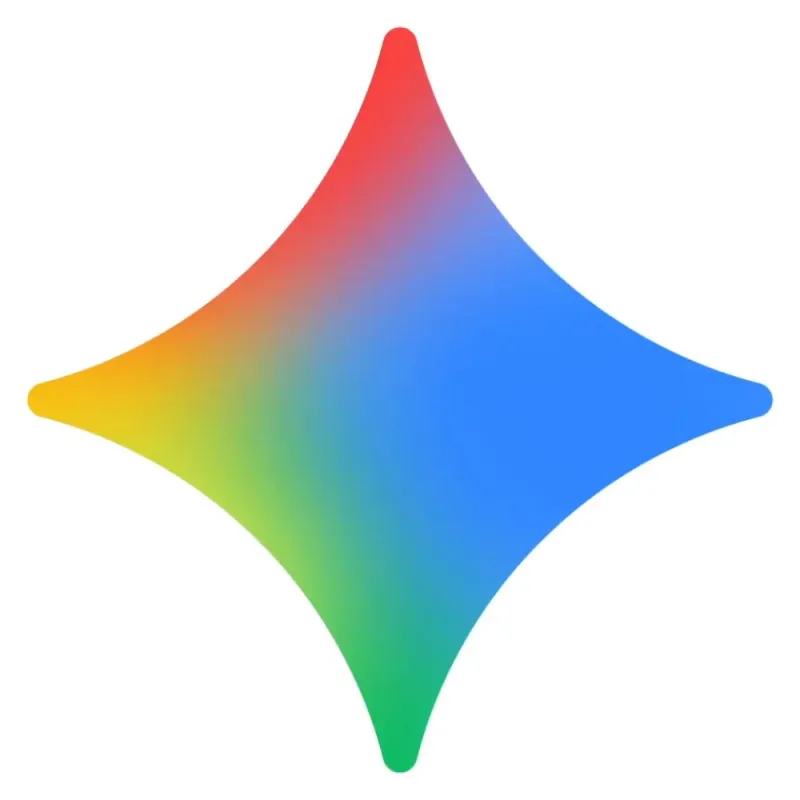}} &
      \makebox[8pt]{\color{rcGbarLt}\rule[2pt]{2pt}{0.8pt}\hskip0.6pt\rule[2pt]{2pt}{0.8pt}\hskip0.6pt\rule[2pt]{2pt}{0.8pt}} &
      Gemini-3-Flash\\[0.5pt]
      \raisebox{-0.5pt}{\includegraphics[height=4pt]{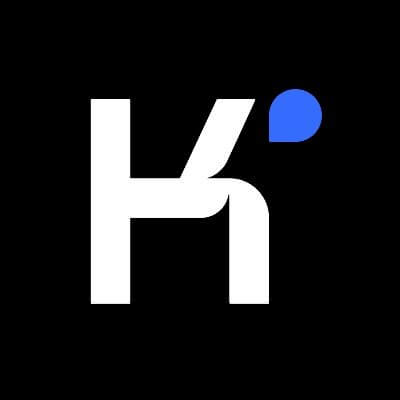}} &
      \makebox[8pt]{\color{rcCbar}\rule[2pt]{6pt}{0.8pt}} &
      Kimi-K2.5%
    \end{tabular}%
  };

\node[font=\scriptsize,anchor=north]
  at ($(group c1r1.south)+(0,-10pt)$) {\textbf{(a)} Average Reasoning Tokens};
\node[font=\scriptsize,anchor=north]
  at ($(group c1r2.south)+(0,-10pt)$) {\textbf{(b)} Average Cost (\$)};

\end{tikzpicture}
\end{minipage}%
\hspace{2pt}%
\begin{minipage}[t]{0.52\textwidth}
\vspace{0pt}
\centering
\scriptsize
\setlength{\tabcolsep}{0.4mm}
\resizebox{\linewidth}{!}{%
\begin{tabular}{l|ccccc|c}
  \hline
  \noalign{\vskip 3pt}
    \multirow{2}{*}{\textbf{Model}}  &  \textbf{L1} & \textbf{L2} & \textbf{L3} &\textbf{ L4 }& \textbf{L5} & \multirow{2}{*}{\textbf{Overall}} \\
    &  \textbf{1 cell} & \textbf{2--10 cells} & \textbf{11--20 cells} & \textbf{21--30 cells} & \textbf{$>$30 cells} &  \\
  & \multicolumn{1}{c}{\scriptsize($n$=43)} & {\scriptsize($n$=150)} & {\scriptsize($n$=150)} & {\scriptsize($n$=150)} & {\ v($n$=150)} & \\
  \noalign{\vskip 3pt}
  \hline
  \noalign{\vskip 1pt}
  \multicolumn{7}{l}{\textbf{Proprietary LMMs}} \\
  \noalign{\vskip 1pt}
  \hline
  Gemini-3-Pro & \rgb{83.7}\textbf{83.7} & \rgb{71.3}71.3 & \rgb{72.7}72.7 & \rgb{56.7}56.7 & \rgb{57.3}57.3 & \rgb{65.8}65.8 \\
  Gemini-3-Flash & \rgb{67.4}\textbf{67.4} & \rgb{61.3}61.3 & \rgb{52.7}52.7 & \rgb{40.7}40.7 & \rgb{34.7}34.7 & \rgb{48.7}48.7 \\
  Seed-2.0-Pro & \rgb{76.7}\textbf{76.7} & \rgb{55.3}55.3 & \rgb{44.0}44.0 & \rgb{44.7}44.7 & \rgb{36.0}36.0 & \rgb{47.1}47.1 \\
  GPT-5.2 & \rgb{65.1}\textbf{65.1} & \rgb{47.3}47.3 & \rgb{45.3}45.3 & \rgb{48.7}48.7 & \rgb{41.3}41.3 & \rgb{47.0}47.0 \\
  Claude-Opus-4.6 & \rgb{60.5}\textbf{60.5} & \rgb{50.7}50.7 & \rgb{43.3}43.3 & \rgb{47.3}47.3 & \rgb{35.3}35.3 & \rgb{45.3}45.3 \\
  Claude-Sonnet-4.6 & \rgb{48.8}\textbf{48.8} & \rgb{40.0}40.0 & \rgb{33.3}33.3 & \rgb{31.3}31.3 & \rgb{26.0}26.0 & \rgb{33.7}33.7 \\
  GPT-5-mini & \rgb{32.6}\textbf{32.6} & \rgb{24.0}24.0 & \rgb{27.3}27.3 & \rgb{25.3}25.3 & \rgb{22.7}22.7 & \rgb{25.3}25.3 \\
  o3 & \rgb{20.9}\textbf{20.9} & \rgb{19.3}19.3 & \rgb{14.7}14.7 & \rgb{13.3}13.3 & \rgb{10.7}10.7 & \rgb{14.9}14.9 \\
  GPT-4o & \rgb{14.0}\textbf{14.0} & \rgb{7.3}7.3 & \rgb{6.0}6.0 & \rgb{5.3}5.3 & \rgb{3.3}3.3 & \rgb{6.1}6.1 \\
  \hline
  \noalign{\vskip 1pt}
  \multicolumn{7}{l}{\textbf{Open-source LMMs: Thinking}} \\
  \noalign{\vskip 1pt}
  \hline
  Kimi-K2.5 & \rgb{51.2}51.2 & \rgb{52.0}\textbf{52.0} & \rgb{49.3}49.3 & \rgb{49.3}49.3 & \rgb{46.0}46.0 & \rgb{49.3}49.3 \\
  Qwen3-VL-235B-T & \rgb{51.2}\textbf{51.2} & \rgb{37.3}37.3 & \rgb{31.3}31.3 & \rgb{35.3}35.3 & \rgb{34.0}34.0 & \rgb{35.6}35.6 \\
  Qwen3-VL-32B-T & \rgb{37.2}37.2 & \rgb{40.7}\textbf{40.7} & \rgb{28.7}28.7 & \rgb{23.3}23.3 & \rgb{22.7}22.7 & \rgb{29.4}29.4 \\
  GLM-4.6V & \rgb{34.9}\textbf{34.9} & \rgb{32.0}32.0 & \rgb{14.7}14.7 & \rgb{28.7}28.7 & \rgb{14.7}14.7 & \rgb{23.3}23.3 \\
  Qwen3-VL-8B-T & \rgb{25.6}\textbf{25.6} & \rgb{18.0}18.0 & \rgb{17.3}17.3 & \rgb{12.0}12.0 & \rgb{10.7}10.7 & \rgb{15.2}15.2 \\
  Qwen3-VL-4B-T & \rgb{23.3}\textbf{23.3} & \rgb{13.3}13.3 & \rgb{6.7}6.7 & \rgb{7.3}7.3 & \rgb{2.7}2.7 & \rgb{8.6}8.6 \\
  Qwen3-VL-2B-T & \rgb{11.6}\textbf{11.6} & \rgb{6.0}6.0 & \rgb{5.3}5.3 & \rgb{3.3}3.3 & \rgb{1.3}1.3 & \rgb{4.5}4.5 \\
  \hline
  \noalign{\vskip 1pt}
  \multicolumn{7}{l}{\textbf{Open-source LMMs: Instruct}} \\
  \noalign{\vskip 1pt}
  \hline
  Qwen3-VL-235B-I & \rgb{41.9}\textbf{41.9} & \rgb{31.3}31.3 & \rgb{16.7}16.7 & \rgb{29.3}29.3 & \rgb{18.7}18.7 & \rgb{25.2}25.2 \\
  Qwen3-VL-32B-I & \rgb{39.5}\textbf{39.5} & \rgb{31.3}31.3 & \rgb{18.7}18.7 & \rgb{26.0}26.0 & \rgb{15.3}15.3 & \rgb{24.0}24.0 \\
  Qwen3-VL-8B-I & \rgb{25.6}\textbf{25.6} & \rgb{12.0}12.0 & \rgb{9.3}9.3 & \rgb{11.3}11.3 & \rgb{3.3}3.3 & \rgb{10.1}10.1 \\
  Qwen3-VL-4B-I & \rgb{18.6}\textbf{18.6} & \rgb{9.3}9.3 & \rgb{3.3}3.3 & \rgb{10.7}10.7 & \rgb{6.0}6.0 & \rgb{8.1}8.1 \\
  Qwen3-VL-2B-I & \rgb{16.3}\textbf{16.3} & \rgb{11.3}11.3 & \rgb{1.3}1.3 & \rgb{7.3}7.3 & \rgb{2.0}2.0 & \rgb{6.2}6.2 \\
  \hline
  \end{tabular}%
}
\end{minipage}%
\par\vspace{4pt}\noindent
\begin{minipage}[t]{0.47\textwidth}
\captionof{figure}{\small \textbf{Reasoning budget.}
\textbf{(a)} accuracy \textit{vs.}\ average reasoning tokens;
\textbf{(b)} accuracy \textit{vs.}\ per-query cost (\$) for Gemini-3-Pro (low/high), Gemini-3-Flash (minimal/low/medium/high), and Kimi-K2.5 (disabled/enabled).}
\label{fig:reasoning-cost}
\end{minipage}%
\hfill
\begin{minipage}[t]{0.51\textwidth}
\captionof{table}{\small Accuracy (\%) by multi-level hop reasoning (cells accessed per question). Column headers show the cells-accessed range; $n$ = question count per bin. Cell shading indicates accuracy (darker green = higher). \textbf{Bold} = best per column.}
\label{tab:hop_results}
\end{minipage}
\end{figure*}

\textbf{Image-type sensitivity.} 
We also compare model performance on spreadsheet and non-spreadsheet images to assess
whether image type affects table understanding. 
The question distribution across the two image types is relatively balanced, at 54.6\% versus 45.4\%. Effects are heterogeneous across models rather than following a single direction: Seed-2.0-Pro shows the largest spreadsheet advantage (+8.7\%), with Claude-Opus-4.6 also higher (+4.9\%), whereas Gemini-3-Flash and Kimi-K2.5 score 7.2\% and 7.7\% \emph{lower} on spreadsheets. A similar reversal appears across question categories: spreadsheet questions are easier for Verification (+5.1\%) but harder for Hypothetical reasoning ($-$9.1\%). These results suggest that image type alone does not determine difficulty; rather, its effect depends on the reasoning required.

\subsection{Error Analysis}
\label{sec:error_analysis}
Figure~\ref{fig:error-ana}~(a) shows a strikingly consistent pattern across representative models: most failures arise before explicit reasoning, in the perceptual stages of \textit{Locating} and \textit{Recognition}. This analysis uses a balanced subset of five categories, reducing the effect of category imbalance. For nearly all models, these two categories together account for the majority of errors, indicating that the main bottleneck on WildTableBench is reliable visual grounding rather than downstream deduction alone. In other words, models often fail either to identify the correct region in a dense real-world table or to accurately read the relevant cell content once the region has been found. As overall accuracy decreases, both error types increase substantially, with locating errors growing especially sharply for weaker models, while \textit{Reasoning} and \textit{Comprehension} remain secondary but still non-negligible. 
%
%
Figure~\ref{fig:error-ana}~(b) further illustrates two examples missed by all models, both requiring precise cell selection over large search spaces: one combines conditional filtering with listing in a dense attendance table, and the other requires color-based reasoning across multiple rows and consecutive entries. Taken together, these results suggest that future gains on real-world table VQA will depend at least as much on improving fine-grained table perception, visual parsing, and grounding as on strengthening high-level reasoning.

\providecolor{erLoc}{RGB}{199,234,228}   
\providecolor{erRec}{RGB}{167,232,189}   
\providecolor{erRes}{RGB}{252,188,184}   
\providecolor{erCmp}{RGB}{232,168,191}   

\newcommand{\micon}[1]{\,\raisebox{-1.2pt}{\includegraphics[height=6pt]{icon/#1.png}}}

\begin{figure*}[t]
\centering
\begin{minipage}[t]{0.46\textwidth}\vspace{0pt}
\makebox[\linewidth][r]{%
\begin{tikzpicture}
\begin{axis}[
  xbar stacked,
  width=.95\linewidth,
  height=1.05\textwidth,
  bar width=7pt,
  enlarge y limits=0.045,
  ytick={0,1,2,3,4,5,6,7,8,9,10,11,12},
  yticklabels={
    {Gemini-3-Pro\micon{gemini}},
    {Kimi-K2.5\micon{kimi}},
    {Gemini-3-Flash\micon{gemini}},
    {Claude-Opus-4.6\micon{claude}},
    {Claude-Sonnet-4.6\micon{claude}},
    {GLM-4.6V\micon{zai}},
    {GPT-5-mini\micon{openai}},
    {o3\micon{openai}},
    {GPT-4o\micon{openai}},
    {Qwen3-VL-4B-T\micon{qwen}},
    {Qwen3-VL-4B-I\micon{qwen}},
    {Qwen3-VL-2B-I\micon{qwen}},
    {Qwen3-VL-2B-T\micon{qwen}},
  },
  yticklabel style={font=\fontsize{5.5}{6.5}\selectfont, anchor=east},
  xticklabel style={font=\scriptsize, text=black!40},
  axis line style={draw=none},
  xtick style={draw=none},
  ytick style={draw=none},
  label style={font=\scriptsize},
  xlabel={\scriptsize Error count},
  xmin=0, xmax=650,
  xtick={0,100,200,300,400,500,600},
  xmajorgrids=true,
  ymajorgrids=false,
  grid style={line width=0.15pt, black!8},
  every axis plot/.append style={rounded corners=1pt},
  legend style={
    at={(.99,0.015)},
    anchor=south east,
    font=\fontsize{6}{7}\selectfont,
    draw=black!30,
    fill=white,
    rounded corners=2pt,
    line width=0.4pt,
    inner xsep=2pt,
    inner ysep=1pt,
    row sep=-2pt,
  },
  legend columns=1,
  legend cell align=left,
  legend image code/.code={%
    \fill[#1,rounded corners=0.5pt] (0cm,-0.06cm) rectangle (0.25cm,0.14cm);},
  clip=false,
]
\addplot[fill=erLoc, draw=erLoc!55!black, line width=0.15pt] coordinates {
  (75,0) (100,1) (120,2) (130,3) (170,4) (180,5) (195,6) (220,7) (275,8) (260,9) (230,10) (230,11) (255,12)
};
\addplot[fill=erRec, draw=erRec!55!black, line width=0.15pt] coordinates {
  (45,0) (95,1) (90,2) (115,3) (155,4) (170,5) (160,6) (160,7) (165,8) (170,9) (190,10) (200,11) (195,12)
};
\addplot[fill=erRes, draw=erRes!55!black, line width=0.15pt] coordinates {
  (25,0) (45,1) (40,2) (45,3) (65,4) (80,5) (95,6) (95,7) (80,8) (95,9) (110,10) (115,11) (115,12)
};
\addplot[fill=erCmp, draw=erCmp!55!black, line width=0.15pt] coordinates {
  (10,0) (25,1) (20,2) (20,3) (35,4) (40,5) (40,6) (40,7) (30,8) (40,9) (45,10) (40,11) (45,12)
};
\legend{Locating, Recognition, Reasoning, Comprehension}
\end{axis}
\end{tikzpicture}}%
\par\vspace{-1pt}
{\centering\small\textbf{(a)} Error type distribution\par}
\end{minipage}\hfill
\begin{minipage}[t]{0.53\textwidth}\vspace{0pt}
  \setlength{\fboxsep}{0pt}\setlength{\fboxrule}{0.25pt}%
  \savebox{\imgbox}{\adjustbox{trim={.02\width} {.06\height} {.02\width} {.26\height},clip,width=\dimexpr0.46\linewidth-0.5pt\relax}{\includegraphics{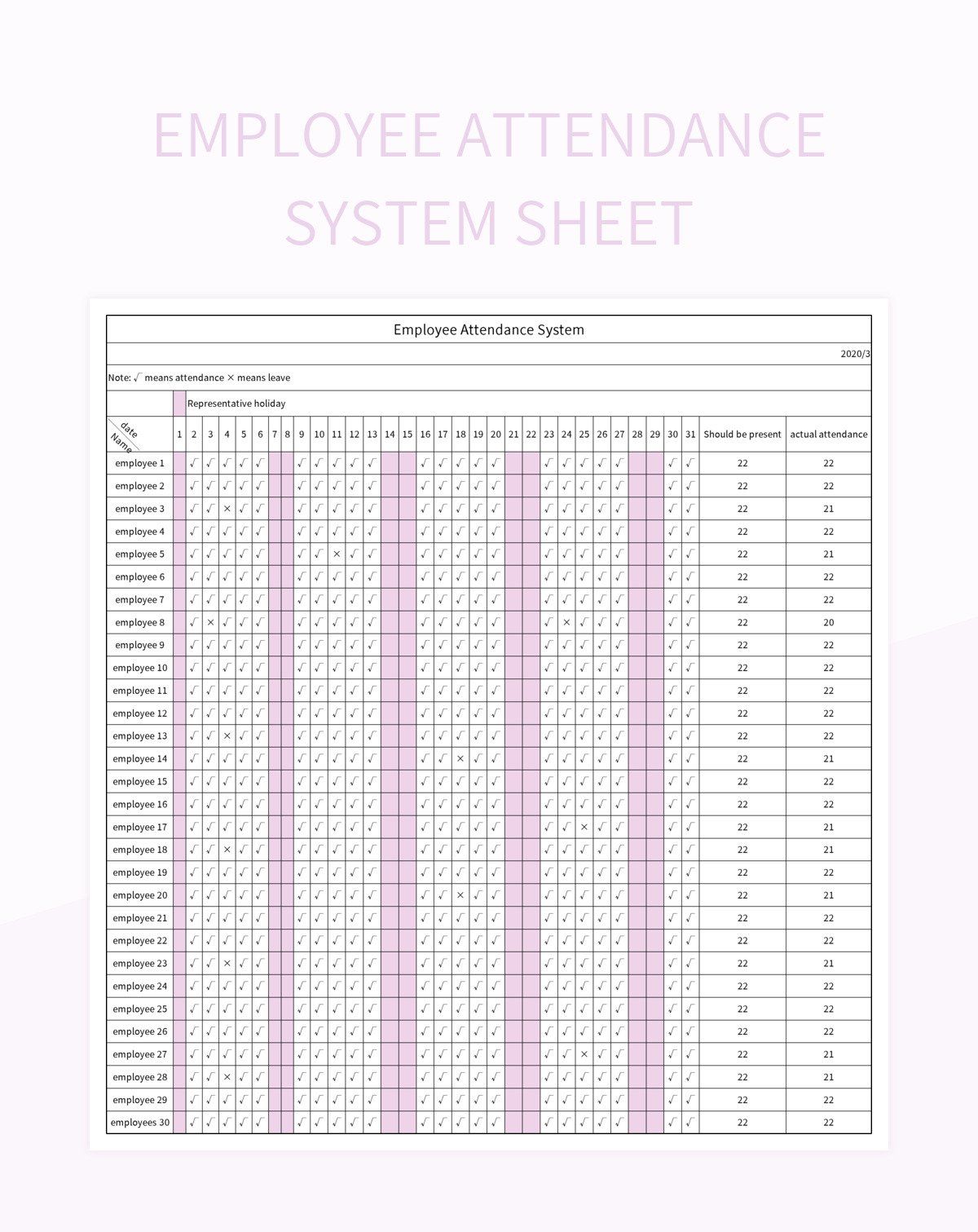}}}%
  \noindent
  \begin{minipage}[t]{0.46\linewidth}\vspace{0pt}%
    \fbox{\usebox{\imgbox}}%
  \end{minipage}\hfill
  \begin{minipage}[t]{0.52\linewidth}\vspace{0pt}%
    \raggedright\tiny
    \textbf{Question:} \textit{Which dates had at least one employee absent, excluding representative holidays?}\par
    \vspace{1pt}
    \textbf{Answer:}\;\,\colorbox{green!45}{\rule[-2pt]{0pt}{9pt}\;\textbf{3, 4, 11, 18, 24 and 25}\;}%
    \par\vspace{2pt}
    {\color{black!50}\hrule height 0.3pt}\par\vspace{2pt}
    \fontsize{5.5}{7}\selectfont
    \textcolor{black!55}{Image Domain:}\;\cattag{blue!8}{Business \& Management}\\
    \textcolor{black!55}{Question Category:}\\
    \cattag{teal!10}{Multi-step Conditional}\\
    \textcolor{black!55}{Required Skills:}\\
    \cattag{orange!10}{Cell Locating}\;\cattag{red!7}{Visual\ Parsing}%
  \end{minipage}%
  \par\vspace{2pt}
  \savebox{\imgbox}{\adjustbox{trim=0 {.00\height} 0 0,clip,width=\dimexpr0.46\linewidth-0.5pt\relax}{\includegraphics{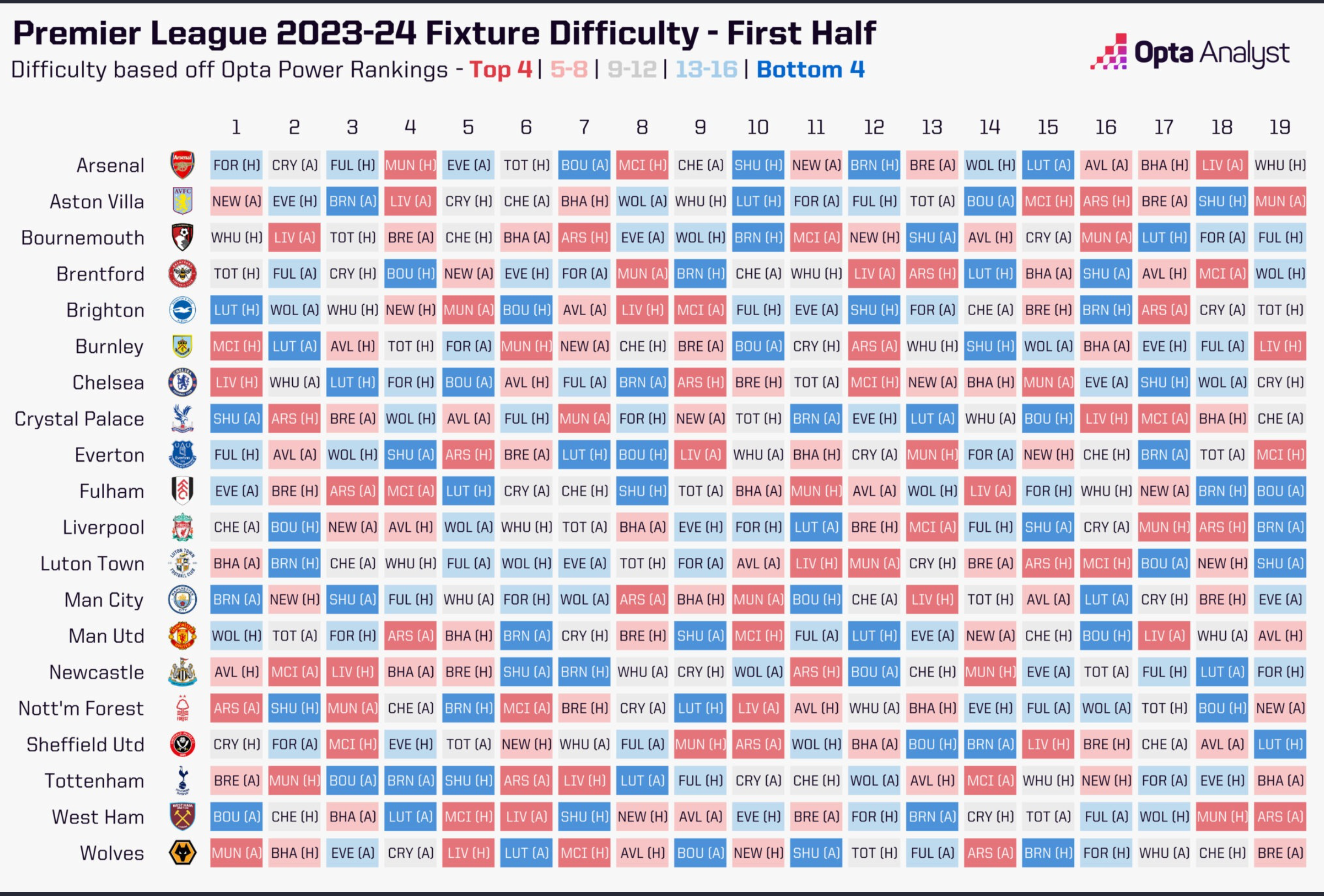}}}%
  \noindent
  \begin{minipage}[t]{0.46\linewidth}\vspace{0pt}%
    \fbox{\usebox{\imgbox}}%
  \end{minipage}\hfill
  \begin{minipage}[t]{0.52\linewidth}\vspace{0pt}%
    \raggedright\tiny
    \textbf{Question:} \textit{List teams facing 'bottom 4' opponents in two consecutive matches per Opta difficulty ratings.}\par
    \vspace{1pt}
    \textbf{Answer:}\;\,
    {\begingroup
    \tiny\bfseries
    \sethlcolor{green!45}%
    \hl{Everton, Newcastle, Sheffield Utd, Tottenham}%
    \endgroup}
    \par\vspace{1pt}
    {\color{black!50}\hrule height 0.3pt}\par\vspace{2pt}
    \fontsize{5.5}{7}\selectfont
    \textcolor{black!55}{Image Domain:}\;\cattag{blue!8}{Sports \& Health}\\
    \textcolor{black!55}{Question Category:}\\
    \cattag{teal!10}{Color-based Reasoning}\\
    \textcolor{black!55}{Required Skills:}\\
    \cattag{orange!10}{Color Filtering}\;\cattag{red!7}{Multi-hop Reasoning}%
  \end{minipage}%
  \par\vspace{10pt}
  {\centering\small\textbf{(b)} All models failed examples\par}
\end{minipage}

\vspace{-8pt}
\caption{\small Error analysis. \textbf{(a)} Error-type breakdown (locating, recognition, reasoning, comprehension) across representative models. Each horizontal stacked bar reports absolute error counts. Perception-related errors (Locating + Recognition) dominate across most model families. \textbf{(b)} Examples of questions that all models answer incorrectly.}
\label{fig:error-ana}
\end{figure*}

\section{Conclusion}
We present WildTableBench, a benchmark for evaluating multimodal
foundation models on real-world table image understanding.
Unlike prior benchmarks that rely on structured text representations or
programmatically rendered images, WildTableBench is built from 402 table images collected directly from real-world web sources. Based on these images, we annotate 928 questions spanning five categories and 17 subtypes.
We evaluate 21 frontier and representative multimodal foundation models and conduct fine-grained analyses by \textbf{category, subtype, reasoning depth, cell position, image type, reasoning budget, and error source}.
Our results show that real-world table understanding remains difficult for current models. Only one evaluated model exceeds 50\% accuracy, and strong open-source models still lag behind frontier proprietary systems. Model performance also varies substantially across skills: verification is relatively tractable, while color-based, hypothetical, and higher-hop questions remain much harder. More importantly, our analyses show that \textit{the main bottleneck is reliable visual grounding rather than high-level reasoning alone}: most errors come from locating and recognition failures, and even strong models exhibit persistent positional biases. We hope WildTableBench helps diagnose these limitations and guide future progress.

\section*{Acknowledgments}
We thank the publicly available websites, forums, and online communities whose shared real-world table images made this benchmark possible. We also thank the annotators and reviewers who contributed to question design, answer verification, and quality control.

\section*{Ethics Statement}
WildTableBench is constructed from publicly available table images collected from real-world web sources. We focus on tables that support research on multimodal table understanding and avoid intentionally collecting private or highly sensitive content. Nevertheless, because web data may contain bias or context-specific conventions, the benchmark may reflect domain imbalance and annotation bias inherited from the source materials.

Our benchmark is intended solely for research use in evaluating multimodal foundation models on realistic table image understanding. It should not be treated as evidence that a model is reliable for high-stakes applications such as medical, financial, legal, or safety-critical decision-making. We hope this dataset is used to diagnose model limitations and improve robustness, especially in visual grounding and reasoning over real-world tables.

\bibliography{ref}
\bibliographystyle{configs/colm2026_conference}

\newpage
\appendix
\onecolumn
\appendixpage

\begin{spacing}{1}
	\section*{Contents}

	\startcontents[appendices]
	\printcontents[appendices]{}{-1}{\setcounter{tocdepth}{2}}
\end{spacing}

\section{Keyword-Based Image Collection}
\label{app:keyword}

\begin{figure}[t]
\centering
\includegraphics[width=\textwidth]{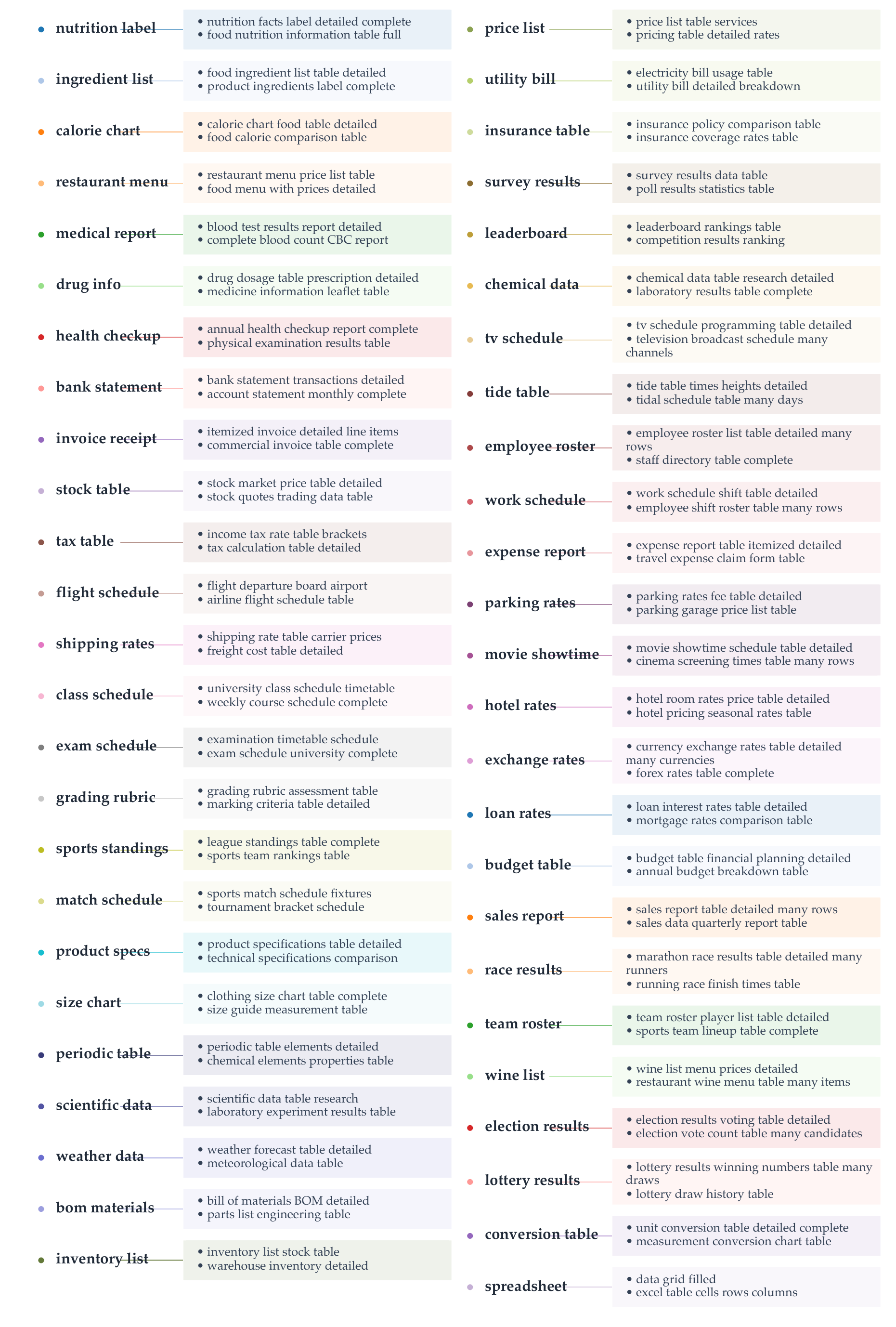}
\caption{\textbf{Overview of keyword-based image collection.}
Keyword schema used for candidate retrieval. Each entry denotes a representative table type and shows example spreadsheet-style and scenario-grounded queries used to retrieve candidate table images from the open web. Together, these query families expand coverage across diverse real-world table scenarios.}
\label{fig:keyword_summary_appendix}
\end{figure}

Figure~\ref{fig:keyword_summary_appendix} summarizes the keyword schema used to seed candidate retrieval. Rather than relying on a narrow set of generic spreadsheet prompts, we construct a broader retrieval space by combining spreadsheet-oriented prompts that foreground tabular structure with scenario-grounded web queries that reflect realistic search intents. The schematic is organized by representative table types and lists the corresponding example queries used for retrieval, covering a wide range of real-world settings, including nutrition labels, medical reports, financial records, timetables, schedules, price lists, rankings, and scientific tables. This design increases semantic and visual coverage at the retrieval stage and helps surface candidate images that better reflect the heterogeneity of naturally occurring table images.

\section{Image Domain Taxonomy}
\label{app:categories}

\begin{figure*}[p]
\centering
\setlength{\fboxsep}{0pt}

\begin{minipage}[t]{0.48\linewidth}
  \vspace{0pt}\centering
  \includegraphics[width=\linewidth]{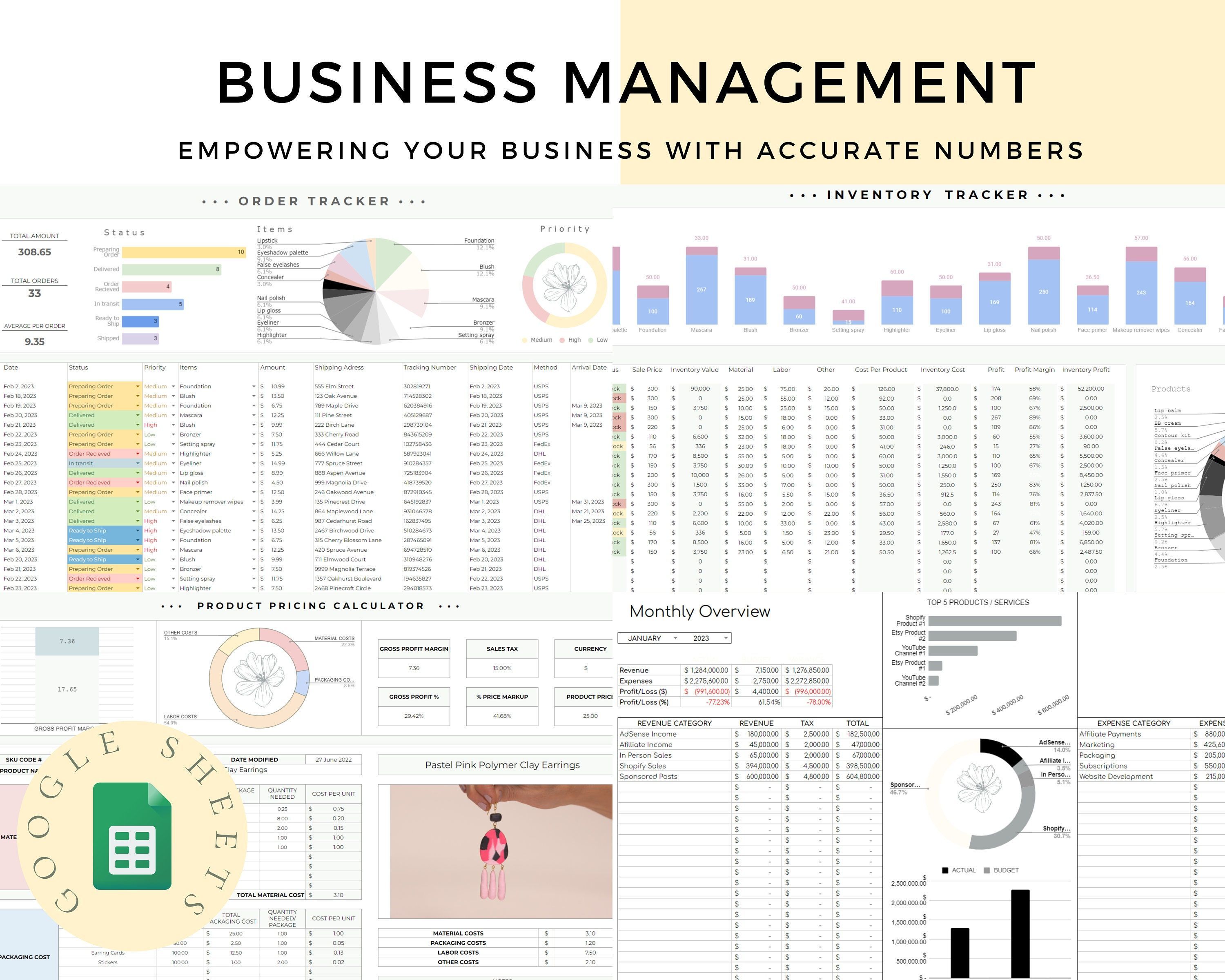}
\end{minipage}\hfill
\begin{minipage}[t]{0.48\linewidth}
  \vspace{0pt}\centering
  \includegraphics[width=\linewidth]{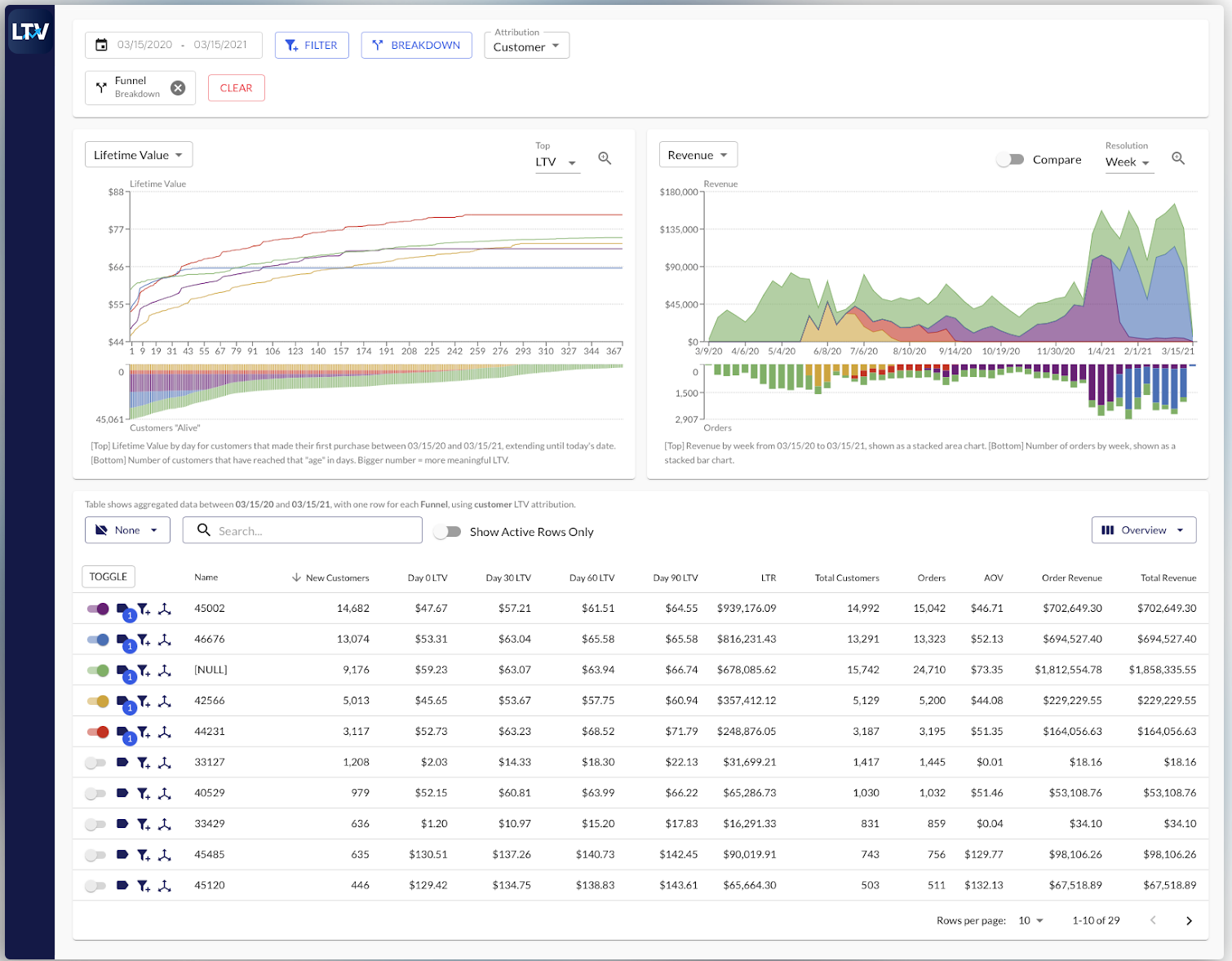}
\end{minipage}\\[4pt]
\begin{minipage}[t]{0.48\linewidth}
  \vspace{0pt}\centering
  \includegraphics[width=\linewidth]{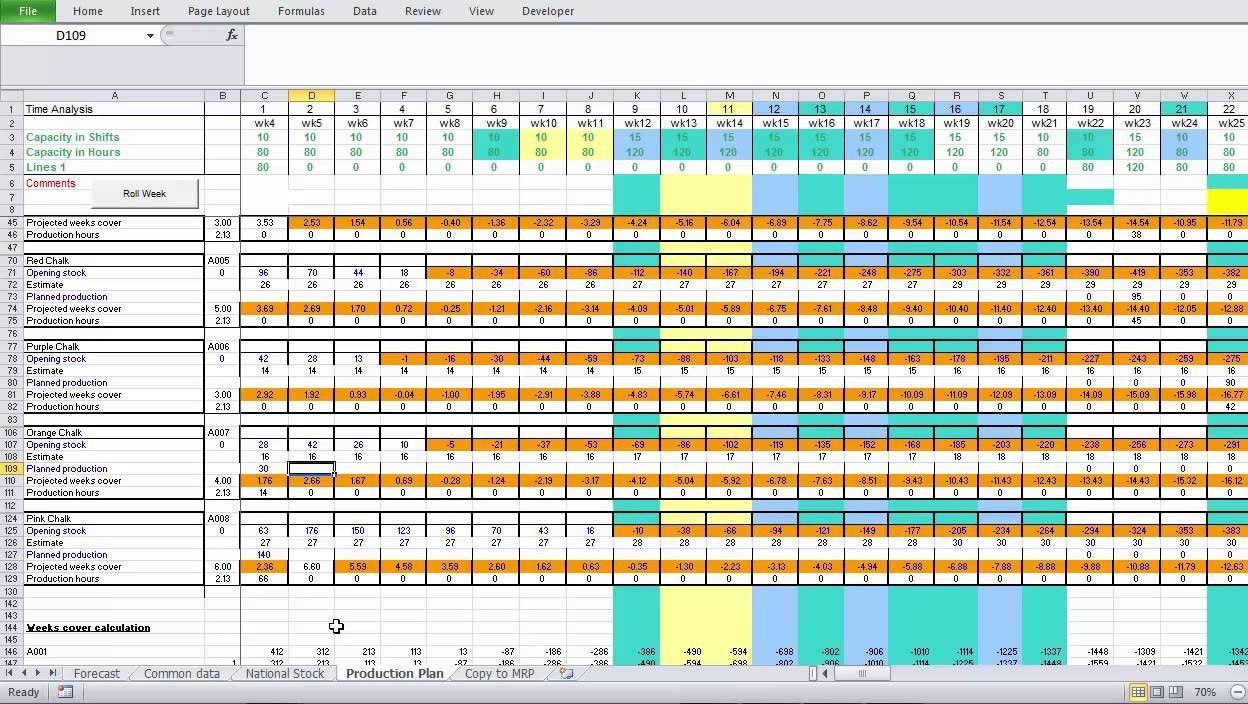}
\end{minipage}\hfill
\begin{minipage}[t]{0.48\linewidth}
  \vspace{0pt}\centering
  \includegraphics[width=\linewidth]{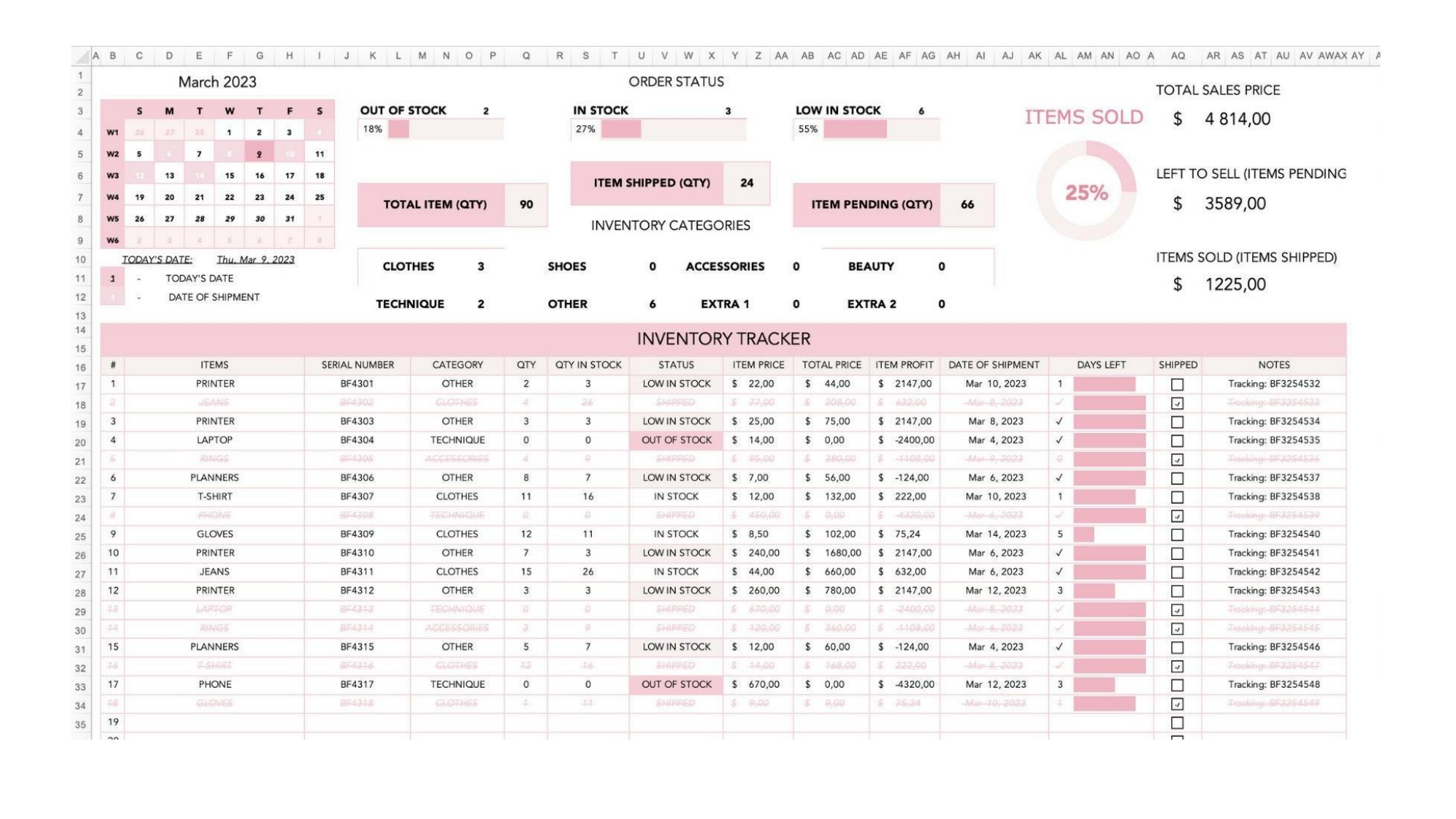}
\end{minipage}\\[3pt]
\captionof{figure}{\textbf{Business \& Management}}
\label{fig:domain-business}

\vspace{10pt}

\begin{minipage}[c]{0.48\linewidth}
  \centering
  \includegraphics[width=\linewidth]{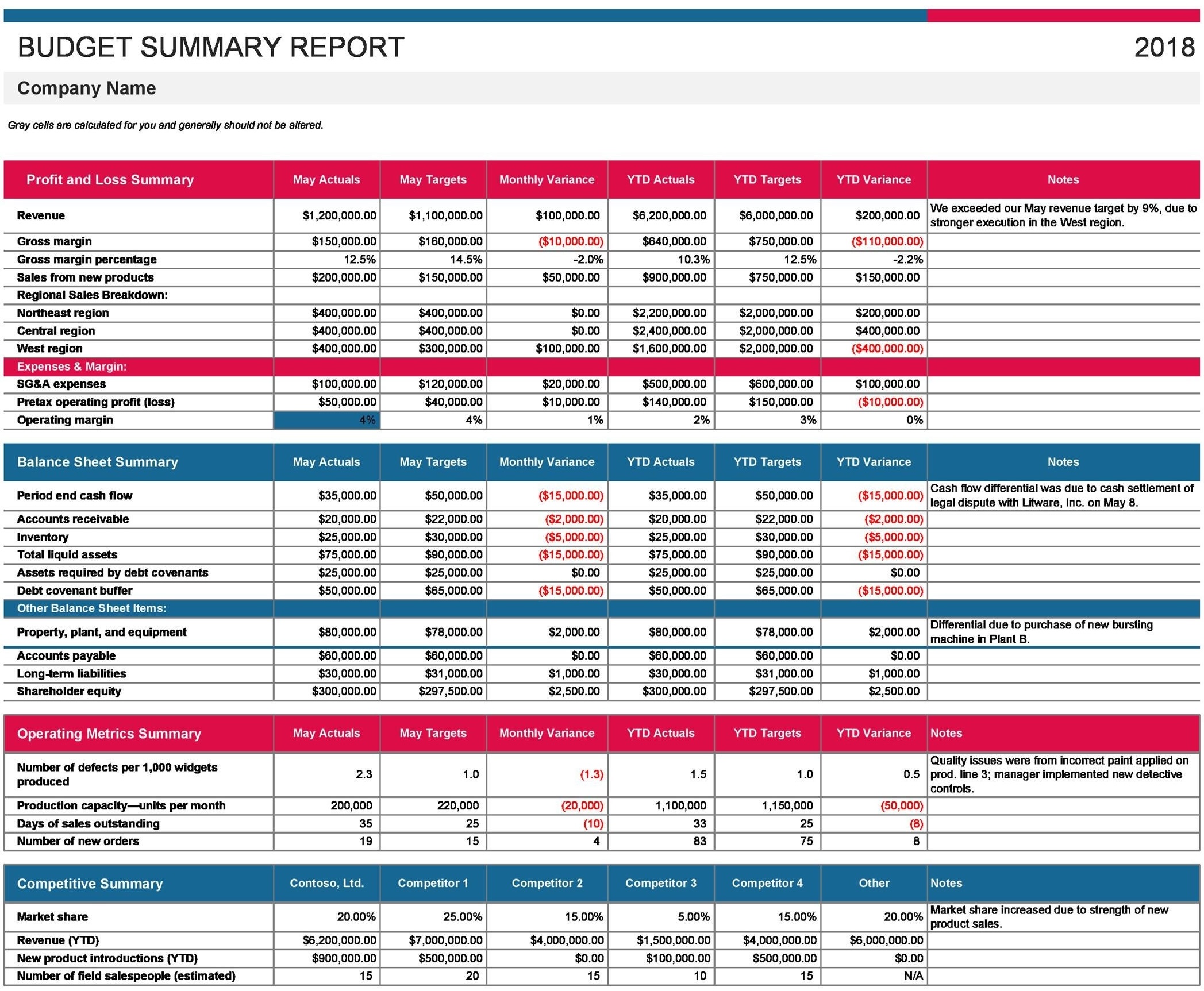}
\end{minipage}\hfill
\begin{minipage}[c]{0.48\linewidth}
  \centering
  \includegraphics[width=\linewidth]{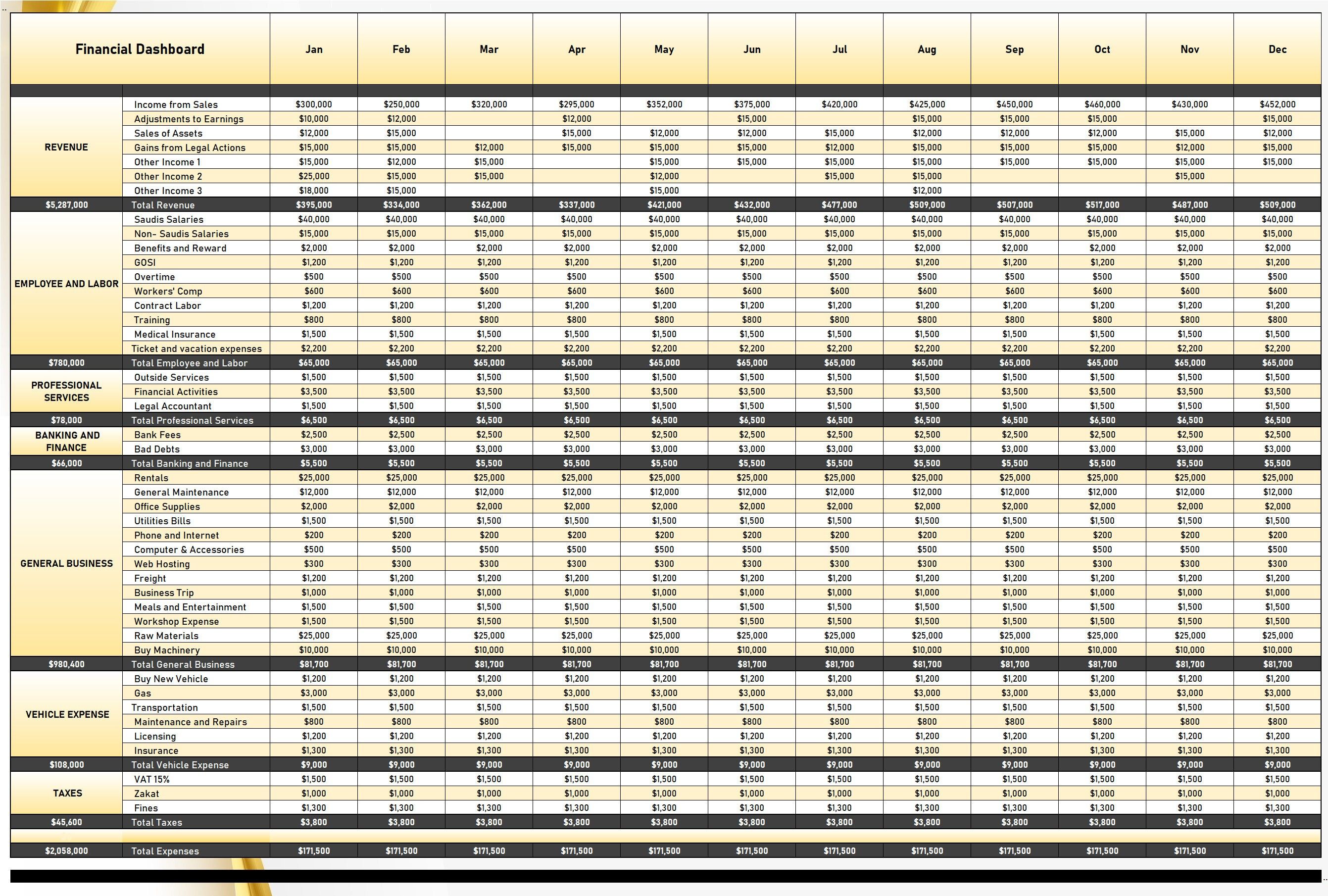}
\end{minipage}\\[4pt]
\begin{minipage}[t]{0.48\linewidth}
  \vspace{0pt}\centering
  \includegraphics[width=\linewidth]{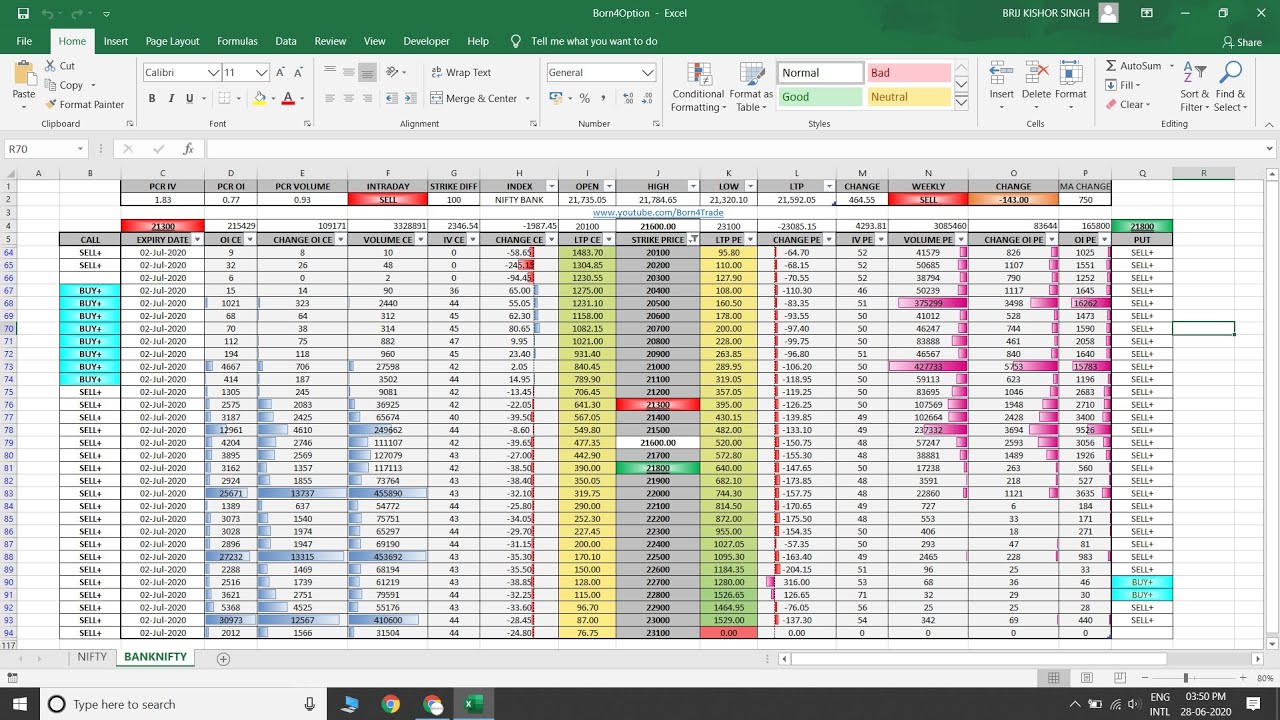}
\end{minipage}\hfill
\begin{minipage}[t]{0.48\linewidth}
  \vspace{0pt}\centering
  \includegraphics[width=\linewidth]{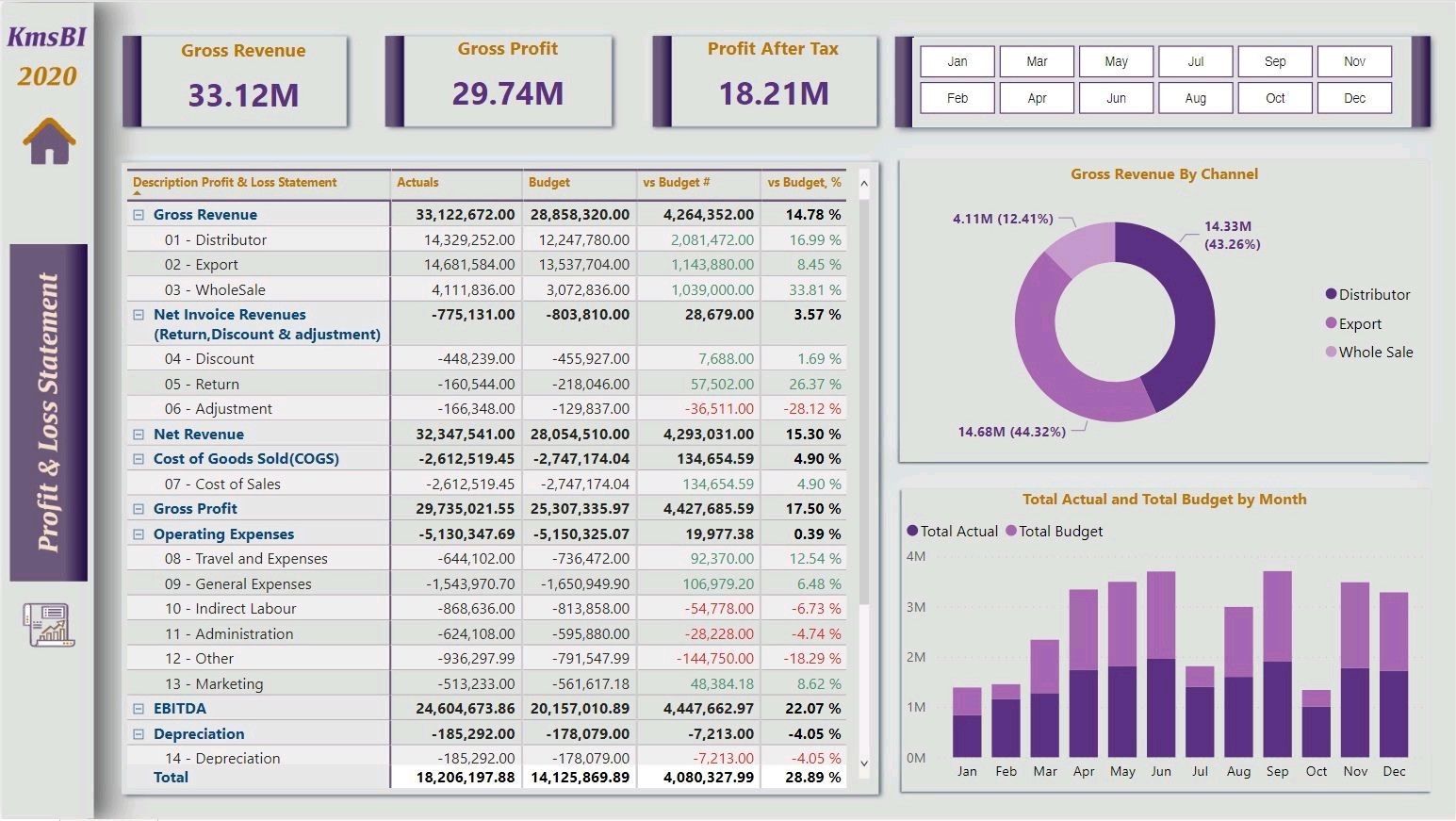}
\end{minipage}\\[3pt]
\captionof{figure}{\textbf{Finance \& Accounting}}
\label{fig:domain-finance}

\vspace{10pt}

\end{figure*}

\begin{figure*}[p]
\centering
\setlength{\fboxsep}{0pt}

\begin{minipage}[b]{0.46\linewidth}
  \centering
  \includegraphics[trim=0 2cm 0 2cm,clip,width=\linewidth]{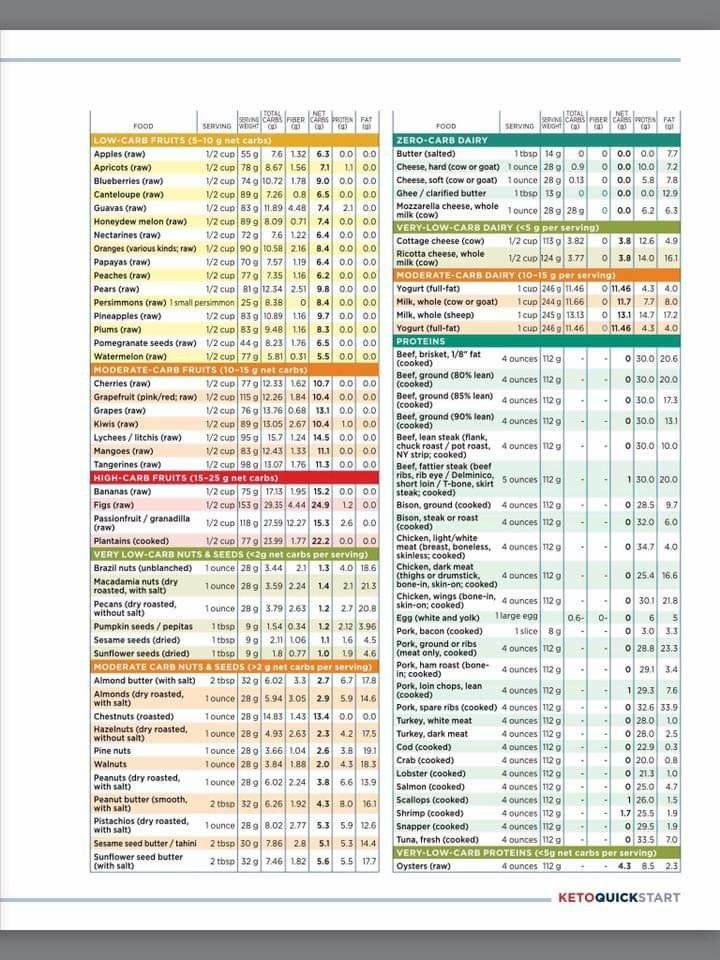}
\end{minipage}\hfill
\begin{minipage}[b]{0.50\linewidth}
  \centering
  \includegraphics[width=\linewidth]{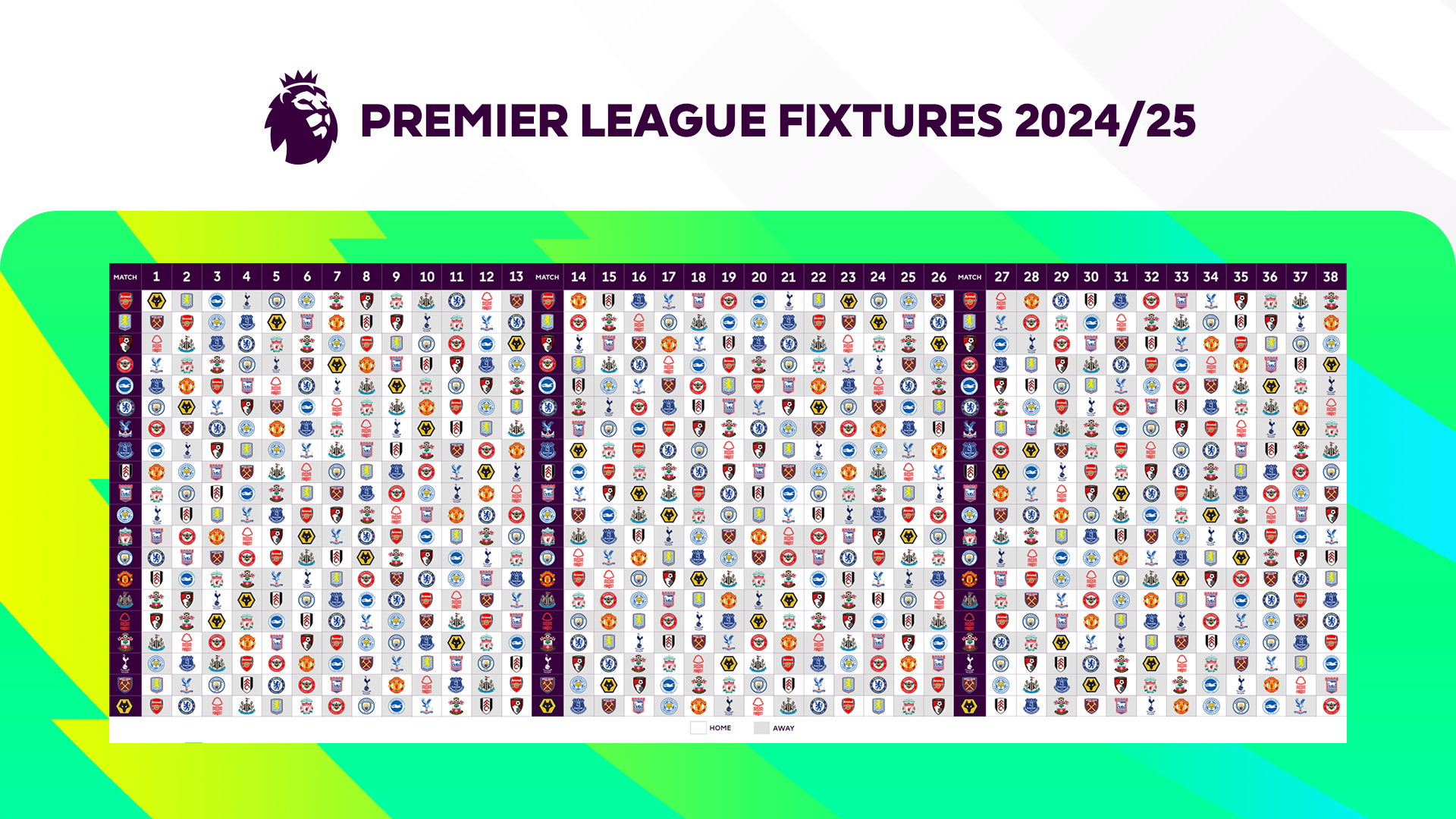}\\[4pt]
  \includegraphics[width=\linewidth]{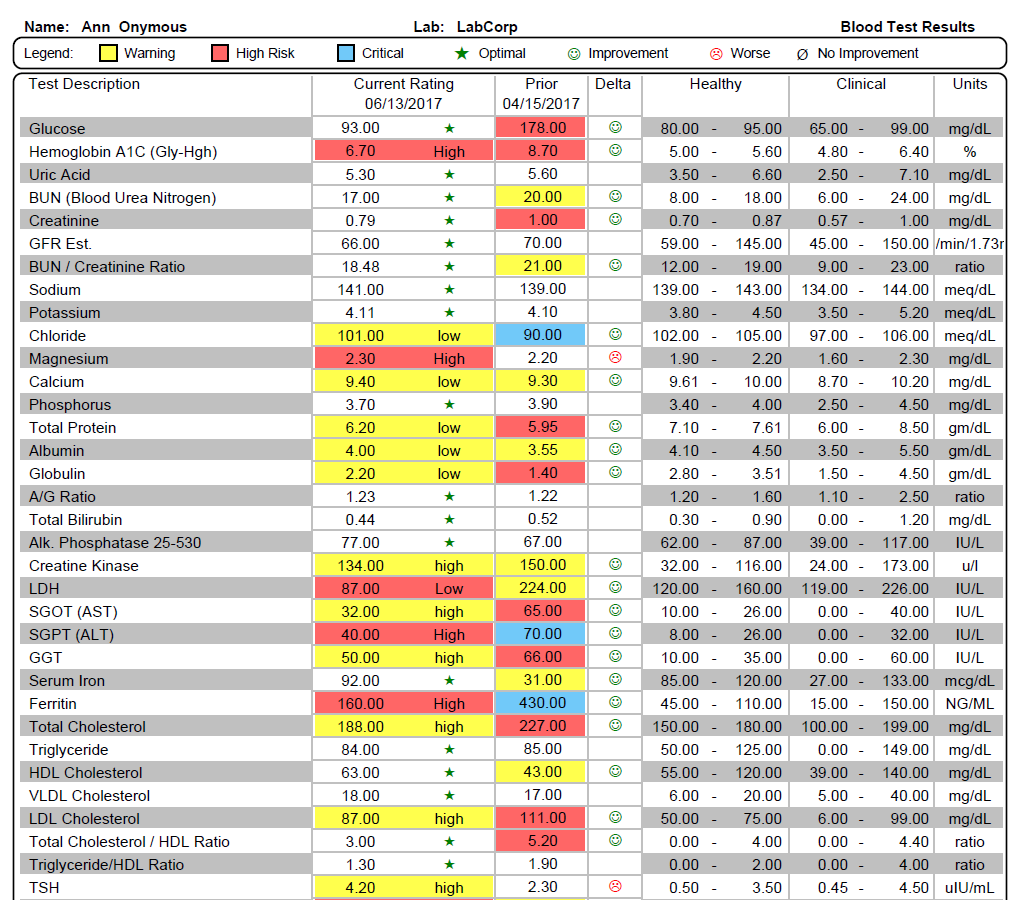}
\end{minipage}\\[3pt]
\captionof{figure}{\textbf{Sports \& Health}}
\label{fig:domain-sports}

\vspace{10pt}

\begin{minipage}[t]{0.48\linewidth}
  \vspace{0pt}\centering
  \includegraphics[width=\linewidth]{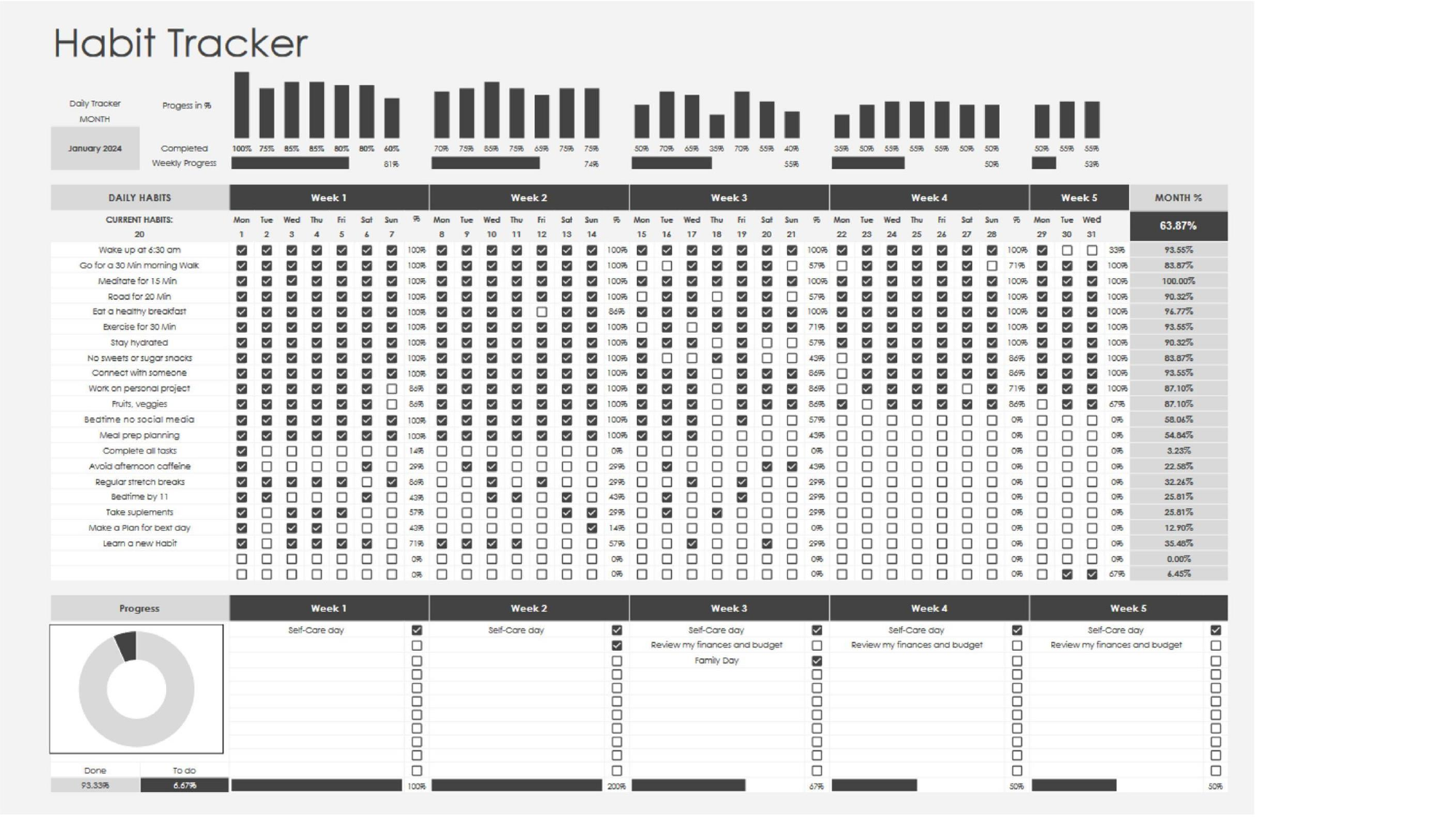}
\end{minipage}\hfill
\begin{minipage}[t]{0.48\linewidth}
  \vspace{0pt}\centering
  \includegraphics[width=\linewidth]{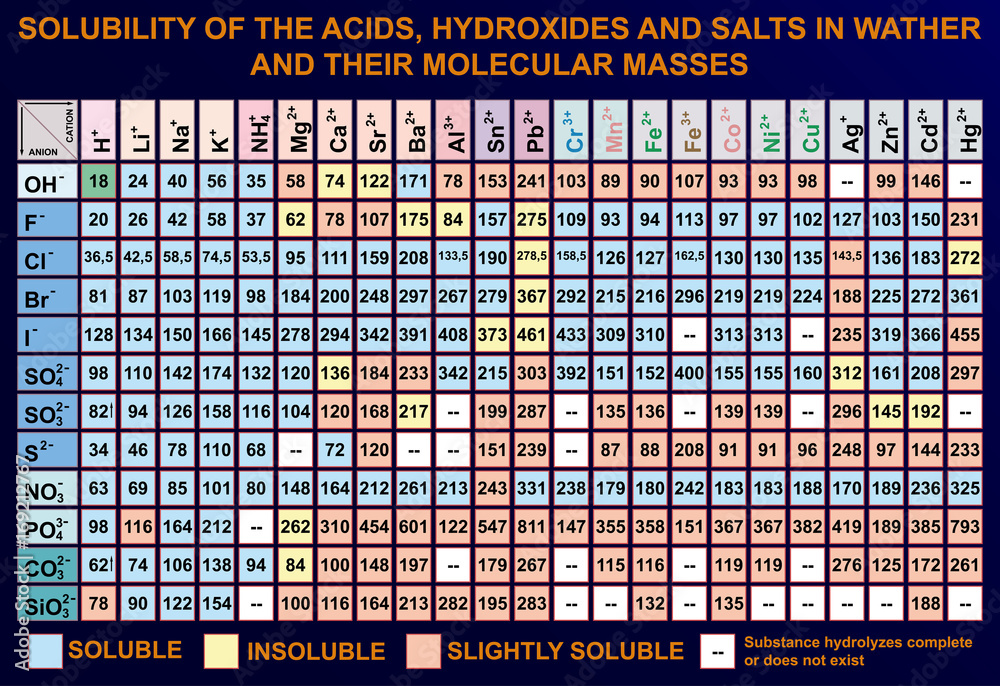}
\end{minipage}\\[4pt]
\begin{minipage}[t]{0.48\linewidth}
  \vspace{0pt}\centering
  \includegraphics[width=0.92\linewidth]{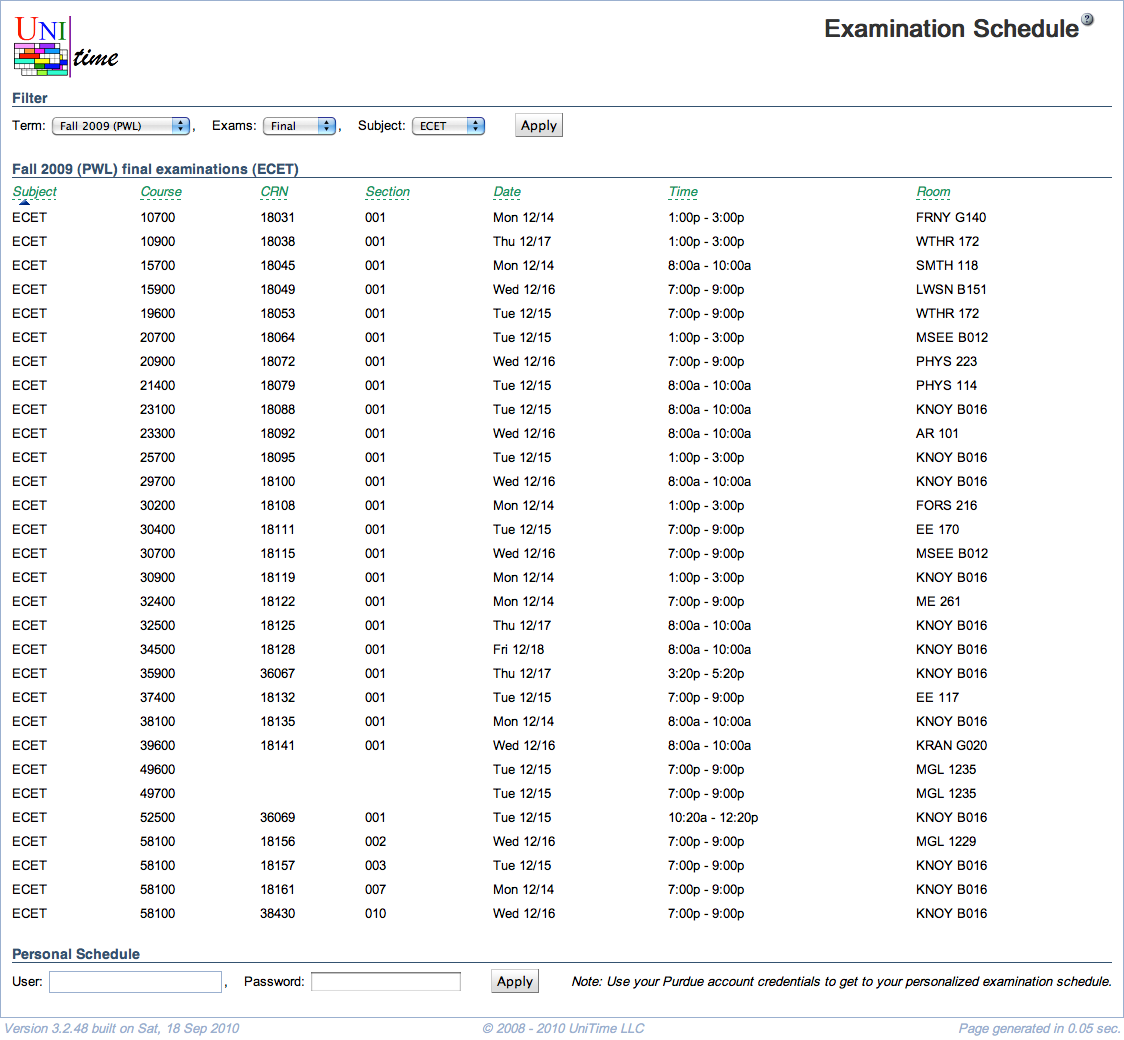}
\end{minipage}\hfill
\begin{minipage}[t]{0.48\linewidth}
  \vspace{0pt}\centering
  \includegraphics[width=\linewidth]{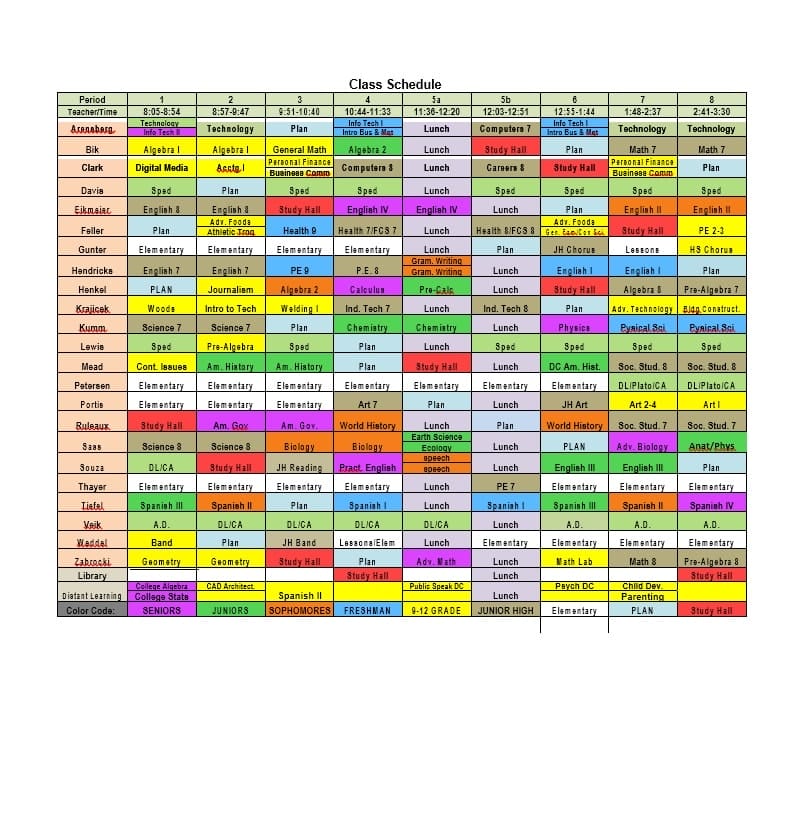}
\end{minipage}\\[3pt]
\captionof{figure}{\textbf{Education \& Science}}
\label{fig:domain-education}

\end{figure*}

\begin{figure*}[p]
\centering
\setlength{\fboxsep}{0pt}

\begin{minipage}[t]{0.46\linewidth}
  \vspace{0pt}\centering
  \includegraphics[width=\linewidth]{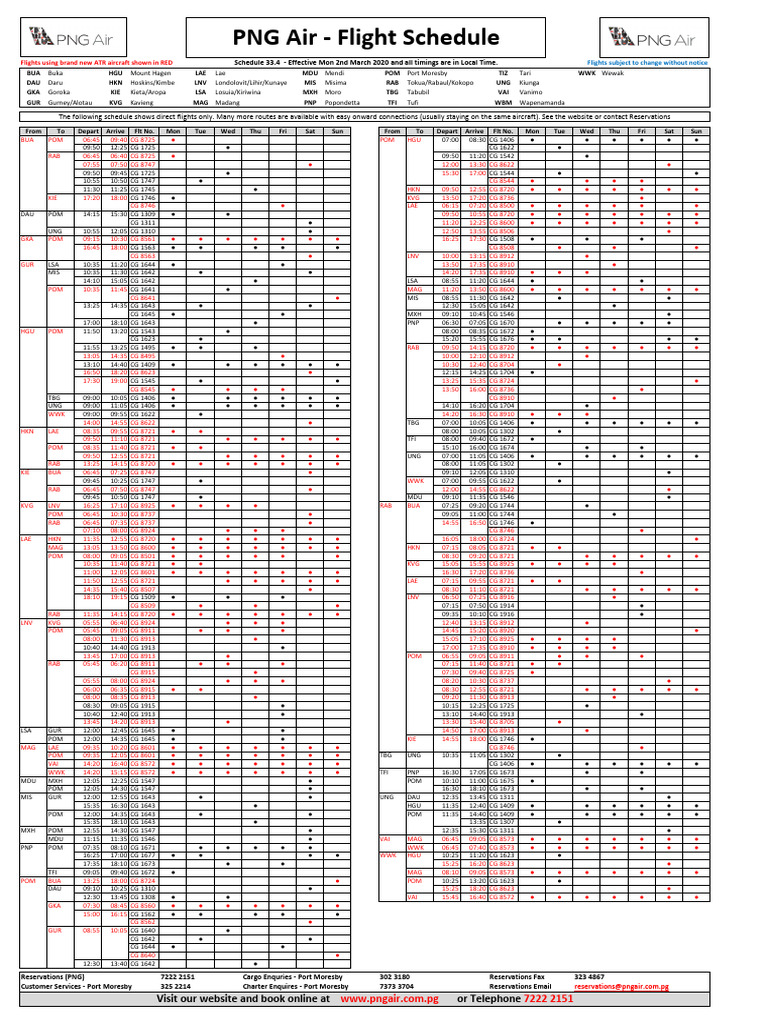}
\end{minipage}\hfill
\begin{minipage}[t]{0.50\linewidth}
  \vspace{0pt}\centering
  \includegraphics[width=\linewidth]{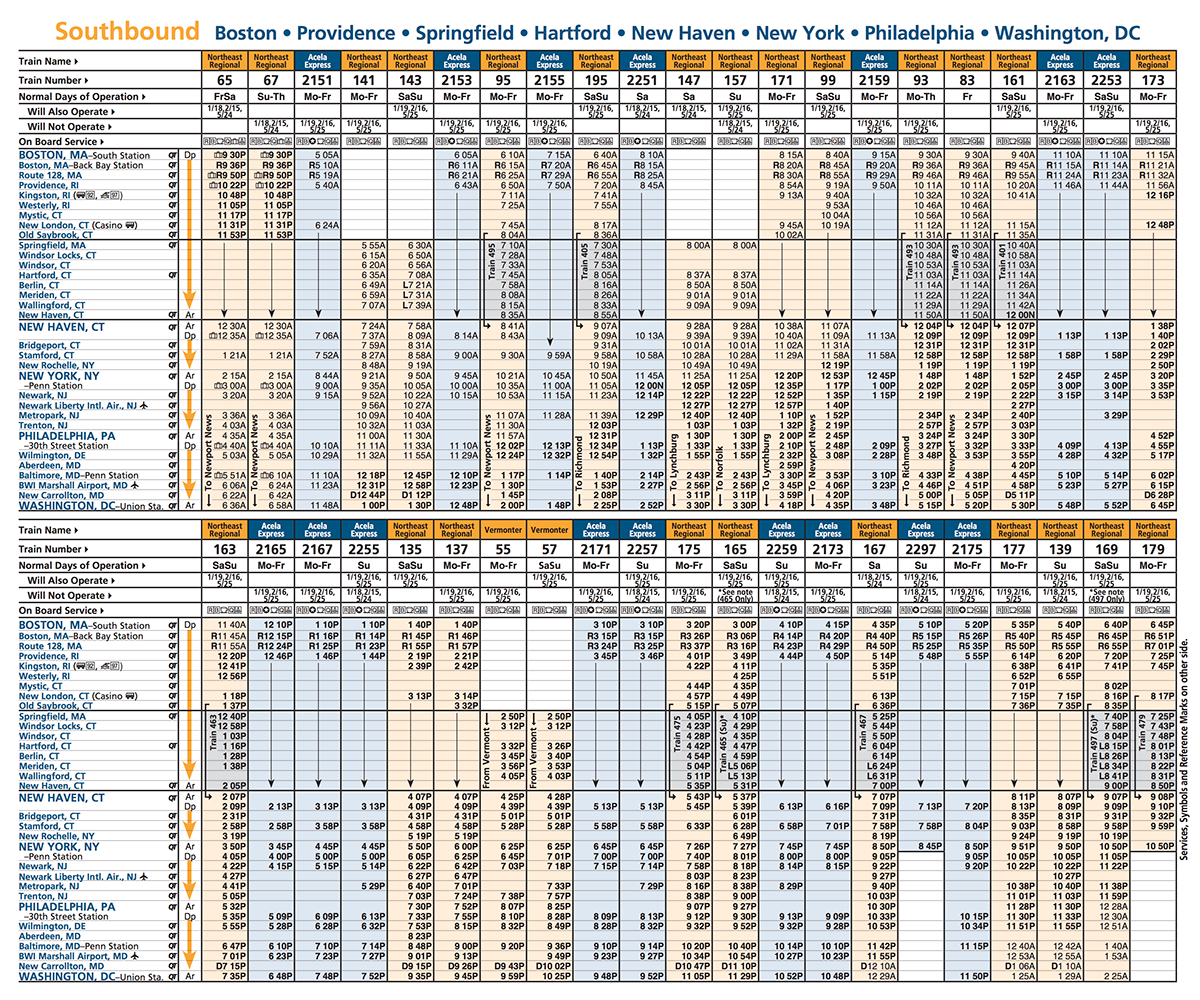}
\end{minipage}\\[3pt]
\captionof{figure}{\textbf{Transportation}}
\label{fig:domain-transportation}

\vspace{10pt}

\begin{minipage}[t]{0.46\linewidth}
  \vspace{0pt}\centering
  \includegraphics[width=\linewidth]{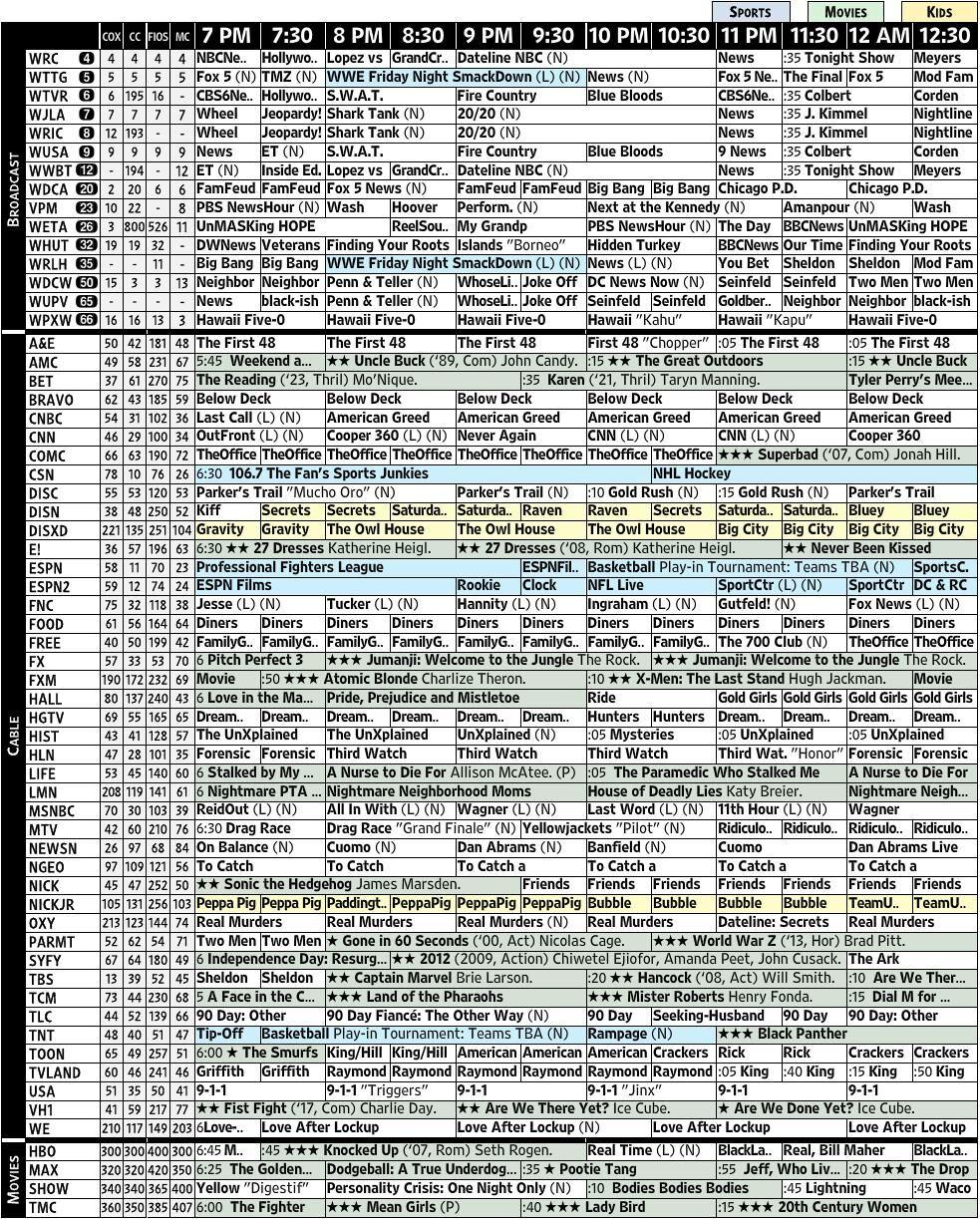}
\end{minipage}\hfill
\begin{minipage}[t]{0.50\linewidth}
  \vspace{0pt}\centering
  \includegraphics[width=\linewidth]{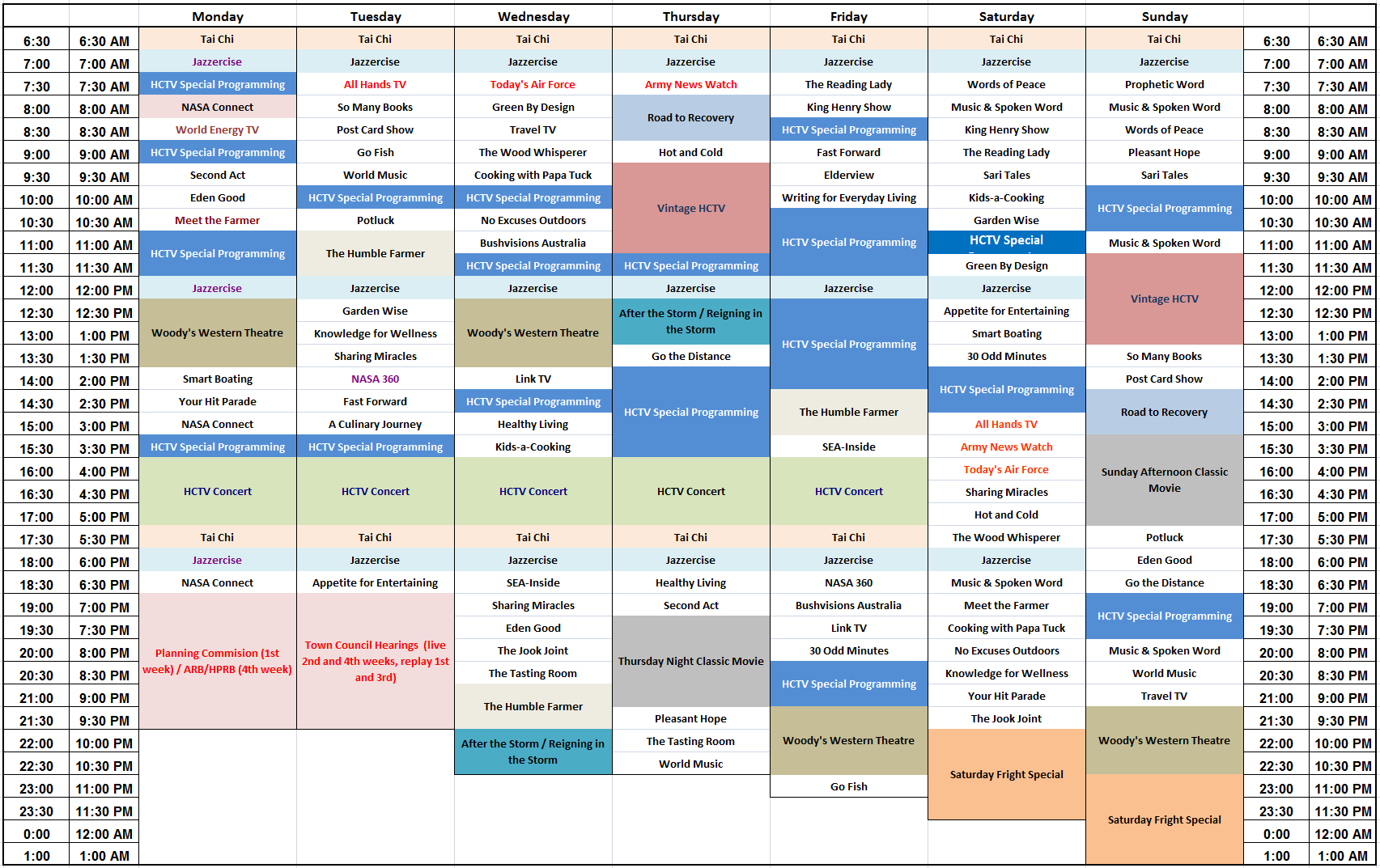}\\[4pt]
  \includegraphics[width=\linewidth]{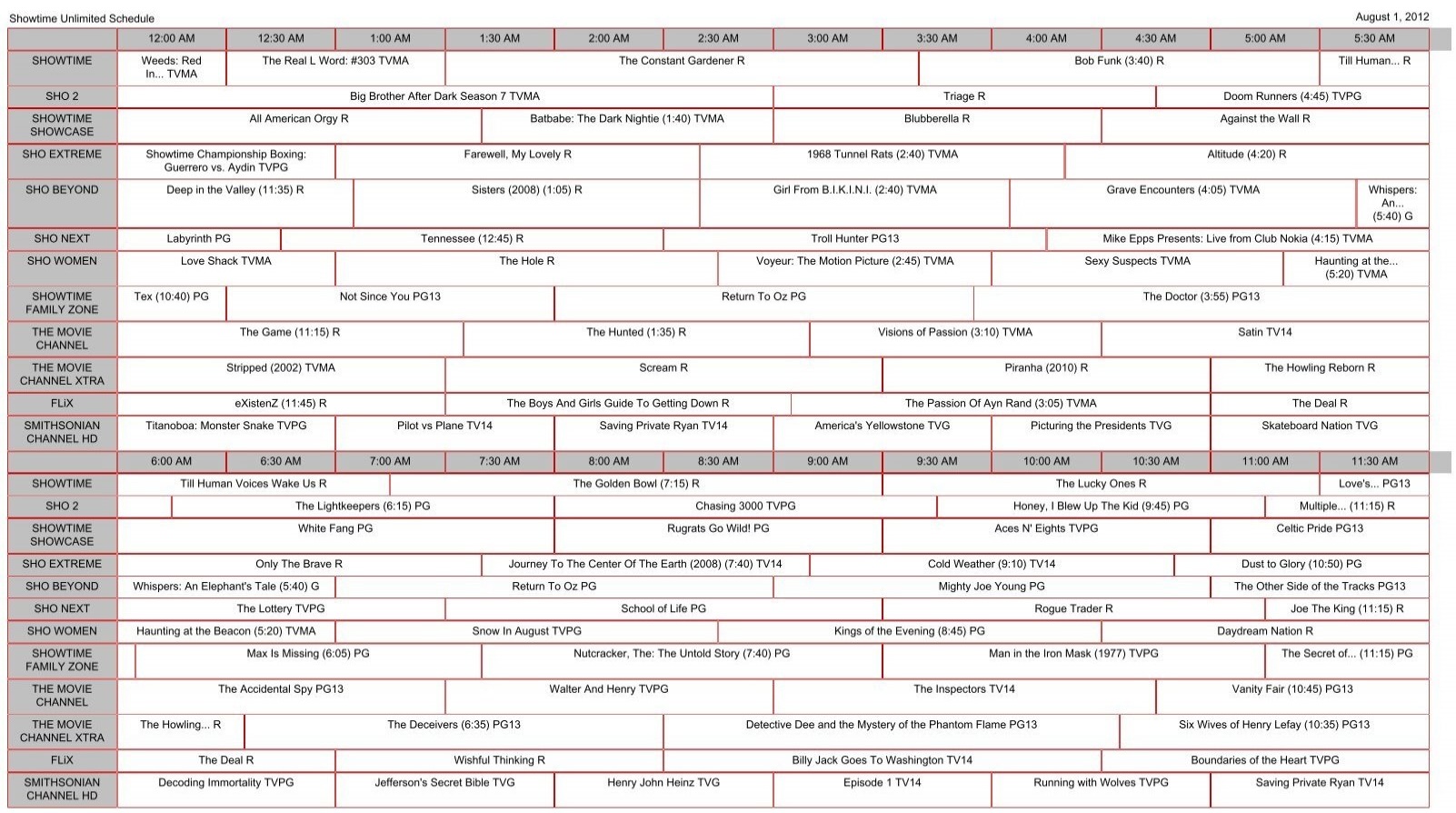}
\end{minipage}\\[3pt]
\captionof{figure}{\textbf{Society \& Media}}
\label{fig:domain-society}

\end{figure*}

Table~\ref{tab:domain_taxonomy} presents the domain taxonomy used to organize the 402 table images in WildTableBench. Each image is assigned to one of seven coarse-grained domains and further annotated with a fine-grained category label (\eg, Finance \& Accounting is subdivided into \textit{financial statement}, \textit{stock table}, \textit{price list}, among others). The taxonomy covers a wide range of real-world settings, including professional, public, scientific, and recreational contexts, providing a structured basis for analyzing model performance across different table scenarios. Figures~\ref{fig:domain-business}--\ref{fig:domain-society} show representative appendix examples (\textit{Others} omitted).

\begin{table}[t]
\centering
\scriptsize
\setlength{\tabcolsep}{3pt}
\renewcommand{\arraystretch}{1.1}
\caption{Image domain taxonomy with fine-grained category labels and dataset statistics.}
\label{tab:domain_taxonomy}
\resizebox{\linewidth}{!}{%
\begin{tabular}{@{}lccp{8.35cm}@{}}
\toprule
\textbf{Domain} & \textbf{Count} & \textbf{Proportion (\%)} & \textbf{Subcategories} \\
\midrule
Business \& Management
& 106
& 26.4
& \begin{tabular}[t]{@{}p{0.48\linewidth}@{\hspace{0.04\linewidth}}p{0.48\linewidth}@{}} $\bullet$ \textit{employee roster} & $\bullet$ \textit{project timeline} \\ \end{tabular} \\
\midrule
Finance \& Accounting  & 70 & 17.4 & \begin{tabular}[t]{@{}p{0.48\linewidth}@{\hspace{0.04\linewidth}}p{0.48\linewidth}@{}} $\bullet$ \textit{financial statement} & $\bullet$ \textit{stock table} \\ $\bullet$ \textit{auction results} & $\bullet$ \textit{price list} \\ $\bullet$ \textit{budget table} & $\bullet$ \textit{loan rates} \\ \end{tabular} \\
\midrule
Sports \& Health       & 66 & 16.4 & \begin{tabular}[t]{@{}p{0.48\linewidth}@{\hspace{0.04\linewidth}}p{0.48\linewidth}@{}} $\bullet$ \textit{match schedule} & $\bullet$ \textit{sports standings} \\ $\bullet$ \textit{race results} & $\bullet$ \textit{leaderboard} \\ $\bullet$ \textit{player statistics} & $\bullet$ \textit{team roster} \\ $\bullet$ \textit{nutrition label} & $\bullet$ \textit{medical report} \\ $\bullet$ \textit{gym class schedule} & \\ \end{tabular} \\
\midrule
Education \& Science   & 65 & 16.2 & \begin{tabular}[t]{@{}p{0.48\linewidth}@{\hspace{0.04\linewidth}}p{0.48\linewidth}@{}} $\bullet$ \textit{academic transcript} & $\bullet$ \textit{class schedule} \\ $\bullet$ \textit{exam schedule} & $\bullet$ \textit{attendance record} \\ $\bullet$ \textit{chemical data} & $\bullet$ \textit{bom materials} \\ \end{tabular} \\
\midrule
Transportation         & 43 & 10.7 & \begin{tabular}[t]{@{}p{0.48\linewidth}@{\hspace{0.04\linewidth}}p{0.48\linewidth}@{}} $\bullet$ \textit{transit timetable} & $\bullet$ \textit{ferry schedule} \\ $\bullet$ \textit{flight schedule} & $\bullet$ \textit{shipping rates} \\ \end{tabular} \\
\midrule
Society \& Media       & 35 &  8.7 & \begin{tabular}[t]{@{}p{0.48\linewidth}@{\hspace{0.04\linewidth}}p{0.48\linewidth}@{}} $\bullet$ \textit{tv schedule} & $\bullet$ \textit{election results} \\ $\bullet$ \textit{lottery results} & $\bullet$ \textit{movie showtime} \\ $\bullet$ \textit{survey results} & $\bullet$ \textit{comparison table} \\ \end{tabular} \\
\midrule
Others                 & 17 &  4.2 & \begin{tabular}[t]{@{}p{0.48\linewidth}@{\hspace{0.04\linewidth}}p{0.48\linewidth}@{}} $\bullet$ \textit{tide table} & --- \\ \end{tabular} \\
\midrule
\textbf{Total}         & \textbf{402} & \textbf{100.0} & \\
\bottomrule
\end{tabular}
\vspace{-2pt}%
}
\end{table}

\section{Implementation Details}
\label{app:impl}

All API-accessed models are queried through the OpenRouter unified endpoint, with the exception of Seed-2.0-Pro, which is accessed via a separate API. All requests use a per-request timeout of 120 seconds; failed requests due to transient errors are retried up to five times with exponential backoff; all questions are guaranteed to have a valid model response in the final evaluation. Locally deployed Qwen3-VL models are run on NVIDIA GPUs.
\section{Evaluation Prompts}
\label{app:prompts}

We use a two-step evaluation pipeline. First, each evaluated model receives the table image together with a task prompt and produces a free-form response. Second, GPT-5.2 serves as an LLM judge to determine correctness by comparing the model output against the ground-truth answer already annotated in the benchmark under explicit, format-insensitive decision rules. This appendix reports the exact prompts used for response generation and answer verification.

This design separates answer generation from answer verification. The generation prompt is standardized across models to reduce prompt-induced variance, whereas the judging prompts are specialized for the answer format required by different task types. In particular, Transcription questions require exact recovery of multiple table entries and therefore use a stricter judging template than the rest of the benchmark. Accordingly, the two-step pipeline is implemented with one response-generation prompt and two answer-verification templates, shown in Box~\ref{box:model-eval}, Box~\ref{box:judge-general}, and Box~\ref{box:judge-transcription}.

\subsection{Answer Generation Stage}
\label{app:response_prompt}
To elicit structured responses from vision-language models, we instruct each model to solve the question step-by-step and to end with a clearly marked final answer (Box~\ref{box:model-eval}). This prompt serves two purposes. First, the explicit \texttt{Final answer:} marker enables deterministic extraction of the answer span for downstream evaluation. Second, the intermediate reasoning provides additional context when the extracted final answer is missing, incomplete, or ambiguously phrased. We keep the prompt intentionally short so that it standardizes answer formatting without injecting task-specific hints beyond the question itself. The \texttt{\{question\}} placeholder is replaced with the question text for each instance.

\refstepcounter{promptboxctr}\label{box:model-eval} 
\begin{evalpromptbox}{Box~\thepromptboxctr. Prompt Template for Model Answer Generation}
\label{prompt:model-eval}
You are given a question that refers to an image. Solve this question step-by-step.
Provide your final answer in the following format: 
\begin{itemize}
    \item Final answer: 
    \begin{itemize}
        \item {
            [your concise answer]
        }
    \end{itemize}
    \item \#\#\# Question
    \begin{itemize}
    \item \{question\}
    \end{itemize}
\end{itemize}
\end{evalpromptbox}
\subsection{Automatic Answer Verification: General Questions}
\label{app:general_prompt}
Following prior work~\citep{zheng2023judging,chen2024mllm}, we use GPT-5.2 as an LLM judge to evaluate answer correctness. The general judging prompt shown in Box~\ref{box:judge-general} is used for all question types except Transcription (C1-T). The judge receives four inputs—the question, the ground truth answer, the model's extracted final answer, and the model's full response—and is instructed to decide from the extracted final answer whenever possible, consulting the full response only as a fallback. The decision rules are strict about semantic content while tolerant to superficial formatting variation: numerical answers are matched by value regardless of currency symbols or separators, textual answers are compared case-insensitively, and boolean questions accept equivalent affirmative or negative forms.

\refstepcounter{promptboxctr}\label{box:judge-general}
\begin{judgepromptbox}{Box~\thepromptboxctr. Prompt Template for LLM-based General Answer Verification}
\label{prompt:judge-general}
You are a STRICT judge evaluating if a model's answer matches the ground truth.
Question: 
\begin{itemize}
    \item \{question\}
\end{itemize}
Ground Truth Answer: 
\begin{itemize}
    \item \{ground\_truth\}
\end{itemize}

\textbf{Model's Final Answer:} (what the model gave after "Final answer:")
\begin{itemize}
    \item \{model\_final\_answer\}
\end{itemize}

\textbf{Model's Full Response:} (reasoning + answer, for reference when final answer is unclear)
\begin{itemize}
    \item \{model\_full\_response\}
\end{itemize}

\textbf{Instructions:} First judge correctness using \textbf{Model's Final Answer} only. If the final answer clearly matches (or clearly does not match) the ground truth, base your decision on that. Only when the final answer is ambiguous, missing, or you cannot tell---use \textbf{Model's Full Response} (reasoning) to decide. Same value = CORRECT; ignore formatting (bold, spaces, extra words).

STRICT Evaluation Rules:
\begin{enumerate}
    \item {
        \textbf{Numerical/currency answers}: If the model's stated value equals the ground truth (same number), answer CORRECT. Format does not matter.
        \begin{itemize}
            \item ``\$151,830'' = ``151830'' = ``The revenue was \$151,830.'' = ``\textbf{\$151,830}'' (CORRECT)
            \item Only if the value is different (e.g.\ \$150,000 vs \$151,830) answer INCORRECT.
        \end{itemize}
    }
    \item {
        \textbf{Text answers}: Must be semantically equivalent. 
        \begin{itemize}
            \item Case differences OK: ``Apple'' = ``apple'' (CORRECT)
            \item Different content is INCORRECT: ``Product A'' $\neq$ ``Product B''.
        \end{itemize}
    }
    \item {
        \textbf{Boolean/Yes-No answers}: Semantically equivalent forms are OK.
        \begin{itemize}
            \item ``Yes'' = ``True'' = ``Correct'' (CORRECT)
            \item ``No'' = ``False'' = ``Incorrect'' (CORRECT)
        \end{itemize}
    }
   
\end{enumerate}
\end{judgepromptbox}

\subsection{Automatic Answer Verification: Transcription Questions}
\label{app:transcription_prompt}
For Transcription questions (C1-T), which require models to reproduce all cell values in the correct order, we use the stricter judging prompt shown in Box~\ref{box:judge-transcription}. Unlike short-answer or reasoning questions, transcription demands completeness and order preservation, so the prompt rejects omissions, insertions, and permutations. It remains tolerant to presentation-level variation such as different separators (\eg, commas, pipes, newlines), equivalent number formats (\eg, ``0.10'' and ``0.1''), and case differences.

\refstepcounter{promptboxctr}\label{box:judge-transcription}
\begin{judgetranscriptionbox}{Box~\thepromptboxctr. Prompt Template for LLM-based Transcription Verification}
\label{prompt:judge-transcription}
You are a STRICT judge evaluating if a model's transcription matches the ground truth.

Question: 
\begin{itemize}
    \item \{question\}
\end{itemize}
Ground Truth Answer: 
\begin{itemize}
    \item \{ground\_truth\}
\end{itemize}

\textbf{Model's Final Answer:} (what the model gave after ``Answer:'')\\
\begin{itemize}
    \item \{model\_final\_answer\}
\end{itemize}

\textbf{Model's Full Response:} (reasoning + answer, for reference when final answer is unclear)
\begin{itemize}
    \item \{model\_full\_response\}
\end{itemize}

\textbf{Instructions:} First judge correctness using \textbf{Model's Final Answer} only. Follow this logic:
\begin{enumerate}
    \item If the final answer clearly matches the ground truth $\to$ CORRECT.
    \item If the final answer contains wrong values $\to$ INCORRECT immediately (do NOT check full response).
    \item If the final answer appears \textbf{partial} (fewer values than expected, seems truncated, but the values present are all correct) $\to$ use \textbf{Model's Full Response} to find the complete answer and judge based on that.
    \item If the final answer is missing or ambiguous $\to$ use \textbf{Model's Full Response} to decide.
\end{enumerate}

STRICT Rules for Transcription:
\begin{enumerate}
  \item {\textbf{All values must be present} --- missing even one value = INCORRECT.}
  \item \textbf{Order must match} --- if the question specifies top-to-bottom or left-to-right, the order must be correct.
  \item \textbf{No extra values} --- values not in the ground truth = INCORRECT.
  \item \textbf{Separators are flexible} --- commas / pipes / newlines / spaces are interchangeable.
  \item \textbf{Number formatting is flexible} --- ``0.10'' = ``0.1'', ``1.00'' = ``1'' (CORRECT).
  \item \textbf{Case is flexible} --- ``apple'' = ``Apple'' (CORRECT).
  \item \textbf{Minor punctuation differences are OK} --- trailing period, extra spaces (CORRECT).
\end{enumerate}

Is the model's transcription CORRECT or INCORRECT?

Return ONLY one word: ``CORRECT'' or ``INCORRECT''.
\end{judgetranscriptionbox}

\section{Cell-Position Sensitivity: Full Results}
\label{app:needle_full}

\paragraph{Experimental Setup.}
We construct a cell-retrieval evaluation set of 2{,}489 needles from 50 real-world spreadsheet images, filtered to the common subset evaluated by all models.
Each needle asks the model to report the exact value at a given cell address (\eg, ``What is the exact value in cell B7?''); responses are graded by string normalisation with numeric tolerance.
Row depth and column depth are each divided into 10 equal bands (0--10\%, 10--20\%, \ldots, 90--100\%), forming a 10$\times$10 position grid, and accuracy is computed per bin.

Figure~\ref{fig:needle_heatmap_appendix} shows cell-retrieval accuracy heatmaps for eight additional evaluated models (two proprietary and six open-source).
Each subplot shows per-model min--max normalised accuracy across the 10$\times$10 row $\times$ column grid.

\begin{figure}[H]
  \centering
  \pgfplotsset{
    colormap={RdBu}{
      rgb255(0)=(0,0,255)
      rgb255(250)=(128,128,255)
      rgb255(500)=(255,255,255)
      rgb255(750)=(255,128,128)
      rgb255(1000)=(255,0,0)
    },
    /pgfplots/heatmap base/.style={
      view={0}{90},
      width=0.22\textwidth,
      height=0.22\textwidth,
      colormap name=RdBu,
      point meta min=0, point meta max=1,
      y dir=reverse,
      xmin=0, xmax=9, ymin=0, ymax=9,
      enlargelimits=false,
      xtick={0,2.25,4.5,6.75},
      xticklabels={0,25,50,75},
      ytick={0,2.25,4.5,6.75},
      yticklabels={0,25,50,75},
      ztick=\empty, zlabel={},
      tick label style={font=\fontsize{5}{6}\selectfont},
      title style={font=\scriptsize\bfseries, at={(0.5,1.0)}, anchor=south, yshift=-8pt, align=center},
      xlabel style={font=\scriptsize, at={(0.5,0)}, anchor=north, yshift=-6pt},
      ylabel style={font=\scriptsize, at={(0,0.5)}, anchor=south, yshift=8pt, xshift=0pt},
      axis on top,
      axis line style={thin, black!50},
      tick style={thin, black!40},
      scale only axis,
      3d box=complete,
      after end axis/.code={
        \draw[densely dotted, black!60, line width=0.8pt]
          (rel axis cs:0.5,0) -- (rel axis cs:0.5,1);
        \draw[densely dotted, black!60, line width=0.8pt]
          (rel axis cs:0,0.5) -- (rel axis cs:1,0.5);
      },
    }
  }

  \begin{tikzpicture}

  \begin{axis}[
    heatmap base,
    name=plotA,
    at={(0,0)}, anchor=north west,
    title={Qwen3-VL-2B-I (15.0\%)},
    xlabel={Column Depth (\%)},
    ylabel={Row Depth (\%)},
  ]
  \addplot3[surf, shader=interp, mesh/cols=10] table[meta=C] {
x y C
0 0 1.000
1 0 0.652
2 0 0.409
3 0 0.428
4 0 0.410
5 0 0.425
6 0 0.301
7 0 0.387
8 0 0.586
9 0 0.285
0 1 0.733
1 1 0.431
2 1 0.263
3 1 0.217
4 1 0.288
5 1 0.491
6 1 0.460
7 1 0.414
8 1 0.516
9 1 0.285
0 2 0.628
1 2 0.362
2 2 0.205
3 2 0.139
4 2 0.185
5 2 0.358
6 2 0.399
7 2 0.331
8 2 0.394
9 2 0.263
0 3 0.655
1 3 0.381
2 3 0.252
3 3 0.180
4 3 0.160
5 3 0.252
6 3 0.214
7 3 0.205
8 3 0.318
9 3 0.167
0 4 0.592
1 4 0.388
2 4 0.302
3 4 0.165
4 4 0.119
5 4 0.202
6 4 0.127
7 4 0.192
8 4 0.333
9 4 0.253
0 5 0.552
1 5 0.387
2 5 0.275
3 5 0.121
4 5 0.040
5 5 0.096
6 5 0.099
7 5 0.168
8 5 0.290
9 5 0.378
0 6 0.477
1 6 0.356
2 6 0.240
3 6 0.087
4 6 0.016
5 6 0.080
6 6 0.102
7 6 0.100
8 6 0.197
9 6 0.252
0 7 0.397
1 7 0.321
2 7 0.161
3 7 0.052
4 7 0.000
5 7 0.091
6 7 0.161
7 7 0.143
8 7 0.220
9 7 0.184
0 8 0.508
1 8 0.412
2 8 0.180
3 8 0.119
4 8 0.103
5 8 0.192
6 8 0.260
7 8 0.205
8 8 0.243
9 8 0.142
0 9 0.476
1 9 0.340
2 9 0.193
3 9 0.209
4 9 0.282
5 9 0.297
6 9 0.244
7 9 0.221
8 9 0.366
9 9 0.350
  };
  \end{axis}

  \begin{axis}[
    heatmap base,
    name=plotB,
    at={(plotA.north east)}, anchor=north west, xshift=2pt,
    title={Qwen3-VL-8B-I (24.6\%)},
    xlabel={Column Depth (\%)},
    yticklabels={},
  ]
  \addplot3[surf, shader=interp, mesh/cols=10] table[meta=C] {
x y C
0 0 0.849
1 0 0.579
2 0 0.295
3 0 0.246
4 0 0.293
5 0 0.313
6 0 0.257
7 0 0.246
8 0 0.212
9 0 0.029
0 1 0.703
1 1 0.472
2 1 0.282
3 1 0.223
4 1 0.234
5 1 0.308
6 1 0.331
7 1 0.397
8 1 0.357
9 1 0.112
0 2 0.544
1 2 0.423
2 2 0.270
3 2 0.217
4 2 0.220
5 2 0.280
6 2 0.270
7 2 0.347
8 2 0.365
9 2 0.164
0 3 0.474
1 3 0.442
2 3 0.306
3 3 0.253
4 3 0.266
5 3 0.299
6 3 0.189
7 3 0.219
8 3 0.282
9 3 0.208
0 4 0.637
1 4 0.462
2 4 0.257
3 4 0.153
4 4 0.167
5 4 0.199
6 4 0.131
7 4 0.185
8 4 0.224
9 4 0.249
0 5 0.709
1 5 0.429
2 5 0.206
3 5 0.076
4 5 0.045
5 5 0.033
6 5 0.060
7 5 0.158
8 5 0.132
9 5 0.207
0 6 0.572
1 6 0.334
2 6 0.188
3 6 0.114
4 6 0.029
5 6 0.000
6 6 0.095
7 6 0.218
8 6 0.116
9 6 0.112
0 7 0.519
1 7 0.345
2 7 0.175
3 7 0.120
4 7 0.028
5 7 0.041
6 7 0.185
7 7 0.299
8 7 0.234
9 7 0.190
0 8 0.707
1 8 0.545
2 8 0.170
3 8 0.033
4 8 0.009
5 8 0.124
6 8 0.283
7 8 0.299
8 8 0.255
9 8 0.207
0 9 1.000
1 9 0.768
2 9 0.251
3 9 0.027
4 9 0.061
5 9 0.211
6 9 0.321
7 9 0.278
8 9 0.188
9 9 0.098
  };
  \end{axis}

  \begin{axis}[
    heatmap base,
    name=plotC,
    at={(plotB.north east)}, anchor=north west, xshift=2pt,
    title={Qwen3-VL-4B-I (26.5\%)},
    xlabel={Column Depth (\%)},
    yticklabels={},
  ]
  \addplot3[surf, shader=interp, mesh/cols=10] table[meta=C] {
x y C
0 0 0.900
1 0 0.692
2 0 0.531
3 0 0.616
4 0 0.528
5 0 0.429
6 0 0.451
7 0 0.523
8 0 0.488
9 0 0.312
0 1 0.725
1 1 0.601
2 1 0.498
3 1 0.445
4 1 0.359
5 1 0.333
6 1 0.406
7 1 0.474
8 1 0.409
9 1 0.272
0 2 0.606
1 2 0.513
2 2 0.475
3 2 0.419
4 2 0.337
5 2 0.280
6 2 0.321
7 2 0.359
8 2 0.274
9 2 0.130
0 3 0.795
1 3 0.544
2 3 0.485
3 3 0.503
4 3 0.464
5 3 0.369
6 3 0.314
7 3 0.299
8 3 0.235
9 3 0.084
0 4 1.000
1 4 0.614
2 4 0.459
3 4 0.442
4 4 0.445
5 4 0.383
6 4 0.379
7 4 0.459
8 4 0.398
9 4 0.263
0 5 0.911
1 5 0.564
2 5 0.346
3 5 0.287
4 5 0.277
5 5 0.217
6 5 0.244
7 5 0.416
8 5 0.352
9 5 0.255
0 6 0.794
1 6 0.456
2 6 0.250
3 6 0.187
4 6 0.184
5 6 0.164
6 6 0.226
7 6 0.345
8 6 0.211
9 6 0.059
0 7 0.735
1 7 0.391
2 7 0.181
3 7 0.141
4 7 0.163
5 7 0.204
6 7 0.300
7 7 0.348
8 7 0.167
9 7 0.000
0 8 0.804
1 8 0.494
2 8 0.183
3 8 0.121
4 8 0.184
5 8 0.189
6 8 0.247
7 8 0.349
8 8 0.192
9 8 0.012
0 9 0.813
1 9 0.604
2 9 0.258
3 9 0.136
4 9 0.173
5 9 0.129
6 9 0.205
7 9 0.331
8 9 0.226
9 9 0.111
  };
  \end{axis}

  \begin{axis}[
    heatmap base,
    name=plotD,
    at={(plotC.north east)}, anchor=north west, xshift=2pt,
    title={GPT-o3 (29.4\%)},
    xlabel={Column Depth (\%)},
    yticklabels={},
  ]
  \addplot3[surf, shader=interp, mesh/cols=10] table[meta=C] {
x y C
0 0 1.000
1 0 0.836
2 0 0.838
3 0 0.953
4 0 0.962
5 0 0.960
6 0 0.944
7 0 0.935
8 0 0.675
9 0 0.580
0 1 0.994
1 1 0.859
2 1 0.714
3 1 0.885
4 1 0.972
5 1 0.975
6 1 0.866
7 1 0.838
8 1 0.663
9 1 0.527
0 2 0.736
1 2 0.854
2 2 0.652
3 2 0.761
4 2 0.840
5 2 0.878
6 2 0.748
7 2 0.689
8 2 0.553
9 2 0.465
0 3 0.719
1 3 0.761
2 3 0.604
3 3 0.747
4 3 0.816
5 3 0.854
6 3 0.666
7 3 0.618
8 3 0.541
9 3 0.566
0 4 0.690
1 4 0.652
2 4 0.546
3 4 0.538
4 4 0.725
5 4 0.896
6 4 0.576
7 4 0.555
8 4 0.578
9 4 0.696
0 5 0.631
1 5 0.532
2 5 0.413
3 5 0.292
4 5 0.475
5 5 0.651
6 5 0.335
7 5 0.403
8 5 0.567
9 5 0.683
0 6 0.517
1 6 0.399
2 6 0.337
3 6 0.205
4 6 0.242
5 6 0.378
6 6 0.210
7 6 0.435
8 6 0.503
9 6 0.470
0 7 0.399
1 7 0.270
2 7 0.202
3 7 0.143
4 7 0.123
5 7 0.197
6 7 0.130
7 7 0.485
8 7 0.559
9 7 0.478
0 8 0.678
1 8 0.455
2 8 0.178
3 8 0.050
4 8 0.078
5 8 0.137
6 8 0.027
7 8 0.278
8 8 0.427
9 8 0.383
0 9 0.778
1 9 0.681
2 9 0.320
3 9 0.047
4 9 0.170
5 9 0.177
6 9 0.000
7 9 0.179
8 9 0.300
9 9 0.244
  };
  \end{axis}

  \begin{axis}[
    heatmap base,
    name=plotE,
    at={(plotA.south west)}, anchor=north west, yshift=-42pt,
    title={Qwen3-VL-235B-I (43.7\%)},
    xlabel={Column Depth (\%)},
    ylabel={Row Depth (\%)},
  ]
  \addplot3[surf, shader=interp, mesh/cols=10] table[meta=C] {
x y C
0 0 0.935
1 0 0.655
2 0 0.446
3 0 0.682
4 0 0.761
5 0 0.529
6 0 0.369
7 0 0.539
8 0 0.469
9 0 0.166
0 1 0.814
1 1 0.618
2 1 0.514
3 1 0.614
4 1 0.642
5 1 0.543
6 1 0.445
7 1 0.635
8 1 0.629
9 1 0.316
0 2 0.747
1 2 0.757
2 2 0.681
3 2 0.697
4 2 0.650
5 2 0.546
6 2 0.509
7 2 0.687
8 2 0.665
9 2 0.317
0 3 1.000
1 3 0.937
2 3 0.747
3 3 0.670
4 3 0.593
5 3 0.479
6 3 0.402
7 3 0.530
8 3 0.476
9 3 0.157
0 4 0.923
1 4 0.845
2 4 0.646
3 4 0.438
4 4 0.373
5 4 0.324
6 4 0.223
7 4 0.371
8 4 0.389
9 4 0.191
0 5 0.591
1 5 0.634
2 5 0.509
3 5 0.270
4 5 0.201
5 5 0.203
6 5 0.160
7 5 0.348
8 5 0.358
9 5 0.263
0 6 0.581
1 6 0.624
2 6 0.526
3 6 0.322
4 6 0.193
5 6 0.222
6 6 0.285
7 6 0.511
8 6 0.387
9 6 0.227
0 7 0.665
1 7 0.583
2 7 0.419
3 7 0.352
4 7 0.217
5 7 0.239
6 7 0.307
7 7 0.567
8 7 0.462
9 7 0.185
0 8 0.689
1 8 0.501
2 8 0.270
3 8 0.216
4 8 0.163
5 8 0.216
6 8 0.318
7 8 0.571
8 8 0.476
9 8 0.100
0 9 0.603
1 9 0.430
2 9 0.238
3 9 0.123
4 9 0.163
5 9 0.251
6 9 0.344
7 9 0.594
8 9 0.474
9 9 0.000
  };
  \end{axis}

  \begin{axis}[
    heatmap base,
    name=plotF,
    at={(plotE.north east)}, anchor=north west, xshift=2pt,
    title={Qwen3-VL-32B-I (47.9\%)},
    xlabel={Column Depth (\%)},
    yticklabels={},
  ]
  \addplot3[surf, shader=interp, mesh/cols=10] table[meta=C] {
x y C
0 0 0.884
1 0 0.785
2 0 0.772
3 0 0.895
4 0 0.715
5 0 0.558
6 0 0.491
7 0 0.569
8 0 0.491
9 0 0.293
0 1 0.951
1 1 0.848
2 1 0.687
3 1 0.753
4 1 0.681
5 1 0.629
6 1 0.565
7 1 0.597
8 1 0.463
9 1 0.212
0 2 1.000
1 2 0.828
2 2 0.585
3 2 0.576
4 2 0.564
5 2 0.587
6 2 0.547
7 2 0.571
8 2 0.471
9 2 0.259
0 3 0.976
1 3 0.709
2 3 0.500
3 3 0.552
4 3 0.488
5 3 0.449
6 3 0.360
7 3 0.453
8 3 0.442
9 3 0.209
0 4 0.820
1 4 0.619
2 4 0.519
3 4 0.557
4 4 0.520
5 4 0.466
6 4 0.326
7 4 0.358
8 4 0.340
9 4 0.238
0 5 0.610
1 5 0.496
2 5 0.404
3 5 0.314
4 5 0.338
5 5 0.316
6 5 0.263
7 5 0.312
8 5 0.234
9 5 0.224
0 6 0.644
1 6 0.505
2 6 0.336
3 6 0.215
4 6 0.167
5 6 0.111
6 6 0.156
7 6 0.274
8 6 0.216
9 6 0.106
0 7 0.573
1 7 0.430
2 7 0.280
3 7 0.243
4 7 0.094
5 7 0.000
6 7 0.110
7 7 0.256
8 7 0.302
9 7 0.231
0 8 0.447
1 8 0.381
2 8 0.204
3 8 0.157
4 8 0.041
5 8 0.004
6 8 0.135
7 8 0.265
8 8 0.372
9 8 0.344
0 9 0.660
1 9 0.597
2 9 0.256
3 9 0.061
4 9 0.061
5 9 0.088
6 9 0.216
7 9 0.390
8 9 0.398
9 9 0.338
  };
  \end{axis}

  \begin{axis}[
    heatmap base,
    name=plotG,
    at={(plotF.north east)}, anchor=north west, xshift=2pt,
    title={Claude-Sonnet-4.6 (49.5\%)},
    xlabel={Column Depth (\%)},
    yticklabels={},
  ]
  \addplot3[surf, shader=interp, mesh/cols=10] table[meta=C] {
x y C
0 0 1.000
1 0 0.870
2 0 0.703
3 0 0.908
4 0 0.986
5 0 0.853
6 0 0.588
7 0 0.565
8 0 0.400
9 0 0.244
0 1 0.996
1 1 0.840
2 1 0.612
3 1 0.735
4 1 0.839
5 1 0.748
6 1 0.500
7 1 0.496
8 1 0.378
9 1 0.282
0 2 0.988
1 2 0.879
2 2 0.635
3 2 0.748
4 2 0.778
5 2 0.651
6 2 0.481
7 2 0.497
8 2 0.397
9 2 0.366
0 3 0.968
1 3 0.872
2 3 0.641
3 3 0.776
4 3 0.780
5 3 0.625
6 3 0.454
7 3 0.505
8 3 0.349
9 3 0.254
0 4 0.893
1 4 0.814
2 4 0.619
3 4 0.637
4 4 0.658
5 4 0.618
6 4 0.390
7 4 0.431
8 4 0.324
9 4 0.238
0 5 0.955
1 5 0.749
2 5 0.535
3 5 0.557
4 5 0.635
5 5 0.585
6 5 0.346
7 5 0.399
8 5 0.283
9 5 0.132
0 6 0.799
1 6 0.599
2 6 0.422
3 6 0.474
4 6 0.633
5 6 0.634
6 6 0.454
7 6 0.474
8 6 0.252
9 6 0.058
0 7 0.521
1 7 0.446
2 7 0.270
3 7 0.283
4 7 0.448
5 7 0.573
6 7 0.448
7 7 0.404
8 7 0.145
9 7 0.040
0 8 0.627
1 8 0.487
2 8 0.220
3 8 0.094
4 8 0.206
5 8 0.438
6 8 0.474
7 8 0.457
8 8 0.126
9 8 0.019
0 9 0.808
1 9 0.601
2 9 0.335
3 9 0.131
4 9 0.157
5 9 0.344
6 9 0.468
7 9 0.519
8 9 0.172
9 9 0.000
  };
  \end{axis}

  \begin{axis}[
    heatmap base,
    name=plotH,
    at={(plotG.north east)}, anchor=north west, xshift=2pt,
    title={Kimi-K2.5 (70.3\%)},
    xlabel={Column Depth (\%)},
    yticklabels={},
  ]
  \addplot3[surf, shader=interp, mesh/cols=10] table[meta=C] {
x y C
0 0 0.360
1 0 0.336
2 0 0.298
3 0 0.427
4 0 0.439
5 0 0.178
6 0 0.078
7 0 0.565
8 0 0.669
9 0 0.342
0 1 0.645
1 1 0.506
2 1 0.427
3 1 0.496
4 1 0.419
5 1 0.244
6 1 0.165
7 1 0.427
8 1 0.388
9 1 0.209
0 2 0.860
1 2 0.742
2 2 0.645
3 2 0.643
4 2 0.553
5 2 0.267
6 2 0.121
7 2 0.347
8 2 0.289
9 2 0.129
0 3 0.946
1 3 0.761
2 3 0.611
3 3 0.601
4 3 0.564
5 3 0.281
6 3 0.017
7 3 0.214
8 3 0.222
9 3 0.055
0 4 1.000
1 4 0.832
2 4 0.718
3 4 0.586
4 4 0.489
5 4 0.291
6 4 0.003
7 4 0.208
8 4 0.255
9 4 0.147
0 5 0.882
1 5 0.766
2 5 0.755
3 5 0.611
4 5 0.510
5 5 0.265
6 5 0.000
7 5 0.250
8 5 0.338
9 5 0.300
0 6 0.612
1 6 0.502
2 6 0.514
3 6 0.464
4 6 0.437
5 6 0.308
6 6 0.043
7 6 0.168
8 6 0.221
9 6 0.198
0 7 0.373
1 7 0.304
2 7 0.280
3 7 0.328
4 7 0.403
5 7 0.323
6 7 0.094
7 7 0.234
8 7 0.202
9 7 0.153
0 8 0.346
1 8 0.377
2 8 0.305
3 8 0.311
4 8 0.382
5 8 0.243
6 8 0.081
7 8 0.295
8 8 0.246
9 8 0.170
0 9 0.404
1 9 0.464
2 9 0.251
3 9 0.248
4 9 0.376
5 9 0.255
6 9 0.168
7 9 0.325
8 9 0.288
9 9 0.286
  };
  \end{axis}

  \begin{axis}[
    hide axis,
    scale only axis,
    name=cbar_span,
    at={(plotD.north east)}, anchor=north west, xshift=8pt,
    width=0.001\textwidth,
    height={\dimexpr 2\dimexpr 0.22\textwidth\relax + 42pt\relax},
    point meta min=0, point meta max=1,
    colormap name=RdBu,
    colorbar right,
    every colorbar/.append style={
      width=0.12cm,
      ytick={0,0.5,1.0},
      yticklabels={Low,Mid,High},
      yticklabel style={font=\fontsize{5}{6}\selectfont},
      ylabel={},
      extra description/.code={\node[rotate=-90, font=\fontsize{6}{7}\selectfont, anchor=south]
        at (rel axis cs:4.20,0.5) {Relative Accuracy (\%)};},
    },
  ]
  \addplot[draw=none, forget plot] coordinates {(0,0) (1,0)};
  \end{axis}

  \end{tikzpicture}
  \vspace{-1em}
  \caption{Cell-retrieval accuracy heatmaps for eight additional evaluated models. Per-model min--max normalised accuracy across the 10$\times$10 row $\times$ column grid.}
  \label{fig:needle_heatmap_appendix}
\end{figure}

\section{More Analysis on Reasoning}
\label{app:reasoning_budget}

Following Section~\ref{sec:experiments} and Figure~\ref{fig:reasoning-cost}, we further observe that WildTableBench consistently rewards thinking across all question categories.
For example, when comparing high \textit{vs.}\ low reasoning effort in Gemini-3-Pro, questions from the Hypothetical category appear among the most improved categories, with gains of around one quarter.
This highlights both the annotation quality and the challenge of WildTableBench: its questions genuinely require careful reasoning rather than simple, direct answers.
These results also suggest that WildTableBench can serve as a useful standard for measuring how models improve as they evolve from instruction-tuned to thinking variants.

\section{Accuracy by Image Type}
\label{app:image_type}

Table~\ref{tab:image_type} reports per-model accuracy on Spreadsheet (54.6\%) and Non-Spreadsheet (45.4\%) images, broken down by question category. Effects vary by model and by category: C4 (Hypothetical) shows a consistent NS advantage across nearly all models (avg.\ $-$9.1\%), whereas C3 (Verification) tends to favour Spreadsheet images (avg.\ $+$5.1\%).

\begin{table*}[t]
\centering
\renewcommand{\arraystretch}{1.1}
\setlength{\tabcolsep}{3.5pt}
\small
\caption{Accuracy (\%) by image type (Spreadsheet 54.6\% vs.\ Non-Spreadsheet 45.4\%) and question category across all 21 models. \textbf{Avg} = overall accuracy on that image type.}
\label{tab:image_type}
\resizebox{\linewidth}{!}{%
\begin{tabular}{l|ccccc|c|ccccc|c}
\hline
\noalign{\vskip 3pt}
\multirow{2}{*}{\textbf{Model}} & \multicolumn{6}{c|}{\textbf{Spreadsheet} (54.6\%)} & \multicolumn{6}{c}{\textbf{Non-Spreadsheet} (45.4\%)} \\
& C1 & C2 & C3 & C4 & C5 & Avg & C1 & C2 & C3 & C4 & C5 & Avg \\
\noalign{\vskip 3pt}
\hline
\noalign{\vskip 1pt}
\multicolumn{13}{l}{\textbf{Proprietary LMMs}} \\
\noalign{\vskip 1pt}
\hline
Gemini-3-Pro      & \rgb{61.6}61.6 & \rgb{74.3}74.3 & \rgb{69.7}69.7 & \rgb{75.0}75.0 & \rgb{52.9}52.9 & 67.4 & \rgb{68.2}68.2 & \rgb{70.1}70.1 & \rgb{73.5}73.5 & \rgb{75.9}75.9 & \rgb{59.1}59.1 & 69.2 \\
Gemini-3-Flash    & \rgb{46.2}46.2 & \rgb{45.9}45.9 & \rgb{69.7}69.7 & \rgb{46.7}46.7 & \rgb{33.8}33.8 & 46.0 & \rgb{51.2}51.2 & \rgb{53.7}53.7 & \rgb{47.1}47.1 & \rgb{72.4}72.4 & \rgb{45.5}45.5 & 53.2 \\
Seed-2.0-Pro      & \rgb{67.5}67.5 & \rgb{51.9}51.9 & \rgb{66.7}66.7 & \rgb{44.4}44.4 & \rgb{42.3}42.3 & 54.3 & \rgb{41.9}41.9 & \rgb{47.3}47.3 & \rgb{70.4}70.4 & \rgb{40.0}40.0 & \rgb{20.0}20.0 & 45.6 \\
GPT-5.2           & \rgb{58.6}58.6 & \rgb{45.3}45.3 & \rgb{60.6}60.6 & \rgb{51.7}51.7 & \rgb{25.0}25.0 & 47.3 & \rgb{40.9}40.9 & \rgb{47.5}47.5 & \rgb{52.9}52.9 & \rgb{62.1}62.1 & \rgb{31.8}31.8 & 46.3 \\
Claude-Opus-4.6   & \rgb{58.6}58.6 & \rgb{50.7}50.7 & \rgb{72.7}72.7 & \rgb{45.0}45.0 & \rgb{22.1}22.1 & 48.8 & \rgb{38.6}38.6 & \rgb{45.2}45.2 & \rgb{50.0}50.0 & \rgb{72.4}72.4 & \rgb{20.5}20.5 & 43.9 \\
Claude-Sonnet-4.6 & \rgb{37.4}37.4 & \rgb{29.1}29.1 & \rgb{54.5}54.5 & \rgb{43.3}43.3 & \rgb{20.6}20.6 & 33.8 & \rgb{34.1}34.1 & \rgb{39.0}39.0 & \rgb{44.1}44.1 & \rgb{62.1}62.1 & \rgb{15.9}15.9 & 37.8 \\
GPT-5-mini        & \rgb{12.6}12.6 & \rgb{27.0}27.0 & \rgb{48.5}48.5 & \rgb{35.0}35.0 & \rgb{16.2}16.2 & 24.8 & \rgb{29.5}29.5 & \rgb{22.6}22.6 & \rgb{35.3}35.3 & \rgb{31.0}31.0 & \rgb{25.0}25.0 & 25.9 \\
GPT-o3                & \rgb{16.2}16.2 & \rgb{6.8}6.8 & \rgb{30.3}30.3 & \rgb{5.0}5.0 & \rgb{13.2}13.2 & 11.8 & \rgb{20.5}20.5 & \rgb{15.3}15.3 & \rgb{35.3}35.3 & \rgb{20.7}20.7 & \rgb{18.2}18.2 & 18.9 \\
GPT-4o            & \rgb{3.0}3.0 & \rgb{3.4}3.4 & \rgb{21.2}21.2 & \rgb{0.0}0.0 & \rgb{10.3}10.3 & 5.4 & \rgb{6.8}6.8 & \rgb{4.0}4.0 & \rgb{26.5}26.5 & \rgb{0.0}0.0 & \rgb{2.3}2.3 & 6.1 \\
\hline
\noalign{\vskip 1pt}
\multicolumn{13}{l}{\textbf{Open-source LMMs: Thinking}} \\
\noalign{\vskip 1pt}
\hline
Kimi-K2.5         & \rgb{47.5}47.5 & \rgb{52.0}52.0 & \rgb{66.7}66.7 & \rgb{51.7}51.7 & \rgb{26.5}26.5 & 47.8 & \rgb{51.2}51.2 & \rgb{51.7}51.7 & \rgb{67.6}67.6 & \rgb{62.1}62.1 & \rgb{61.5}61.5 & 55.5 \\
Qwen3-VL-235B-T   & \rgb{38.4}38.4 & \rgb{32.4}32.4 & \rgb{51.5}51.5 & \rgb{36.7}36.7 & \rgb{25.0}25.0 & 34.8 & \rgb{36.4}36.4 & \rgb{35.6}35.6 & \rgb{32.4}32.4 & \rgb{44.8}44.8 & \rgb{29.5}29.5 & 35.4 \\
Qwen3-VL-32B-T    & \rgb{29.3}29.3 & \rgb{28.4}28.4 & \rgb{45.5}45.5 & \rgb{20.0}20.0 & \rgb{30.9}30.9 & 29.2 & \rgb{25.0}25.0 & \rgb{28.1}28.1 & \rgb{35.3}35.3 & \rgb{20.7}20.7 & \rgb{29.5}29.5 & 28.0 \\
GLM-4.6V          & \rgb{16.2}16.2 & \rgb{22.3}22.3 & \rgb{33.3}33.3 & \rgb{21.7}21.7 & \rgb{25.0}25.0 & 22.1 & \rgb{22.7}22.7 & \rgb{25.4}25.4 & \rgb{35.3}35.3 & \rgb{37.9}37.9 & \rgb{27.3}27.3 & 27.4 \\
Qwen3-VL-8B-T     & \rgb{9.1}9.1 & \rgb{16.2}16.2 & \rgb{30.3}30.3 & \rgb{10.0}10.0 & \rgb{13.2}13.2 & 14.2 & \rgb{9.1}9.1 & \rgb{14.1}14.1 & \rgb{26.5}26.5 & \rgb{31.0}31.0 & \rgb{11.4}11.4 & 15.9 \\
Qwen3-VL-4B-T     & \rgb{5.0}5.0 & \rgb{6.0}6.0 & \rgb{24.2}24.2 & \rgb{1.7}1.7 & \rgb{13.2}13.2 & 7.8 & \rgb{4.5}4.5 & \rgb{4.9}4.9 & \rgb{26.5}26.5 & \rgb{13.8}13.8 & \rgb{3.7}3.7 & 7.6 \\
Qwen3-VL-2B-T     & \rgb{4.0}4.0 & \rgb{2.0}2.0 & \rgb{6.1}6.1 & \rgb{0.0}0.0 & \rgb{1.5}1.5 & 2.4 & \rgb{4.5}4.5 & \rgb{2.7}2.7 & \rgb{23.5}23.5 & \rgb{3.4}3.4 & \rgb{9.3}9.3 & 6.1 \\
\hline
\noalign{\vskip 1pt}
\multicolumn{13}{l}{\textbf{Open-source LMMs: Instruct}} \\
\noalign{\vskip 1pt}
\hline
Qwen3-VL-235B-I   & \rgb{21.2}21.2 & \rgb{23.0}23.0 & \rgb{39.4}39.4 & \rgb{18.3}18.3 & \rgb{23.5}23.5 & 23.3 & \rgb{27.3}27.3 & \rgb{29.9}29.9 & \rgb{29.4}29.4 & \rgb{27.6}27.6 & \rgb{22.7}22.7 & 28.4 \\
Qwen3-VL-32B-I    & \rgb{27.3}27.3 & \rgb{17.6}17.6 & \rgb{42.4}42.4 & \rgb{21.7}21.7 & \rgb{23.5}23.5 & 23.5 & \rgb{20.5}20.5 & \rgb{26.6}26.6 & \rgb{41.2}41.2 & \rgb{37.9}37.9 & \rgb{13.6}13.6 & 26.5 \\
Qwen3-VL-8B-I     & \rgb{13.1}13.1 & \rgb{6.8}6.8 & \rgb{18.2}18.2 & \rgb{6.7}6.7 & \rgb{7.4}7.4 & 9.3 & \rgb{4.5}4.5 & \rgb{10.2}10.2 & \rgb{17.6}17.6 & \rgb{13.8}13.8 & \rgb{13.6}13.6 & 11.0 \\
Qwen3-VL-4B-I     & \rgb{6.9}6.9 & \rgb{4.0}4.0 & \rgb{24.2}24.2 & \rgb{8.3}8.3 & \rgb{5.9}5.9 & 7.3 & \rgb{11.4}11.4 & \rgb{6.6}6.6 & \rgb{8.8}8.8 & \rgb{3.4}3.4 & \rgb{16.7}16.7 & 8.7 \\
Qwen3-VL-2B-I     & \rgb{4.0}4.0 & \rgb{1.4}1.4 & \rgb{27.3}27.3 & \rgb{1.7}1.7 & \rgb{7.4}7.4 & 5.1 & \rgb{2.3}2.3 & \rgb{5.1}5.1 & \rgb{17.6}17.6 & \rgb{3.4}3.4 & \rgb{11.4}11.4 & 6.7 \\
\hline
\end{tabular}}
\end{table*}

\end{document}